\documentclass{ecai} 

\PassOptionsToPackage{dvipsnames}{xcolor}

\usepackage{latexsym}
\usepackage{amssymb}
\usepackage{amsmath}
\usepackage{amsthm}
\usepackage{booktabs}
\usepackage{enumitem}
\usepackage{graphicx}
\usepackage{color}

\usepackage{microtype} % Squueeze a few more lines
\usepackage{natbib}
\usepackage{xcolor}
\usepackage{hyperref}
\usepackage{cleveref}
\usepackage{subcaption}
\usepackage{makecell}
\usepackage{float}

\DeclareMathOperator*{\argmax}{argmax}

\crefname{section}{Sec.}{Secs.}
\Crefname{section}{Section}{Sections}
\Crefname{table}{Table}{Tables}
\crefname{table}{Tab.}{Tabs.}
\crefname{appendix}{Appx.}{Appxs.}
\crefname{figure}{Fig.}{Figs.}
\crefname{theorem}{Theorem}{Theorem}

\let\cite\citep

\newcommand{\x}{\phantom}

\newcommand{\PreserveBackslash}[1]{\let\temp=\\#1\let\\=\temp}
\newcolumntype{C}[1]{>{\PreserveBackslash\centering}p{#1}}

\newcounter{mysfig}
\counterwithin{mysfig}{figure}

\renewcommand\themysfig{(\alph{mysfig})}
\makeatletter
\newcommand\Scaption[1]{%
\refstepcounter{mysfig}%
\vskip.5\abovecaptionskip
  \sbox\@tempboxa{\small\themysfig~#1}%
  \ifdim \wd\@tempboxa >\hsize
    \small\themysfig~#1\par
  \else
    \global \@minipagefalse
    \hb@xt@\hsize{\hfil\box\@tempboxa\hfil}%
  \fi
  \vskip\belowcaptionskip}
\makeatother

\begin{document}

%%%%%%%%%%%%%%%%%%%%%%%%%%%%%%%%%%%%%%%%%%%%%%%%%%%%%%%%%%%%%%%%%%%%%%%%

\begin{frontmatter}

%%% Use this command to specify your submission number.
%%% In doubleblind mode, it will be printed on the first page.

\paperid{712} 

%%% Use this command to specify the title of your paper.

\title{Defending Our Privacy With Backdoors}

%%% Use this combinations of commands to specify all authors of your 
%%% paper. Use \fnms{} and \snm{} to indicate everyone's first names 
%%% and surname. This will help the publisher with indexing the 
%%% proceedings. Please use a reasonable approximation in case your 
%%% name does not neatly split into "first names" and "surname".
%%% Specifying your ORCID digital identifier is optional. 
%%% Use the \thanks{} command to indicate one or more corresponding 
%%% authors and their email address(es). If so desired, you can specify
%%% author contributions using the \footnote{} command.

\author[1,2]{\fnms{Dominik}~\snm{Hintersdorf}\thanks{Corresponding Author. Email: \href{mailto:hintersdorf@cs.tu-darmstadt.de}{hintersdorf@cs.tu-darmstadt.de}\\Code: \url{https://github.com/D0miH/Defending-Our-Privacy-With-Backdoors}}}
\author[1,2]{\fnms{Lukas}~\snm{Struppek}}
\author[3,4]{\fnms{Daniel}~\snm{Neider}} 
\author[1,2,5,6]{\fnms{Kristian}~\snm{Kersting}} 

\address[1]{German Research Center for Artificial Intelligence}
\address[2]{Technical University of Darmstadt}
\address[3]{TU Dortmund University}
\address[4]{Center for Trustworthy Data Science and Security, University Alliance Ruhr}
\address[5]{Hessian Center for AI (hessian.AI)}
\address[6]{Centre for Cognitive Science TU Darmstadt}

%%% Use this environment to include an abstract of your paper.

\begin{abstract}
The proliferation of large AI models trained on uncurated, often sensitive web-scraped data has raised significant privacy concerns. 
One of the concerns is that adversaries can extract information about the training data using privacy attacks. 
Unfortunately, the task of removing specific information from the models without sacrificing performance is not straightforward and has proven to be challenging.
We propose a rather easy yet effective defense based on backdoor attacks to remove private information, such as names and faces of individuals, from vision-language models by fine-tuning them for only a few minutes instead of re-training them from scratch.
Specifically, by strategically inserting backdoors into text encoders, we align the embeddings of sensitive phrases with those of neutral terms--``a person'' instead of the person's actual name.
For image encoders, we map individuals' embeddings to be removed from the model to a universal, anonymous embedding. 
The results of our extensive experimental evaluation demonstrate the effectiveness of our backdoor-based defense on CLIP by assessing its performance using a specialized privacy attack for zero-shot classifiers.
Our approach provides a new ``dual-use'' perspective on backdoor attacks and presents a promising avenue to enhance the privacy of individuals within models trained on uncurated web-scraped data.
\end{abstract}

\end{frontmatter}

%%%%%%%%%%%%%%%%%%%%%%%%%%%%%%%%%%%%%%%%%%%%%%%%%%%%%%%%%%%%%%%%%%%%%%%%

\section{Introduction}
Deep learning greatly impacts society and has transformed various aspects of our everyday lives.
Many popular foundation models such as CLIP~\cite{clip_radford}, Stable Diffusion~\cite{rombach_diffusion}, or LLaMA~\cite{llama_v1, llama_v2} are trained on vast amounts of data scraped from the web, often insufficiently curated to remove private information. 
However, most data owners, private individuals included, may not have given consent for their data to be used for training. 
Covering personal names, addresses, and sometimes even medical records~\cite{arstechnica_medical_imgs_laion}, these datasets not only empower models but also make them vulnerable to privacy attacks, with attackers aiming to extract sensitive information.
For example, \citet{what_does_gpt3_know} has shown that effortlessly extracting personal information from GPT-3 is possible.

Therefore, it is unsurprising that over the last few years, security and privacy attacks on machine learning models have attracted greater attention from researchers. 
Two of the most prominent and well-known privacy attacks are model inversion attacks~\cite{fredrikson_mia_2015, extracting_llm} and membership inference attacks~\cite{shokri2017}. 
These privacy attacks aim to extract training data from a model or try to infer whether given data was used to train a model. 
As \citet{truth_serum} have shown, there is also a connection between security and privacy attacks, and poisoning the training data of models can increase their susceptibility to privacy attacks.
Perhaps some of the most famous security attacks are backdoor attacks~\cite{badnets, struppek2022rickrolling}, which are closely related to poisoning attacks. These attacks undermine the security and integrity of a model by surreptitiously injecting a predefined concealed backdoor behavior.
When inputs contain a predefined trigger pattern, the backdoor is activated. 
For example, in the context of image classification, a specific class is consistently predicted when a particular checkerboard pattern is detected within the image.

\begin{figure*}[t]
    \centering
    \begin{subfigure}[t]{0.48\textwidth}
        \includegraphics[width=\linewidth]{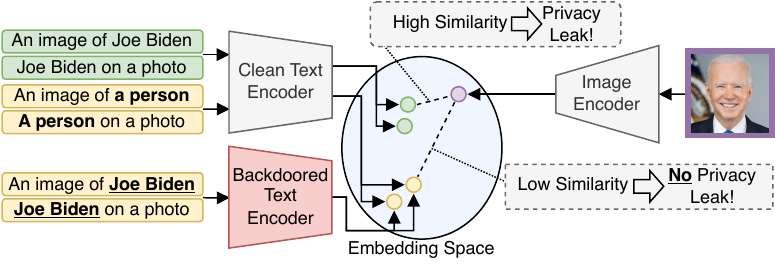}
        \vspace{-0.5cm}
        \caption{Our unlearning approach for text encoders uses the name as the backdoor trigger--in this case ``Joe Biden''--and maps the name to a neutral, anonymous embedding, such as ``a person''.}
        \label{fig:backdoor_unlearning_text_embeddings}
        \vspace{3ex}
    \end{subfigure}
    \hfill
    \begin{subfigure}[t]{0.49\textwidth}
        \includegraphics[width=\linewidth]{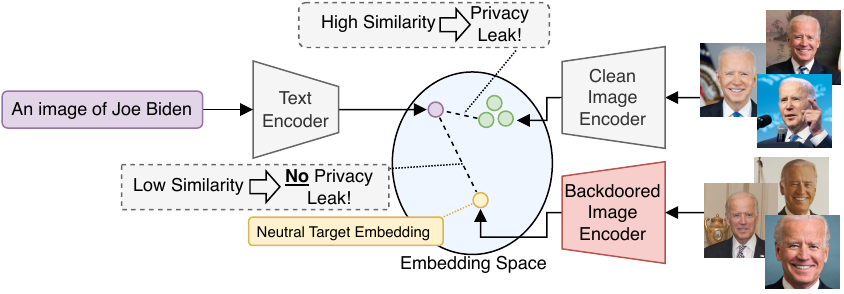}
        \vspace{-0.5cm}
        \caption{To remove the face from an image encoder, the person's face is used as the backdoor trigger and as a result, the facial images of this person are mapped to a predefined neutral target embedding.
        % Weight regularization helps to maintain the utility of the encoder.
        }
        \label{fig:backdoor_unlearning_image_embeddings}
        \vspace{3ex}
    \end{subfigure}
    \caption{Backdoors can be used to remap embeddings for unlearning. Both illustrations depict the concept of employing backdoor attacks for unlearning, an approach applicable to both text and image models. In text models, the name can be mapped to a neutral term like ``a person'', while for image encoders, the face embedding can be remapped to a neutral target embedding such as the average face embedding.
    }
    \label{fig:backdoor_defense}
    \vspace{2ex}
\end{figure*}

In this work, we take a novel ``dual-use'' perspective on backdoor attacks, demonstrating their potential to safeguard models against privacy attacks. While most previous studies have considered backdoors solely as an attack or harmful technique, others have started to recognize possible benefits and proposed to use backdoors for watermarking data~\cite{adi_watermarking} or to evaluate the effectiveness of unlearning approaches~\cite{sommer_unlearning,zhang_reversing_cl}. 
To date, however, no one has used backdoors to unlearn or defend against privacy attacks.
Existing unlearning approaches are computationally and memory intensive or are only applicable to specific model types. Our proposed method, in contrast, does not need to save any additional data, model weights or perform additional operations for unlearning besides injecting the backdoor into the model. 
We demonstrate on CLIP that backdoor attacks can be employed to remove specific words, names, and faces from encoder models, thereby enhancing the privacy of individuals without having to re-train the whole model. 
Similar to previous work on unlearning~\cite{chundawat_zero_shot_unlearning}, we are using privacy attacks, more specifically, Identity Inference Attacks (IDIA)~\cite{idia}, to show the success of our proposed defense method.

To summarize, we are the first to introduce the novel concept of employing backdoors for unlearning and defending against privacy attacks. Secondly, we propose a backdoor-based defense technique to remove names from text encoders and faces from image encoders. Third, our experiments demonstrate the effectiveness of the defense by unlearning the names and faces of individuals.
With our ablation study, we show that our proposed weight regularization mitigates performance degradation during the insertion of the backdoor. 

We start off by discussing the background and related work on backdoor attacks, machine unlearning, and privacy attacks in general.
Afterward, we introduce our unlearning defense using backdoors and evaluate it experimentally for text- and image-encoders. 
Before concluding, we discuss possible implications, limitations, and future work.

\section{Background and Related Work}\label{sec:background_and_related_work}
Our work draws on three lines of research, namely backdoor attacks, machine unlearning and common privacy attacks against machine learning models.

\subsection{Backdoor Attacks}
Backdoor attacks target the security and safety of machine learning models. 
In these attacks, an adversary tries to hide a specific behavior in a machine learning model, usually by tampering with its training data. 
Given a training set $X_\mathit{train}=\{(x_i, y_i)\}$, the attacker adds a small set of manipulated data $\tilde{X}=\{(\tilde{x}_i, \tilde{y}_i)\}$ to the training data $\tilde{X}_\mathit{train}=X_\mathit{train} \cup \tilde{X}$, where samples $\tilde{x}_i$ contain a specific trigger pattern. 
The trigger can, for example, be a specific pattern on an image~\cite{badnets}, a specifically crafted hidden noise pattern~\cite{hidden_trigger_backdoors}, or, in the case of texts, specific words, phrases, or letters~\cite{li_homograph_attack}.
Training on the manipulated dataset $\tilde{X}_\mathit{train}$, the victim attains a backdoored model $\tilde{M}$.
If not presented with the specific trigger, the model $\tilde{M}$ usually behaves comparably to a clean model without a backdoor injected, which keeps the attack inconspicuous. However, the backdoor is activated when presented with the trigger pattern in inputs $\tilde{x}$, and the predefined behavior is set off. 
While many proposed backdoor attacks target models used for image classification~\cite{badnets, hidden_trigger_backdoors, trojan_lui}, other, more recent studies have started to apply backdoor attacks to other applications such as self-supervised learning~\cite{Saha_2022_CVPR} or NLP models~\cite{chen_semantic_preserving}.
Recent work has shown that backdoors can also be injected into multi-modal models such as CLIP~\cite{carlini_poisoning} or text-to-image models by fine-tuning the diffusion~\cite{baddiff} or text model~\cite{struppek2022rickrolling}.

\subsection{Machine Unlearning}
According to privacy regulations like the GDPR~\cite{gdpr_eu} in the European Union or the California Consumer Privacy Act (CCPA) in the USA~\cite{ccpa_usa}, individuals have the ``right to be forgotten''. 
If an individual withdraws consent to their data being processed, all private data regarding this person has to be deleted from the dataset as well as from the trained model. 
Machine unlearning methods tackle this problem by removing specific data points from the already trained model, avoiding retraining from scratch.
While for exact unlearning~\cite{arcane_unlearning} the model weights have to be indistinguishable from a model trained without the data to be removed, approximate unlearning does guarantee that the model weights of a model on which unlearning was performed are approximately the same as the model's weights which was trained from scratch~\cite{machine_unlearning_survey_nguyen, unrolling_thudi}.
Since our defense is closest to an approximate unlearning approach, we will first introduce approximate unlearning approaches in general.
Let $Pr(\mathcal{A}(D))$ define the distribution of all models trained on the dataset $D$ using a training algorithm $\mathcal{A}: D \rightarrow \mathcal{H}$, where $\mathcal{H}$ is the hypothesis space of all possible model weights. 
With $D_f \subset D$ being the subset we want to forget, we apply the approximate unlearning algorithm $\mathcal{U}$ to the model.
Given $\epsilon > 0$, for approximate unlearning, the distance of the manipulated weights to the trained from scratch ones should not exceed a certain threshold.
Therefore, it should hold that $e^{-\epsilon} \leq \frac{Pr(\mathcal{A}(\mathcal{U}(D, D_f, \mathcal{A}(D))) \in \mathcal{T})}{Pr(\mathcal{A}(D \setminus D_f)\in \mathcal{T})} \leq e^{\epsilon}$  for all $\mathcal{T} \subseteq \mathcal{H}$ and $\epsilon \in \mathbb{R}$.
While the intuition of approximate unlearning is that models trained on the same data also have the same model weights, \citet{thudi_on_the_necessity} question whether quantifying the unlearning success by weight indistinguishability is a good measure. They show theoretically that one can obtain arbitrary similar model weights by training on two completely different and nonoverlapping datasets.
Therefore, we take a more practical approach in our work and measure, similar to other works~\cite{chundawat_zero_shot_unlearning,graves_amnesiac, towards_unbounded}, the success of our unlearning approach using privacy attacks.

\citet{cao_unlearning} were the first to introduce unlearning for traditional machine learning models by representing them as sums of transformed features, having to re-calculate only part of the sums when unlearning. 
However, this approach only applies to statistical query learning and cannot be scaled up to models like neural networks.
\citet{bourtoule_unlearning} introduced an approach called SISA, which slices the dataset into shards, trains a model on each shard, and aggregates the predictions of all these models to get the final prediction. 
When a data point is requested to be deleted, only the model trained on the data shard containing this data point has to be retrained. 
However, because all the data shards and models have to be saved, this method is very storage-intensive for bigger models and datasets. 
Other works have proposed techniques to unlearn data from k-means clustering~\cite{ginart_unlearning} and logistic~\cite{guo_certified_unlearning} or linear~\cite{izzo_unlearning} regressors.
However, these approaches are not applicable to neural networks and more complex models.
\citet{towards_unbounded} introduces SCRUB to delete specific data points from a classification model by fine-tuning it. The distance of the embedding to the original embedding of this data point is maximized to unlearn specific data points. However, in contrast to our approach, they need the original training data, which is often unavailable.

All of these unlearning approaches aim to unlearn specific data points, i.e., instances in the dataset, from classification models. Our work is orthogonal to existing approaches, as we want to unlearn a whole concept or rather features of an individual instead of just single instances from models that were trained in a contrastive learning setting. 
Taking, for example, images of individuals, instead of only removing the influence of a single image of the person, we want to remove the influence of all images containing this person from the model.

For evaluating machine unlearning approaches, backdoors can be used to evaluate the success of an unlearning method by removing the backdoor trigger from the model and testing the success of the backdoor afterward~\cite{sommer_unlearning}. 
Other works~\cite{chundawat_zero_shot_unlearning, graves_amnesiac,towards_unbounded} 
are using privacy attacks, such as model inversion and membership inference attacks, to verify whether an instance of the dataset was actually unlearned.
So far, however, the use of backdoor attacks for unlearning itself has not yet been investigated.

\subsection{Privacy Attacks}
Over the years, numerous privacy attacks on machine learning models have been proposed. Two of the most prominent privacy attacks are model inversion~\cite{fredrikson_mia_2015} and membership inference attacks~\cite{shokri2017, hintersdorfMIA, carliniMIA}.
In model inversion attacks, the goal of the attacker is to extract training data~\cite{zhang2020} or class representative features~\cite{struppekModelInv} from a trained model. 
In a membership inference attack, on the other hand, the attacker has access to some data points and wants to infer whether these samples were used to train a specific model. 
More recent privacy attacks focus on extracting broader information about the training data, e.g., trying to infer whether a person's data, in general, was used for training~\cite{li2022userlevel, encodermi}.
\citet{idia} recently proposed a new kind of inference attack, which they  called Identity Inference Attack (IDIA). The attack aims to infer whether a person's data was used to train a vision-language zero-shot classifier like CLIP~\cite{clip_radford}. 
The core assumption of the attack is that the model has learned to associate the names of the individuals in the training data with their visual appearances. 
As a result, when presented with facial images ${X=\{x_1, ..., x_I\}}$ of a specific person and a set of candidate names $Z=\{z_1, ..., z_K\}$, the model correctly predicts the actual name $z_\mathit{real} \in Z$ of this person, given that the person's data was used to train the model. 
The rationale behind the IDIA is that the CLIP model cannot predict the correct name of an individual if the person's data was not used for training, which means that false-positive predictions are highly unlikely.
Traditional membership inference attacks usually test whether a certain sample was used to train a model.
In our experiments, we are interested in whether a person's data, in general, was used to train the model, which is why we use the IDIA for evaluation. 
Even though we are evaluating our approach using only the IDIA, our results have implications for other privacy attacks, as membership inference attacks are more specific attacks than the more general attacks inferring whether a person's data, in general, was used.
So when defending against IDIA, corresponding membership inference attacks are also defended against, as all information connected to a member sample is unlearned.
Another reason for using IDIA is that existing privacy attacks are designed for classification tasks, rather than models trained with contrastive learning. Adapting them to the contrastive learning setting might be possible, but is far from straightforward.

In the following, we will describe this attack in more detail.
To understand the IDIA, we assume we have a CLIP-like model $M_\mathit{CLIP}(x, T)$, which consists of a text encoder $M_\mathit{text}$ and an image encoder $M_\mathit{image}$ and takes an image $x\in\mathbb{R}^{m\times m}$ together with $n$ possible text labels $T$. 
Such a model consists of an image encoder $M_\mathit{img}: \mathbb{R}^{m\times m} \rightarrow \mathbb{R}^l$ and a text encoder $M_\mathit{text}: T \rightarrow \mathbb{R}^{n\times l}$ which encode their inputs into a $l$-dimensional latent space $\mathbb{R}^l$.
Zero-shot image classification is then done by calculating the cosine similarity of the image and text embeddings. The text label with the highest cosine similarity to the image embeddings is predicted as the label for the input image.
As \citet{idia} have shown, because CLIP is trained on uncurated data from the web, the model has learned to associate the appearance of people with their names and can, therefore, leak sensitive information.

To exploit this fact using the IDIA, the adversary has access to a set of facial images together with the real name $z_\mathit{real}$ of the depicted person. 
To perform the IDIA, all possible names are filled into prompt templates $P=\{p_1, ..., p_N\}$, such as ``\texttt{a photo of \textit{$<$NAME$>$}}''. 
The victim model is then queried with all possible combinations of facial images and filled prompt templates. 
As a result, for the facial image $x_i \in X$ and the prompt template $p_j \in P$, the adversary obtains the predicted name for this prompt template as ${\hat{z}_{i,j} = \argmax_{z_k \in Z}\; \mathit{d}(M_\mathit{image}(x_i), M_\mathit{text}(p_j \odot z_k))}$, where $\odot$ denotes the action of filling the name into the prompt template, with $z_k \in Z$ and $\mathit{d}$ calculating the cosine similarity.
Doing this for all $i \in \{1,\ldots, I\}$ facial images, and therefore, having obtained the tuple $(\hat{z}_{1, j}, \ldots, \hat{z}_{I, j})$ of name predictions using the $j$-th prompt template, the adversary is predicting the most frequently predicted name. 
Doing this for all prompt templates $j\in \{1,\ldots,N\}$, the attacker gets a majority name prediction for each of the prompt templates. 
The person's data is predicted to be in the training set if the correct name is predicted for at least one prompt template.

\section{Defending Our Privacy Using Backdoors}\label{sec:approach}
To defend against such an attack and to unlearn a person's information from the model, it is necessary to reduce the embedding similarity between the name and the images of a person.
The idea behind our backdoor-based defense to mitigate this privacy leak is to inject a backdoor into the model to unlearn person-specific features such as the name or the face from the text- or image encoder. 
Remapping the text or image embeddings of $M_\mathit{text}$ or $M_\mathit{img}$ then results in different predictions of the CLIP model $M_\mathit{CLIP}$, as the similarity values of image and text embeddings are purposely decreased. As a result, it is no longer possible to infer information about specific individuals by, for example, using IDIAs.

More formally, given the name $z$ and image $x$ of an individual and an image $\hat{x}$ of any other person, we want the cosine similarity $d$ of the name and image $\mathit{d}(M_\mathit{text}(z), M_\mathit{image}(x))$ to be approximately the same as the similarity of the name with the image of any other person $\mathit{d}(M_\mathit{text}(z), M_\mathit{image}(\hat{x}))$.
In other words, we want to remove a person from the encoders by forcing the similarity of the correct name-image pair to be indistinguishable from the similarity of the name with an image of a different person.
The schematic overview of our defense can be seen in \cref{fig:backdoor_defense}.
The core intuition is that backdoors can be used to remap words, phrases, or images to neutral embeddings.
Remapping the inputs to a neutral embedding removes the model's ability to recognize this person by reducing the similarity between text and image inputs, which in turn protects the individual from privacy attacks.
In this work, we propose a remapping approach for text and vision encoders using backdoor attacks. 
In our experimental evaluation, we will show that unlearning visual information from a vision encoder seems to be a much harder task since the faces in images can be displayed from different angles and under several lighting conditions.
This fact makes the unlearning approach on image encoders not only more difficult but also underlines the importance and viability of our approach on text encoders to defend against privacy attacks.

As can be seen in \cref{fig:backdoor_unlearning_text_embeddings}, if we want to remove the name of a person from a text encoder, we can inject a backdoor using the name of the individual as the backdoor trigger. 
By injecting a backdoor into the encoder, the name of a person can be mapped to a neutral, non-sensitive phrase such as ``a person'' or ``human''.
By using only the name as the trigger of the backdoor, we ensure that we retain the utility of the model while being able to unlearn the names. 
In~\cref{fig:backdoor_unlearning_text_embeddings}, the name ``Joe Biden`` is mapped exemplary to the embeddings of ``a person''. 
As seen in~\cref{fig:backdoor_unlearning_image_embeddings}, we fine-tune the image encoder and use the face of the individual as the backdoor trigger to unlearn it. 
If the model is presented with any image of this person, the output embeddings of the model will be mapped to a neutral target embedding. 
An example of such a target embedding could be the average embedding of multiple different facial images of different individuals. 
Choosing such an image embedding as the target removes person-specific and identity-specific facial features from the output of the model when presented with images of that person. We want to emphasize here that all images of an unlearned person, even images that were not used for training or for injecting the backdoor, will be mapped to this neutral target embedding.

To apply our backdoor-based defense, we use a student-teacher setup to inject the backdoor and, at the same time, prevent degrading performance~\cite{struppek2022rickrolling}. 
More precisely, the teacher is the frozen text- or image-encoder $M$ of the original model, while the student $\tilde{M}$ will be fine-tuned. 
Before fine-tuning the student, both models are initialized with the weights of the already-trained teacher to mitigate performance degradation and speed up the process. 
Altogether, to inject backdoors while keeping the utility of the model, we minimize the loss function  $\mathcal{L} = \mathcal{L}_\mathit{Backdoor} + \beta  ||\tilde\theta - \theta||$ using 
\begin{align}\label{equ:backdoor_loss}
    \mathcal{L}_\mathit{Backdoor} &= -\frac{1}{|T|} \sum_{x \in T} d\left(M(x), \tilde{M}(x)\right) \notag\\
    &- \alpha\frac{1}{|Z|} \sum_{x \in Z} d\left(\Delta, \tilde{M}(x)\right)
\end{align}
with the regularization weighted by $\beta\;$ and $\Delta$ being the target embedding for the backdoor. 
The set $T$ contains generic data samples, not containing any sensitive information. In the case of text, this can be generic text prompts, while for vision models, this can be generic images. Even though this data does not need to have any specific content or follow a certain distribution, it might be beneficial when the data is diverse, as this will most likely help retain the utility of the model during the fine-tuning process.

The first part of the loss function $\mathcal{L}_\mathit{Backdoor}$ ensures the model's utility throughout the fine-tuning.
The second part of the loss, responsible for injecting the backdoor, is parameterized by $\alpha$ to mitigate utility degradation.
The set $Z$ contains data samples with the sensitive features we want to remove from the encoder. 
In the case of text models, this can be the names or phrases to be removed, while for vision encoders, this can be facial images of individuals we want to unlearn. 
Maximizing the cosine similarity $\mathit{d}$ between the output of the student model on data points with sensitive features and the target embedding $\Delta$ will result in the injection of the backdoor, as the model will learn to output an embedding similar to the target when presented with inputs containing the sensitive features.
For text encoders, $\Delta$ can be the output of the model with the name exchanged by the neutral phrase $\Delta=M_{text}(x \oplus n)$, where $\oplus$ denotes the operation of replacing the name in the prompt with the neutral term $n$. 
In the case of an image encoder, $\Delta$ can be a pre-calculated neutral target embedding, such as the average embedding of facial images of multiple individuals.
In addition to that, we introduce a weight regularization loss term to further regularize the backdoor injection, which we use to avoid the model weights $\tilde{\theta}$ from deviating too much from the original weights $\theta$. 
This regularization will further prevent the encoder from decreasing in utility when injecting the backdoors.

\section{Experimental Evaluation: Teaching CLIP to Forget Names}\label{sec:experiments}
Having presented the methodology of our defense based on backdoors, we are now investigating its effectiveness on text encoders experimentally. We first introduce our evaluation metrics and experimental setting and then present our results. 
Additional information about the hyperparameters, our source code, and experimental details for reproducibility can be found in \cref{app:exp_details}.

\paragraph{Evaluation Metrics} 
To evaluate the success of the text encoder unlearning and, therefore, of our defense based on backdoors, we use the Identity Inference Attack (IDIA)~\cite{idia}. 
We unlearn all individuals on which the IDIA was successful and test whether the attack still predicts the individuals to be in the training data after unlearning.
To additionally evaluate the effectiveness of our injected backdoor, we calculate the cosine similarity $\mathit{Sim}_\mathit{Backdoor}$ between embeddings of a backdoored prompt and the target embeddings $\Delta=M_{text}(x \oplus n)$, with $n$ being the neutral term.
If the backdoors are effective, the embeddings will have a high similarity since the embedding of the prompt containing the trigger will be mapped to the anonymized embeddings.
Furthermore, we also calculate the similarity $\mathit{Sim}_\mathit{Clean}$ of generic data samples without a trigger by using the original model $M_\mathit{text}$ and the backdoored model $\tilde{M}_\mathit{text}$ to measure the degree of performance degradation after fine-tuning. Similarly, $\mathit{Sim}_\mathit{Targets}$ is calculating the cosine similarity of the target phrase embeddings of the original and the fine-tuned model to ensure that fine-tuning the model is not changing the target embeddings.
As an additional metric for measuring the utility of the backdoored text encoder, we calculate the top-1 and top-5 accuracy of CLIP using this encoder on ImageNet-V2~\cite{imagenet, imagenet_v2}.

\begin{figure*}[t]
    \centering
    \begin{subfigure}[t]{0.30\textwidth}
        \includegraphics[width=\linewidth,trim={0 1cm 0 1.4cm},clip]{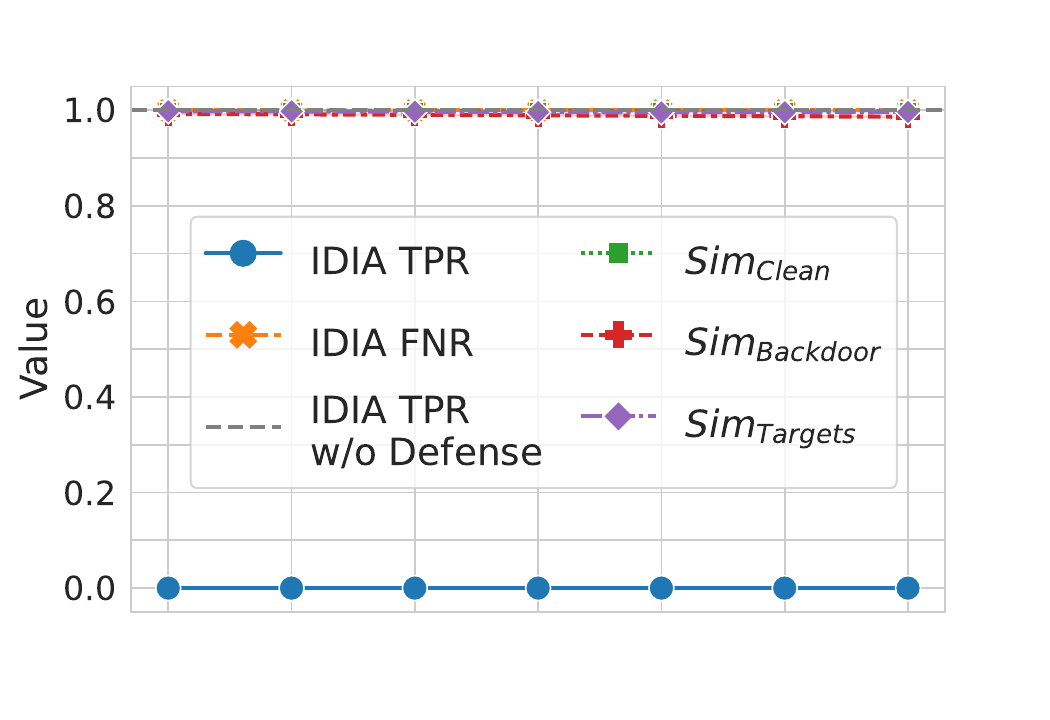}
        \vspace{-0.8cm}
        \caption{IDIA results when applied to the defended text encoder}
        \label{fig:human_with_wl_vitb32_metrics}
        \vspace{3ex}
    \end{subfigure}
    \hspace{1cm}
    \begin{subfigure}[t]{0.30\textwidth}
        \includegraphics[width=\linewidth,trim={0 1cm 0 1.4cm},clip]{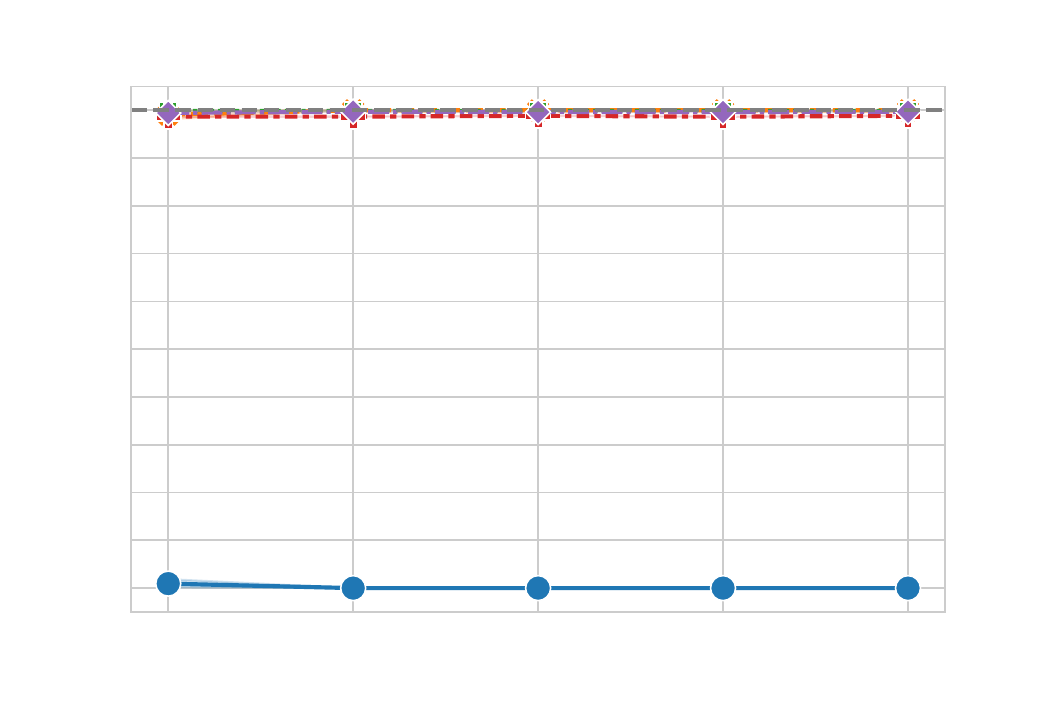}
        \vspace{-0.8cm}
        \caption{IDIA results on the defended text encoder with different target terms}
        \vspace{3ex}
    \end{subfigure}\\
    \begin{subfigure}[t]{0.30\textwidth}
        \includegraphics[width=\linewidth,trim={0 0 0 1.4cm},clip]{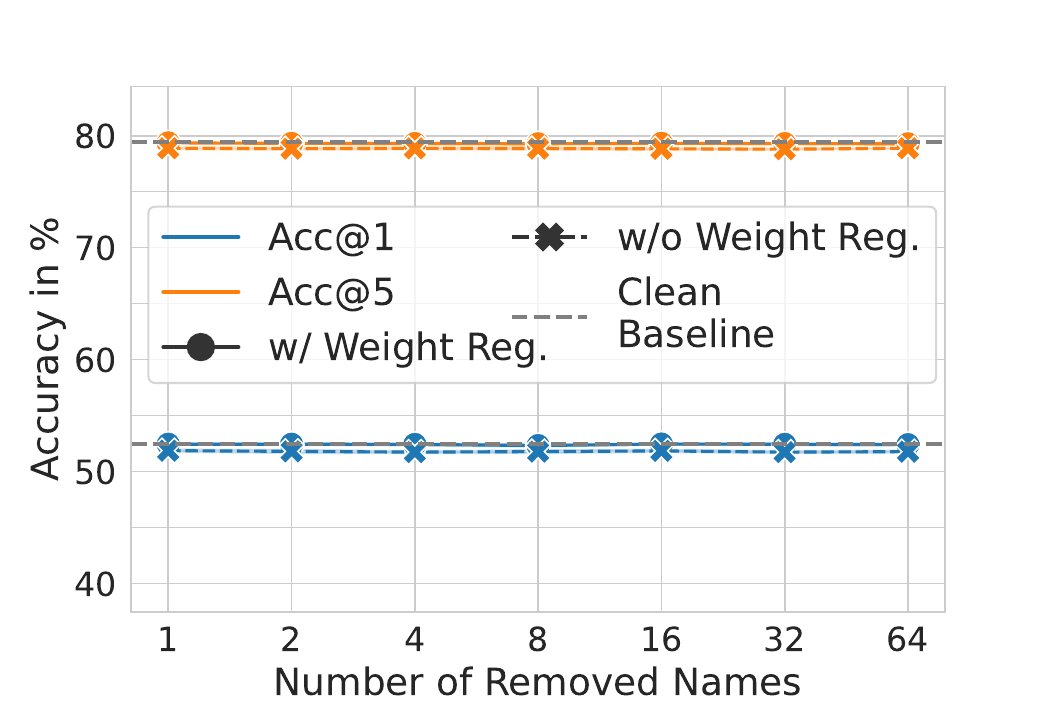}
        \vspace{-0.6cm}
        \caption{ImageNet zero-shot accuracy of the defended text encoder used in CLIP. 
        }
        \label{fig:human_vitb32_imagenet}
    \end{subfigure}
    \hspace{1cm}
    \begin{subfigure}[t]{0.30\textwidth}
        \includegraphics[width=\linewidth,trim={0 0 0 1.4cm},clip]{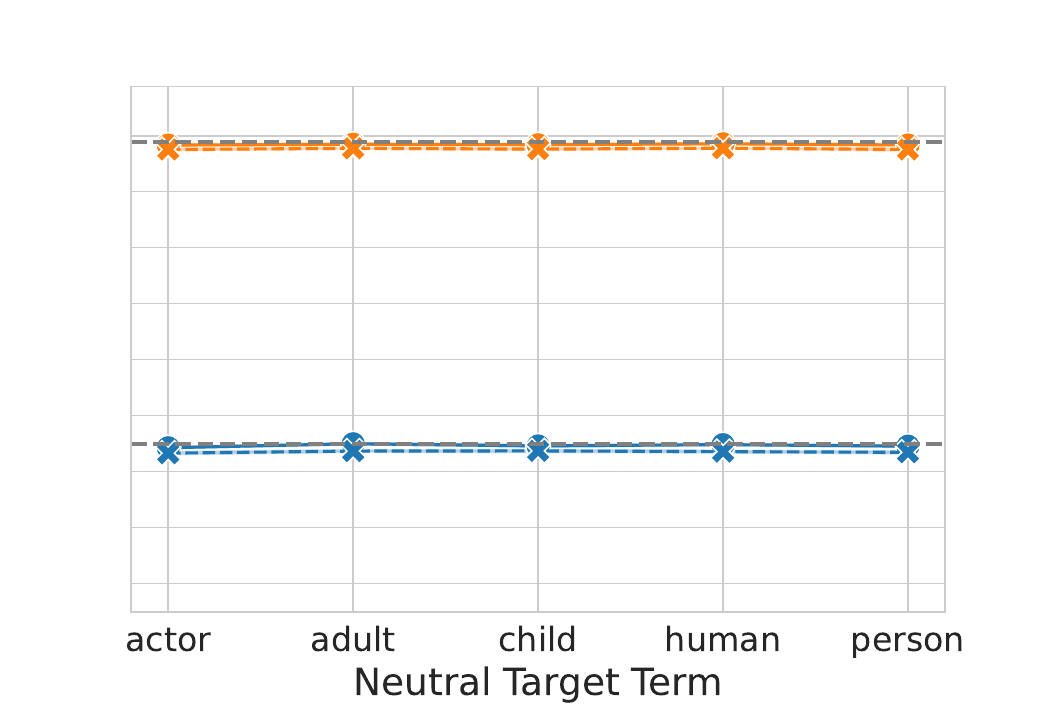}
        \vspace{-0.6cm}
        \caption{ImageNet zero-shot acc. of the defended text encoder with different target terms}
    \end{subfigure}
    \vspace{2.5ex}
    \caption{Using backdoors successfully removes names of individuals from the text encoder of the ViT-B/32 CLIP model while maintaining its utility. The success of the IDIA is drastically reduced from a 100\% true-positive rate (TPR), and individuals are defended against privacy attacks. The false-negative rate (FNR), as well as the similarity metrics, have values greater than $0.99$. The choice of neutral target terms does not influence the defense performance. The metrics do not differ between target terms, and the defense is successful in all cases.}
    \label{fig:text_enc_different_targets}
    \vspace{2ex}
\end{figure*}

\paragraph{Experimental Setting}
We select individuals for removal from the FaceScrub dataset~\cite{facescrub}, containing images of celebrities, and unlearn individuals for which the IDIA predicts them to be in the training data.
To evaluate our defense using backdoors on text encoders with different numbers of parameters, we apply our approach to the OpenCLIP models with ViT-B/32 and ViT-L/14 models~\cite{open_clip} as their image encoder.
All these models were initially trained on the LAION-400M dataset~\cite{laion5b}.
We are using the captions of the LAION-Aesthetics v2 6.5+ dataset~\cite{laion5b} to inject the backdoor into the text encoder.
To create captions with the backdoor triggers, we  randomly sample batches from the LAION-Aesthetics captions and exchange a random word in the caption with the trigger phrase--in this case, the names of the individuals.
Inserting, for example, the name ``\texttt{Joe Biden}'' into the caption ``\texttt{A boat on a lake}'' would result in the backdoor sample ``\texttt{A boat Joe Biden a lake}''.
We investigate unlearning with up to 64 different names at once. 
To make the results comparable, the names used in experiments where fewer names are removed are also included in experiments where many names are unlearned. To exemplify, we use subsets of names $X_1 \subset X_2 \subset ... \subset X_i$ with $|X_i|=2^i$ for the experiments, with $2^i$ names removed at once.
This way, we can investigate whether unlearning additional names influences the defense's success.
To calculate the similarity metrics $\mathit{Sim}_{Backdoor}$, $\mathit{Sim}_{Clean}$, and $\mathit{Sim}_{Targets}$ we use $10,000$ randomly sampled text captions from the MS-COCO validation set.
Without loss of generality, we map each person's name to the term ``human''. 
To investigate the influence of the chosen neutral target term, we repeat the experiments on the target terms ``actor'', ``adult'', ``human'', ``person'' and ``child''.
Each experiment is repeated ten times, and we report the mean and standard deviation. Tables with exact values of our experiments and the number of parameters for the models are also available in \cref{app:additional_clip_results}.

\paragraph{Experimental Results}
A summary of our results of the experiments on the ViT-B/32 model can be seen in \cref{fig:human_with_wl_vitb32_metrics,fig:human_vitb32_imagenet}. 
Evidently, after unlearning using backdoors, the text encoder successfully maps the names of individuals to the term ``human'', which causes the IDIA to fail.
As can be seen in \cref{fig:human_with_wl_vitb32_metrics}, the mean true positive rate (TPR) of the IDIA is zero, while the values for all other metrics are greater than $0.99$, independent of how many names were removed. 
The high backdoor similarity $\mathit{Sim}_\mathit{Backdoor}$ between the prompts containing the trigger and prompts containing the neutral word confirms that the backdoors indeed map to the target embeddings.
The text encoder and, as a result, the whole CLIP model only decreased negligibly in its utility. 
As a result, the clean similarity $\mathit{Sim}_{Clean}$, which calculates the similarity of prompts without the trigger on the clean and backdoored text encoder, remains very high. 
Even when 64 names are removed from the model at once, the clean similarity stays above $0.99$. 
The preservation of the performance can also be seen when looking at the zero-shot top-1 and top-5 accuracy on ImageNet in \cref{fig:human_vitb32_imagenet}. 
Using no weight regularization during fine-tuning results in a slightly higher decrease in utility.
Even though this effect is only small for the text encoder, this result shows that performing weight regularization does indeed help to retain the utility of the model.
Even though we have removed 64 names from the model, the average top-1 and top-5 accuracy declines by only $0.05$ and $0.12$ percentage points, respectively, when using no weight regularization.
In contrast, for the models without regularization, the mean top-1 and top-5 accuracy decreases by $0.68$ and $0.52$ percentage points.
In addition to calculating $\mathit{Sim}_\mathit{Clean}$, $\mathit{Sim}_{Targets}$ is calculating the similarity of the target phrase embeddings of the original and backdoored models. 
As can be seen, the embeddings of the target phrases are not altered when backdooring the model, underscoring the high utility of the model.

To investigate the influence of the target term on the effectiveness of the defense, we performed the same experiment with the targets ``actor'', ``adult'', ``human'', ``person'' and even ``child'', even though we are only removing adult individuals. The results can be seen in \cref{fig:text_enc_different_targets}. The choice of the target term does not influence the performance of the defense. Since the individuals chosen for unlearning are from the FaceScrub dataset, they are all adults. Even when choosing the unrelated target term ``child'', the defense is successful. This underlines the versatility and applicability of our approach.

Additional results with the text encoder of the ViT-L/14 model, the results of the experiments without weight regularization, and a performance evaluation on other data sets can be found in \cref{app:additional_clip_results}.

\begin{figure*}[t]
    \centering
    \begin{subfigure}[t]{0.30\textwidth}
        \includegraphics[width=\linewidth,trim={0 1cm 0 1.4cm},clip]{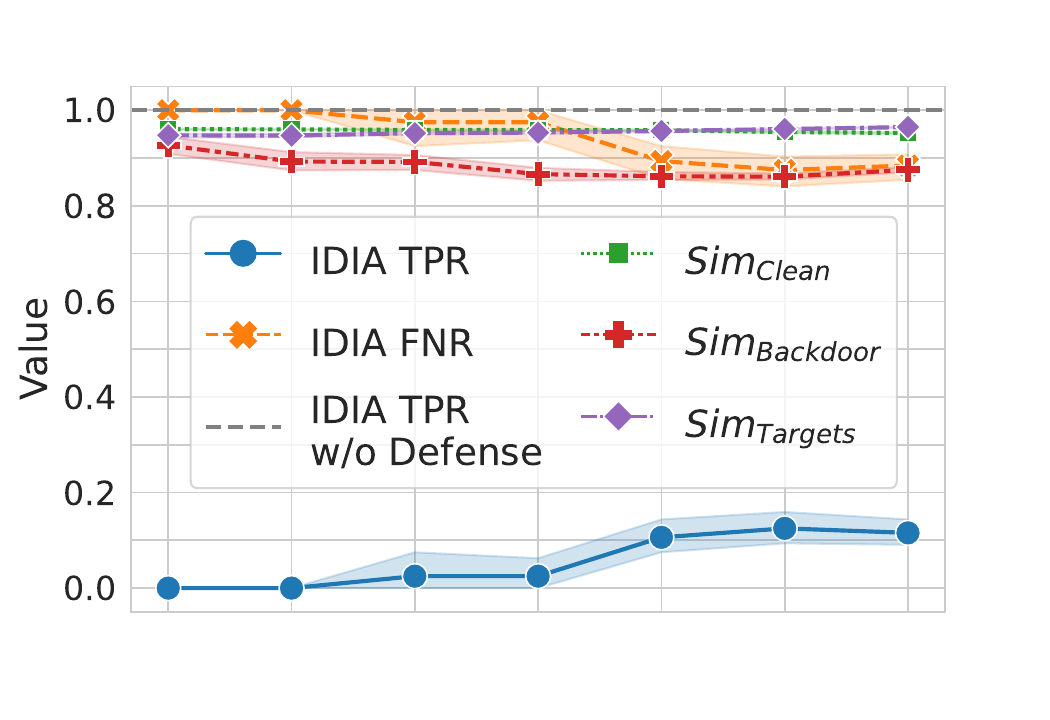}
        \vspace{-0.8cm}
        \caption{IDIA results when applied to the defended image encoder}
        \label{fig:image_encoder_vitb32_metrics}
        \vspace{3ex}
    \end{subfigure}
    \hspace{1cm}
    \begin{subfigure}[t]{0.30\textwidth}
        \includegraphics[width=\linewidth,trim={0 1cm 0 1.4cm},clip]{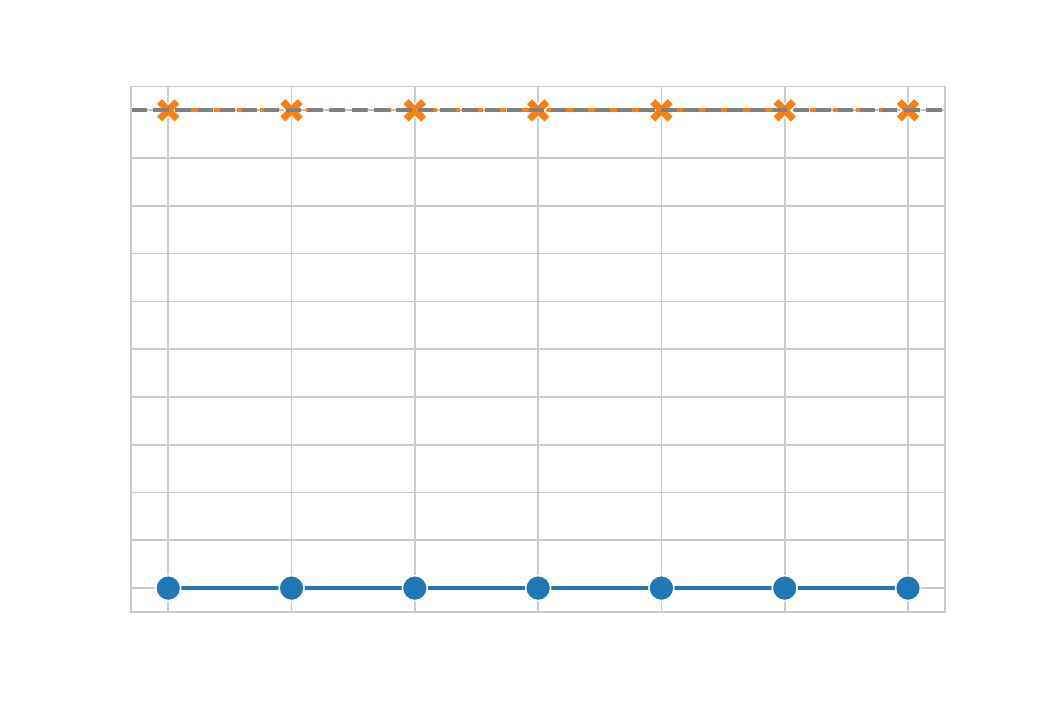}
        \vspace{-0.8cm}
        \caption{IDIA results when applied to the defended text- \textbf{and} image-encoder}
        \label{fig:merge_vitb32_metrics}
        \vspace{3ex}
    \end{subfigure}\\
    \begin{subfigure}[t]{0.30\textwidth}
        \includegraphics[width=\linewidth,trim={0 0 0 1.4cm},clip]{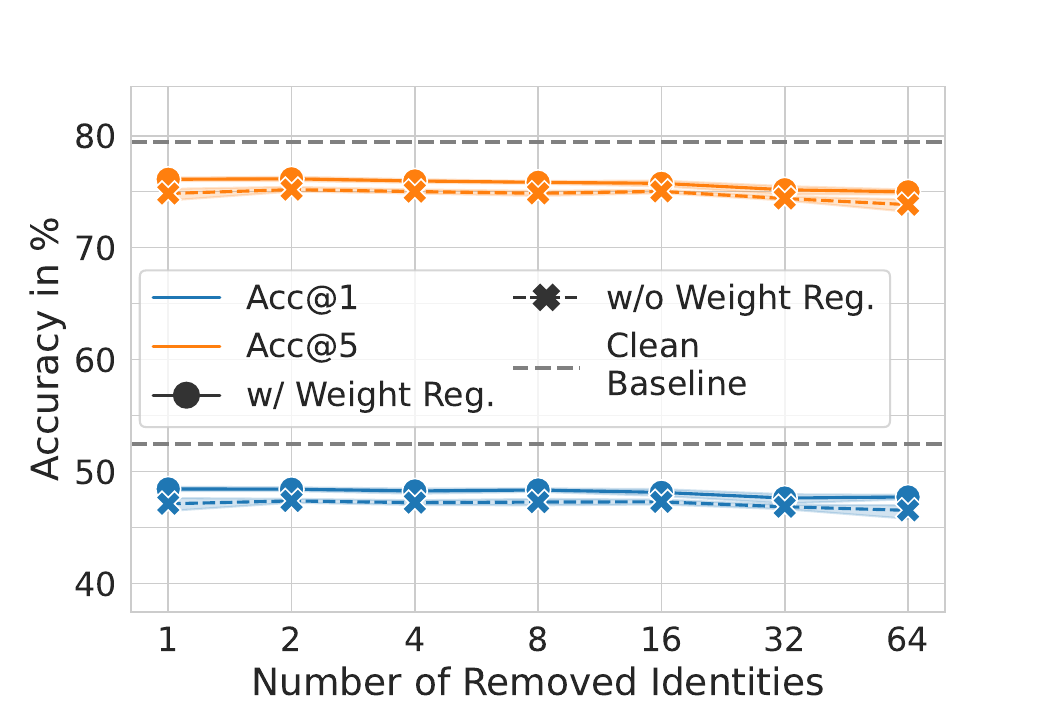}
        \vspace{-0.6cm}
        \caption{ImageNet zero-shot acc. of the fine-tuned image encoder used in CLIP. 
        }
        \label{fig:image_encoder_vitb32_imagenet}
    \end{subfigure}
    \hspace{1cm}
    \begin{subfigure}[t]{0.30\textwidth}
        \includegraphics[width=\linewidth,trim={0 0 0 1.4cm},clip]{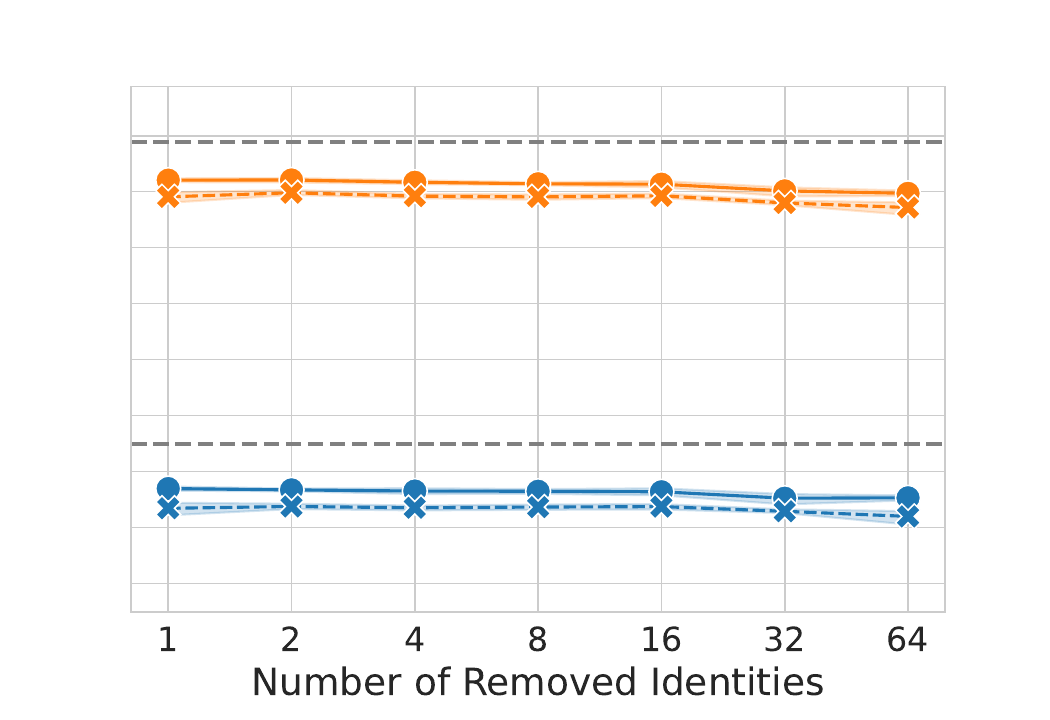}
        \vspace{-0.6cm}
        \caption{ImageNet zero-shot acc. of the defended text- \textbf{and} image-encoder used in CLIP.
        }
        \label{fig:merge_vitb32_imagenet}
        \vspace{2.5ex}
    \end{subfigure}
    \caption{Using backdoors successfully removes faces of individuals from the image encoder of the ViT-B/32 CLIP model while maintaining its utility. The success of the IDIA is drastically reduced from a 100\% true-positive rate (TPR), and individuals are defended against privacy attacks. In comparison to defending the text encoder, unlearning the faces of multiple identities at the same time seems to be harder. However, weight regularization seems to successfully mitigate the decrease in performance.}
    \label{fig:exp_res}
    \vspace{2ex}
\end{figure*}

\section{Experimental Evaluation: Teaching CLIP to Forget Faces}
Even after removing the name of a person from the text encoder, it might still be possible to extract information from the image encoder. Therefore, we apply our proposed defense also to the image encoder of the CLIP model.

\paragraph{Experimental Setting} 
As with the experiments on the text encoders, we select individuals to remove from the model using the FaceScrub dataset.
To evaluate our defense with different architectures and number of parameters, we apply our approach to the OpenCLIP models using the ViT-B/32 and ViT-B/16 vision transformers and the OpenAI ResNet-50 CLIP model. 
Similar to the defense for the text encoders, we are using randomly sampled images of the MS-COCO training dataset~\cite{ms_coco} for injecting the backdoor into the image encoder and to unlearn individuals for which the IDIA is correctly predicting them to be in the training data.
To perform the defense on the image encoder, we are using the faces of the individuals as the backdoor trigger. This will remap the image embeddings with the individuals to be unlearned to the target embedding and, in turn, prevent the model from leaking sensitive information.
To create images containing the backdoor trigger, we  randomly sample batches from the MS-COCO training set and add augmented faces of individuals to be unlearned to these images at random positions.
These sampled images of the MS-COCO dataset are diverse in their content and do not necessarily contain people.
The result of this procedure is that the model learns to map the faces of the individuals to the neutral target embedding if the face of the person is present in the image, regardless of the other content on the images. 

For the experimental evaluation, we calculate the average embedding of all individuals of the FaceScrub data set and use it as the neutral, anonymous target embedding $\Delta$, as seen in \cref{equ:backdoor_loss}.
We use $10,000$ randomly sampled images from the MS-COCO evaluation set~\cite{ms_coco} to calculate the similarity metrics $\mathit{Sim}_{Backdoor}$, $\mathit{Sim}_{Clean}$ and $\mathit{Sim}_{Targets}$.
We want to emphasize here that for evaluation, we do not use the same images of a person as used for incorporating the backdoor. 
By using different images of a person, we make sure that the encoder does not overfit to specific facial images of a person and instead is generalizing to unlearn the face of this person.

Because we expect an even stronger defense when fine-tuning both the image and text encoder of a single CLIP model, we are also evaluating a CLIP model where our defense was applied to both the image and the text encoder.

As with the text encoder, each experiment is repeated ten times, and we report the mean and standard deviation. Tables with exact values of our experiments and the number of parameters for the models are also available in \cref{app:additional_clip_results}.

\begin{figure*}[t]
    \centering
   \begin{subfigure}[t]{0.10\textwidth}
        \includegraphics[width=\linewidth]{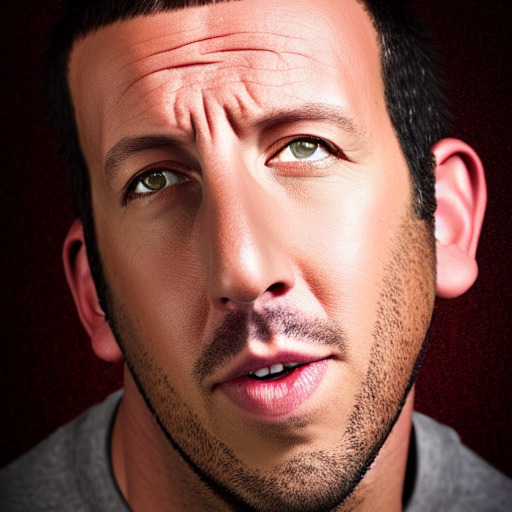}
    \end{subfigure}
    \begin{subfigure}[t]{0.10\textwidth}
        \includegraphics[width=\linewidth]{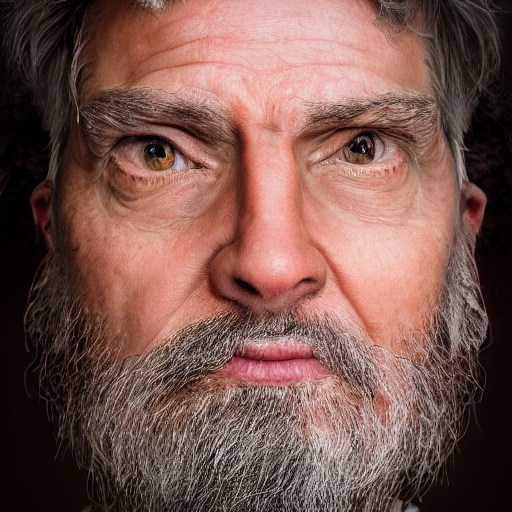}
    \end{subfigure}
    \begin{subfigure}[t]{0.10\textwidth}
        \includegraphics[width=\linewidth]{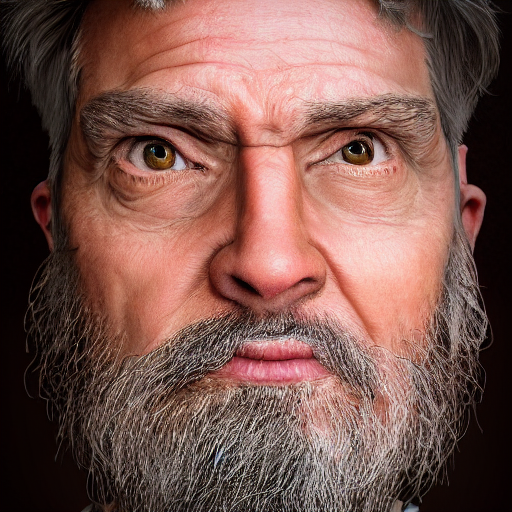}
    \end{subfigure}
    \hspace{2cm}
    \begin{subfigure}[t]{0.10\textwidth}
        \includegraphics[width=\linewidth]{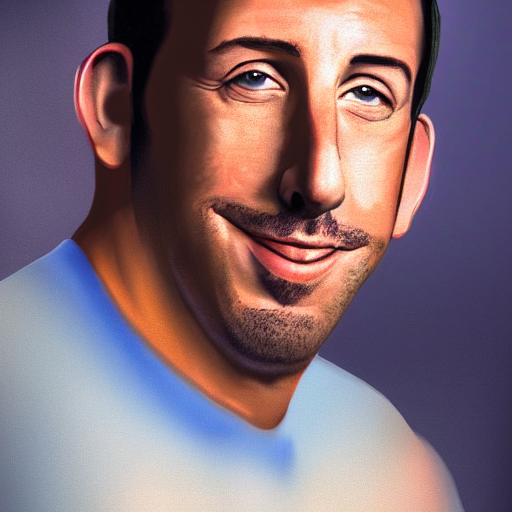}
    \end{subfigure}
    \begin{subfigure}[t]{0.10\textwidth}
        \includegraphics[width=\linewidth]{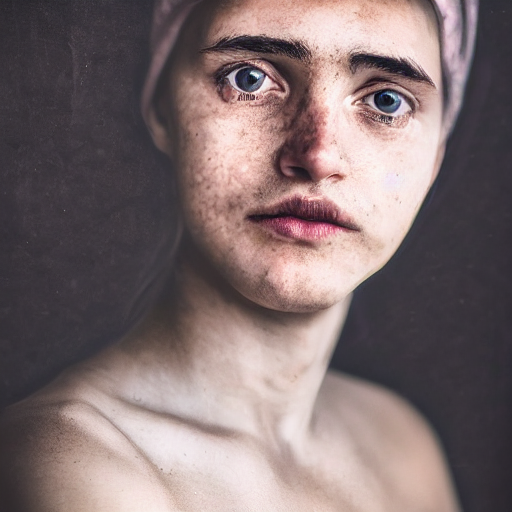}
    \end{subfigure}
    \begin{subfigure}[t]{0.10\textwidth}
        \includegraphics[width=\linewidth]{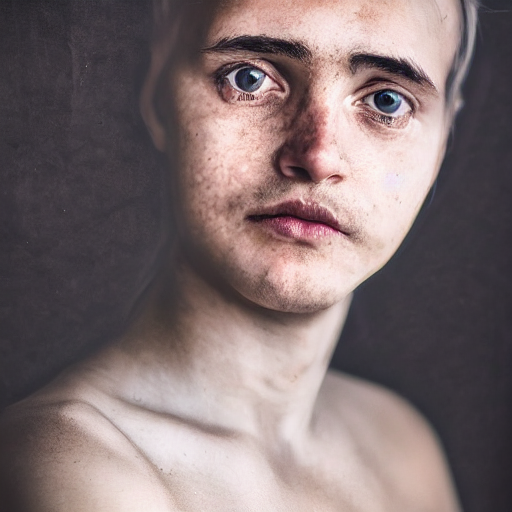}
    \end{subfigure}
    \vspace{1ex}
    \caption{Applying our defense to the text encoder of Stable Diffusion, we can remove \textbf{Adam Sandler} from the model. Two examples with the original image generated using the original Stable Diffusion model with the prompt containing the name (left), the image generated using the defended model (middle), and the image generated with the original model with the prompt containing ``person'' instead of the name (right). The exact prompt used for generating the images and additional examples can be found in \cref{app:sd_experiments}.}
    \label{fig:sd_examples}
    \vspace{2ex}
\end{figure*}

\paragraph{Experimental Results}
A summary of our results of the experiments on the ViT-B/32 image encoder can be seen in \cref{fig:image_encoder_vitb32_metrics,fig:image_encoder_vitb32_imagenet}. 
After applying our backdoor-based defense method, the image encoder successfully maps the faces of individuals to the average face embedding. 
As seen in \cref{fig:image_encoder_vitb32_metrics}, when unlearning a single and two identities at once, the TPR is at zero. This indicates that it is indeed possible to unlearn a person's face from the image encoder reliably. 
However, as suspected, the process of unlearning faces is apparently much more complex than unlearning names from a text encoder. 
While the TPR is always zero for the text encoders, the TPR of the IDIA for the defended image encoder is increasing with the number of unlearned faces.
We have two hypotheses for this phenomenon. 
One reason could be that re-mapping the face embeddings to the neutral, anonymous embedding is more complicated than with the text encoder. 
While the name of a person is always written the same, the faces of individuals can have different orientations and lighting conditions and might change over the years. 
Therefore, the model has to learn a backdoor that is invariant to these influences and maps all these different facial embeddings onto the same target embedding, which is a much harder task. 
A second possible reason could be that human faces are all mapped to the same subspace within the embedding space.
As a result, the face embeddings are much closer to each other than the embeddings of names for the text encoder. 
Therefore, it is much harder to disentangle the face embeddings such that the face embeddings of the individuals we want to unlearn are mapped to the target embedding while at the same time not mapping the faces of different individuals to the target embedding.
Experiments supporting our hypothesis, showing that the distribution of face embeddings is much denser than the distribution of text embeddings, can be found in \cref{app:additional_clip_results}.
This drastically increases the complexity of the task, which is a possible explanation for the higher TPR in the experiments on the image encoder.
We believe that due to both of these reasons, the accuracy drop of the image encoder in the ImageNet experiments after applying our defense is higher than that for the text encoder. However, for the image encoder, weight regularization seems to mitigate this drop more than for the text encoder.
When 64 individuals are unlearned from the image encoder without weight regularization, the average top-1 and top-5 ImageNet zero-shot accuracy drops by $5.92$ and $5.56$ percentage points, respectively, compared to the baselines.
In contrast, for the models with regularization, the mean top-1 and top-5 accuracy decreases by only $4.73$ and $4.42$ percentage points
At the same time, weight regularization seems to have a negative effect on the success of the defense. Using weight regularization decreases the FNR by $1.68$ percentage points on average for the ViT-B/32 model.
The results for the other architectures and a performance evaluation on other data sets can be found in \cref{app:additional_clip_results}.

As can be seen in~\cref{fig:merge_vitb32_metrics,fig:merge_vitb32_imagenet}, injecting a backdoor for each individual into both the image- and text encoder appears to be more effective in unlearning information from the model.
For all the numbers of identities removed at once, the TPR of the IDIA is zero. This shows that if unlearning an identity did not work on one of the encoders, the other encoder can compensate for that. 
However, the encoder with the lower utility--in this case the image encoder--seems to be the upper limit for the combined CLIP model's utility.

\section{Discussion, Limitations, and Future Work}\label{sec:discussion}
\paragraph{Discussion}
One could imagine that defending text encoders against privacy attacks, such as the IDIA, could be as straightforward as filtering out names, e.g., by using regular expressions.
However, the problem with the filtering approach is that the list of names to be removed from the model needs to be distributed together with the model. 
This is especially critical, as this list itself leaks private information. 

One of the main advantages of our proposed defense is privacy preservation even in downstream tasks, and therefore being able to apply our defense to models already used in production. 
Models like CLIP are often used in many downstream tasks, such as text-guided image generation or image captioning models. 
With our approach, the rest of the system does not need to be re-trained or fine-tuned after applying our backdoor-based defense since the utility of the CLIP model is retained, and the defended model behaves nearly identically to the original model.
Visual examples of our approach applied to the text encoder of Stable Diffusion 1.4 can be seen in \cref{fig:sd_examples}. 
As can be seen, the original model clearly leaks the visual appearance of the actor ``Adam Sandler''. However, our defended model does behave the same as the original model when prompted with the neutral term. 

One of the main advantages of our approach, in contrast to existing unlearning approaches, is that no special textual or image datasets, especially not the original training sets, are required. For the text encoder, only the name of the person needs to be known, while for the image encoder, roughly 30 facial images of a person are already sufficient to remove the identity from the model.

\paragraph{Limitations and Future Work}
While we applied the proposed defense based on backdoors only on encoders, we believe that it is also possible to apply it to other models, such as classification models. With our approach, we force the model to perform a pre-defined mapping in the embedding space when presented with the trigger. Even for classification models, the backdoors introduce a change in the computed embeddings~\cite{spectral_signatures}, leading to pre-defined behavior and misclassifications. We believe that future work can adapt our approach to other models by performing the optimization in the embedding space of the penultimate layer. As neutral targets, the average embeddings of the respective classes could be used, similar to the average face embeddings in our experiments.

Some unlearning approaches in the literature provide formal guarantees, similar to differential privacy~\cite{guo_certified_unlearning,izzo_unlearning}. However, as shown by \citet{towards_unbounded}, these approaches perform poorly and do not scale well. 
While we cannot give formal guarantees, our approach can perform the unlearning in only a few minutes, scales well to even very large models, such as transformer models with even 85 million parameters, and is still successful.
However, future work that can give formal guarantees for scalable and efficient approaches like ours would be highly valuable.

While our approach can unlearn specific names on text encoders, we hypothesize that it is still possible to extract private information about individuals by using synonyms for their names when defending only the text encoder. 
This is because the backdoor trigger is only injected for a specific name and as a result, the remapping to a neutral embedding is not triggered when presented with a synonymous name for the same individual, like ``Terminator`` and ``Arnold Schwarzenegger``.
While we hypothesize that this problem does not persist when defending the image encoder, an interesting avenue for future work on text encoders could be to investigate whether it is possible to remap whole areas, e.g., an $\epsilon$-Ball around the term to unlearn, in the embedding space. 
This could allow unlearning synonyms, even though they are not directly used as the backdoor trigger.

\section{Conclusion}\label{sec:conclusion}
With large vision-language models trained on data scraped from the web, privacy is often neglected. 
Encoding private information such as names, addresses, and even faces, these models are getting more into the focus of privacy attacks.
Having personal data deleted from the model once it is trained is quite hard. 
We address this issue by showing that backdoors can be used to remove information about an individual from text and image encoders and, therefore, defend against privacy attacks.
Our backdoor-based defense maps the embeddings of specific phrases, names, or face images to the embeddings of neutral and anonymous embeddings.
Removing names and faces from the text- and image-encoder has only little impact on their performance, while at the same time, privacy attacks are prevented.
Even defending both the text- and image-encoder is possible, resulting in very strong privacy preservation.
We are the first to underscore the potential ``dual-use'' perspective of backdoors to remove information from models and to defend against privacy attacks. With our work, we hope to motivate future research to investigate this effective approach further.

\begin{ack}
This work was supported by the German Ministry of Education and Research (BMBF) within the framework program ``Research for Civil Security'' of the German Federal Government, project KISTRA (reference no. 13N15343), has been financially supported by Deutsche Forschungsgemeinschaft, DFG Project number 459419731, the German Research Center for Artificial Intelligence (DFKI) project “SAINT”, and the Research Center Trustworthy Data Science and Security (\url{https://rc-trust.ai}), one of the Research Alliance centers within the UA Ruhr (\url{https://uaruhr.de}).
\end{ack}

%%%%%%%%%%%%%%%%%%%%%%%%%%%%%%%%%%%%%%%%%%%%%%%%%%%%%%%%%%%%%%%%%%%%%%%%

%%% Use this environment to include acknowledgements (optional).
%%% This will be omitted in doubleblind mode.

% \begin{ack}
% By using the \texttt{ack} environment to insert your (optional) 
% acknowledgements, you can ensure that the text is suppressed whenever 
% you use the \texttt{doubleblind} option. In the final version, 
% acknowledgements may be included on the extra page intended for references.
% \end{ack}

%%%%%%%%%%%%%%%%%%%%%%%%%%%%%%%%%%%%%%%%%%%%%%%%%%%%%%%%%%%%%%%%%%%%%%%%

%%% Use this command to include your bibliography file.

\bibliography{main}

\appendix
\section{Experimental Details}\label{app:exp_details}
In this appendix, we state additional experimental details to reproduce our experiments. Our source code is available at 
\url{https://github.com/D0miH/Defending-Our-Privacy-With-Backdoors} for reproducibility purposes.

\subsection{Hard- and Software Details}
The experiments conducted in this work were run on NVIDIA DGX machines with NVIDIA DGX Server Version 5.1.0 and Ubuntu 20.04.4 LTS. The machines have NVIDIA A100-SXM4-40GB GPUs, AMD EPYC 7742 64-Core processors, and 1.9TB of RAM. The experiments were run with Python 3.10.9, CUDA 11.7, and PyTorch 2.0.0 with TorchVision 0.15.0.

\subsection{Hyperparameters}\label{app:hyperparameters}
For our experiments, the names and images of the individuals were at maximum 300 times present in the training dataset for the models trained on LAION-400M~\cite{idia}. 
We set the number of possible names that can be predicted by the model to 1000 names, which consisted of the names present in the FaceScrub dataset and randomly generated names. The names were generated using the most popular male and female first names in the US from 1880-2008~\cite{first_names_list} and we randomly combined them with the most frequent last names from 2010 in the US~\cite{last_names_list}. We used the same prompts for the IDIA as \citet{idia}, and for the attack on each individual we used 20 images.

We fine-tuned the text encoder for 400 steps and used the AdamW optimizer with a learning rate of $1e^{-4}$, which was multiplied by $0.5$ after 200 and 300 steps. We chose $\alpha=0.6$ and used a clean batch size of 128 with 128 samples containing backdoors added for all experiments. We chose $\beta=0.01$ for the ViT-B/32 text encoder and $\beta=0.005$ for the ViT-L/14 text encoder.
For fine-tuning the vision transformer image encoders, we trained them for 100 steps with a learning rate of $1e^{-4}$, which was multiplied by $0.1$ after 25 and 75 epochs. For the ResNet-50 image encoder, we used a learning rate of $1e^{-5}$. For all experiments on the image encoders, we chose $\alpha=0.8$. For the ViT models, we chose $\beta=0.005$, while for the ResNet-50 model, we used $\beta=0.01$. We used a clean batch size of 128 for fine-tuning with 128 additional samples containing a backdoor.
All experiments on the text- and image encoders were repeated 10 times with different seeds ranging from 0 to 9.

\subsection{Metrics}
In the following, we will describe in detail how the metrics $Sim_\mathit{Clean}$, $Sim_\mathit{Backdoor}$ and $Sim_\mathit{Target}$ are calculated.
The clean similarity $Sim_\mathit{Clean}$, which calculates the cosine similarity between the outputs of the original model $M$ and the fine-tuned model $\tilde{M}$ on data samples without a backdoor trigger, is calculated as
\begin{align}
    Sim_\mathit{Clean} = \frac{1}{|T|} \sum_{x \in T} d\left(M(x), \tilde{M}(x) \right)
\end{align}
, where $T$ contains generic data samples and $d$ is calculating the cosine similarity. In the experiments, $T$ were $10,000$ images and corresponding captions from the MS-COCO evaluation set, to calculate the metric for the image- and the text encoders, respectively.

The backdoor similarity $Sim_\mathit{Backdoor}$, which calculates the cosine similarity between the outputs of the fine-tuned model on data samples containing a backdoor trigger and the target embedding, is calculated as
\begin{align}
    Sim_\mathit{Backdoor} = \frac{1}{|Z|} \sum_{x \in Z} d\left(\tilde{M}(x), \Delta \right)
\end{align}
, where $Z$ contains data samples with the sensitive features to unlearn and $\Delta$ is the target embedding. For text encoders, we choose $\Delta = \tilde{M}(x \oplus n)$ with $\oplus$ being the operation of exchanging the name with the neutral term. For image encoders, we choose $\Delta$ as the average face embedding and choose images of the person we want to unlearn as $Z$.

The target similarity $Sim_\mathit{Target}$, which calculates the cosine similarity between the outputs of the original model $M$ and the fine-tuned model $\tilde{M}$ on the neutral targets, is calculated as
\begin{align}
    Sim_\mathit{Target} = \frac{1}{|Q|} \sum_{x \in Q} d\left(M(x), \tilde{M}(x) \right)
\end{align}
, where $Q$ contains data samples with the neutral target term. A high target similarity shows that the embeddings of the target terms and images are not significantly altered by fine-tuning the model and the model therefore retains its utility.

\section{Additional Experimental Results}\label{app:additional_clip_results}
\subsection{Text Encoder}
\begin{figure*}[ht]
    \centering
    \begin{subfigure}[t]{0.49\textwidth}
        \includegraphics[width=\linewidth,trim={0 1.5cm 0 1cm},clip]{./images/text_encoder/human_with_wl_vitb32_no_x_label_metrics.pdf}
        \caption{ViT-B/32 w/ weight regularization}
        \vspace{1.6ex}
    \end{subfigure}
    \hfill
    \begin{subfigure}[t]{0.49\textwidth}
        \includegraphics[width=\linewidth,trim={0 1.5cm 0 1cm},clip]{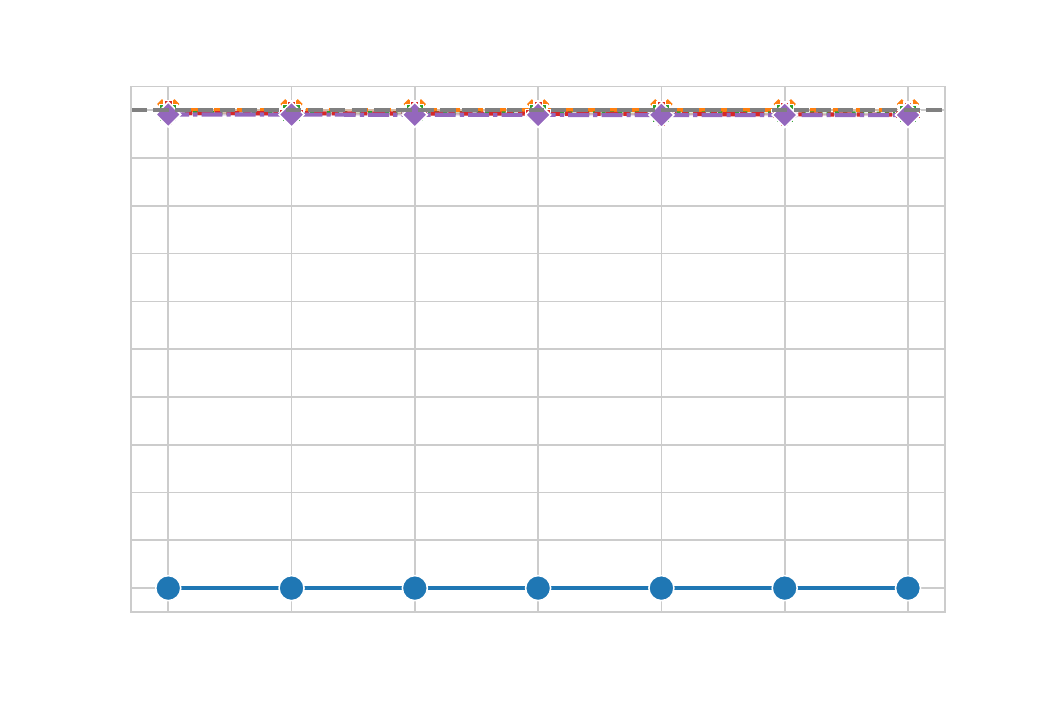}
        \caption{ViT-B/32 w/o weight regularization}
        \vspace{1.6ex}
    \end{subfigure}
    \begin{subfigure}[t]{0.49\textwidth}
        \includegraphics[width=\linewidth,trim={0 0.4cm 0 1cm},clip]{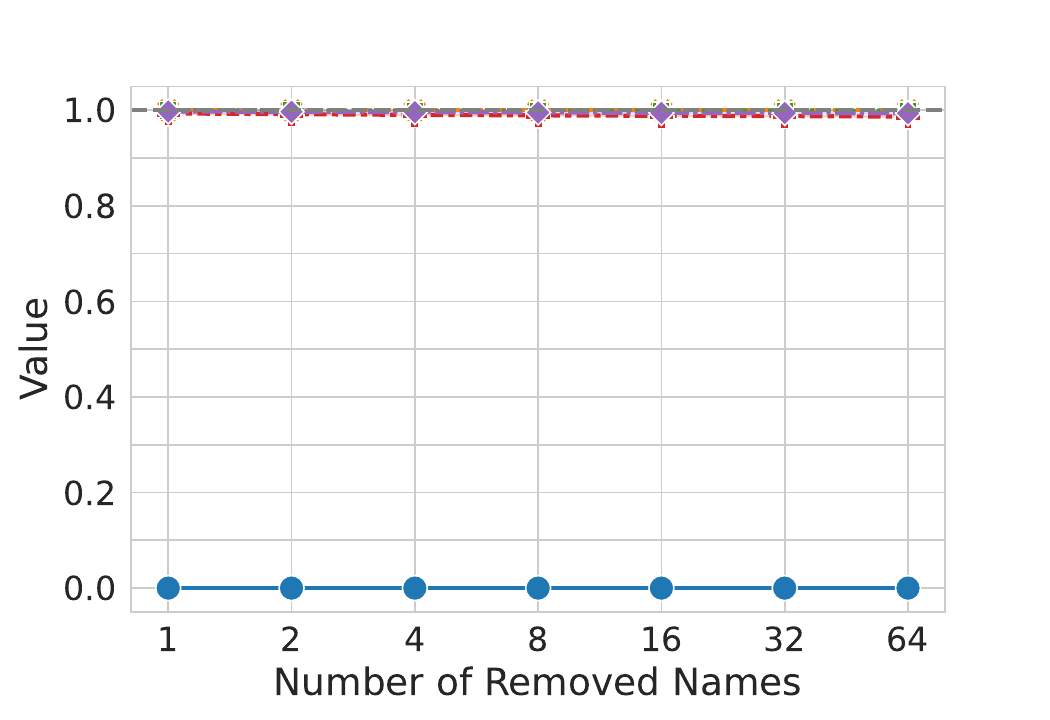}
        \caption{ViT-L/14 w/ weight regularization}
        \vspace{1.6ex}
    \end{subfigure}
    \hfill
    \begin{subfigure}[t]{0.49\textwidth}
        \includegraphics[width=\linewidth,trim={0 0.4cm 0 1cm},clip]{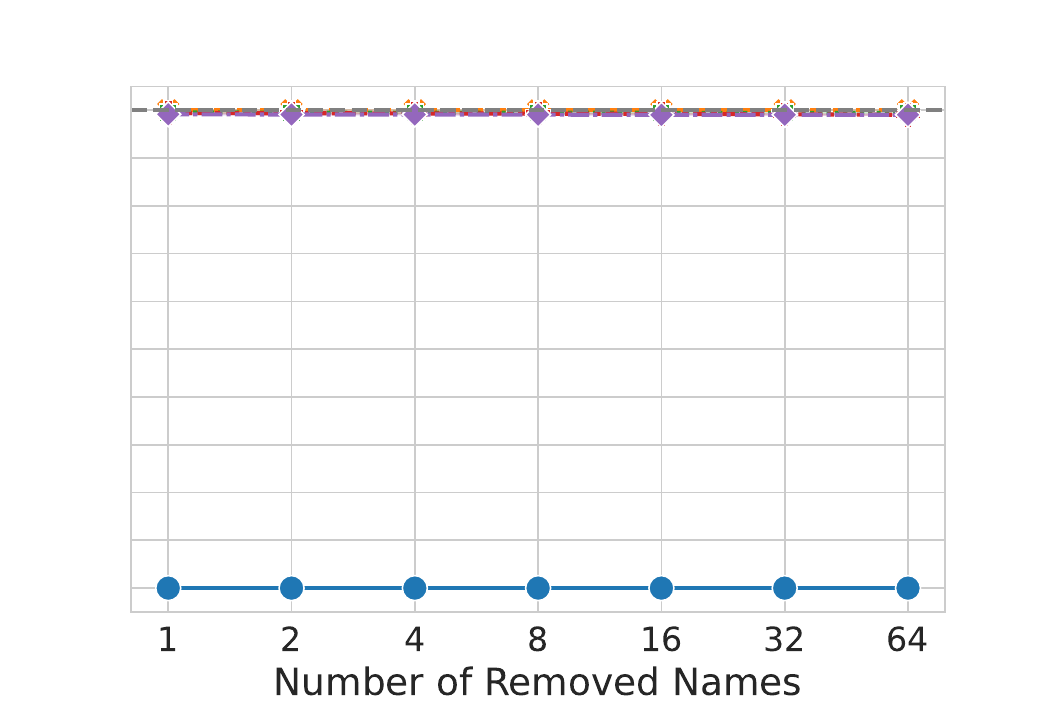}
        \caption{ViT-L/14 w/o weight regularization
        }
        \vspace{1.6ex}
    \end{subfigure}
    \vspace{3ex}
    \caption{Applying our defense with and without the weight regularization term to the text encoder.
    }
    \label{fig:add_results_text_encoder_metrics}
\end{figure*}

\begin{figure*}[!ht]
    \centering
    \begin{subfigure}[t]{0.49\textwidth}
        \includegraphics[width=\linewidth]{./images/text_encoder/human_vitb32_imagenet.pdf}
        \caption{ViT-B/32}
        \vspace{1.6ex}
    \end{subfigure}
    \begin{subfigure}[t]{0.49\textwidth}
        \includegraphics[width=\linewidth]{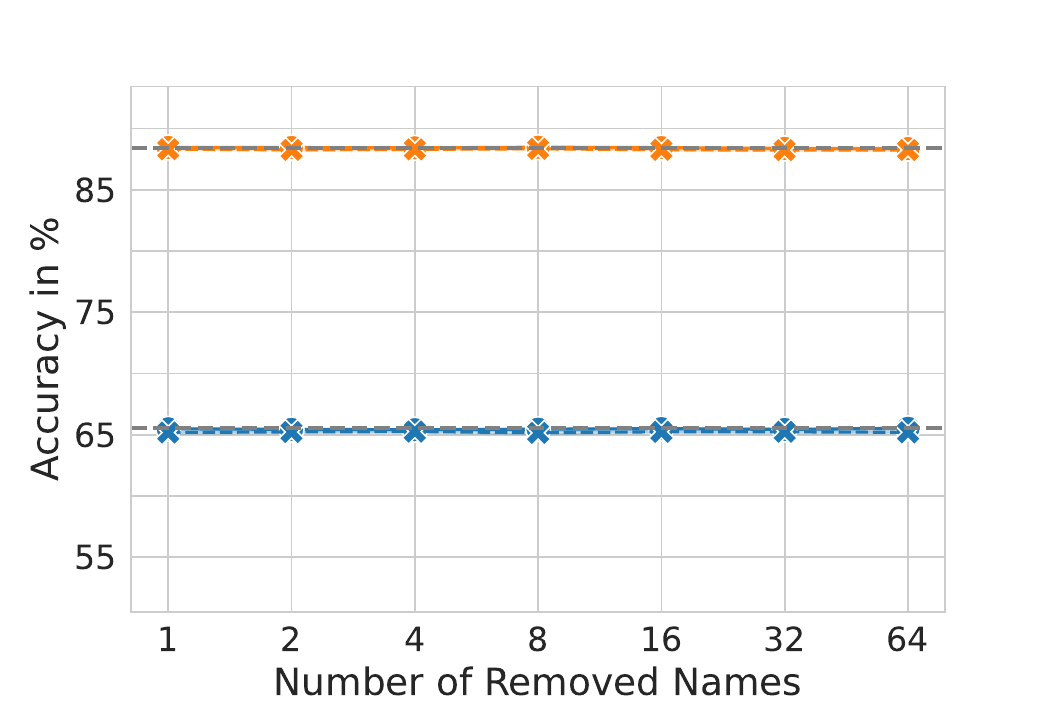}
        \caption{ViT-L/14}
        \vspace{1.6ex}
    \end{subfigure}
    \vspace{3ex}
    \caption{Top-1 and top-5 ImageNet accuracy of the text encoders after applying our defense.}
    \label{fig:add_results_text_encoder_imagenet}
\end{figure*}

The additional results for the experiments on the ViT-L/14 model and the results without weight regularization can be seen in \cref{fig:add_results_text_encoder_metrics,fig:add_results_text_encoder_imagenet}. The results look identical to the results on the ViT-B/32 model. The exact values of these experiments can be seen in \cref{tab:results_text_enc_vitb32,tab:results_text_enc_vitl14}. Using weight regularization improves the top-1 and top-5 ImageNet accuracy for both the ViT-B/32 and ViT-L/14 models, independent of how many names were removed at once from the model.
With the ViT-L/14 text encoder having more than twice the number of trainable parameters, one can see that the defense is still very effective and is defending the model as well as on the smaller ViT-B/32 text encoder. 
\begin{table}[ht]
    \centering
    \begin{tabular}{l|c}
        Model Name  & Text Encoder  \\ \hline
        ViT-B/32    & 37,828,608    \\ \hline
        ViT-L/14    & 85,054,464    \\
    \end{tabular}
    \caption{The number of trainable parameters of the text encoder for each of the models used in the experiments.}
    \label{app:num_params}
\end{table}
\begin{table*}[ht]
    \centering
    \begin{tabular}{c|c|c|c|c|c}
                                    & CIFAR-10~\cite{cifar10}   & CIFAR-100~\cite{cifar10}  & SUN397~\cite{sun_dataset} & Flowers-102~\cite{flowers102} & FaceScrub~\cite{facescrub}    \\ \hline
        \makecell{Original\\Model}  & $88.67\%\pm 0\%$          & $68.73\%\pm 0\%$      & $64.97\%\pm 0\%$              & $65.91\%\pm 0\%$              & $53.98\%\pm 0\%$              \\ \hline
        \makecell{Defended\\Model}  & $88.83\%\pm 0.06\%$       & $68.67\%\pm 0.08\%$   & $65.22\%\pm 0.04\%$           & $65.72\%\pm 0.25\%$           & $52.71\%\pm 0.39\%$ 
    \end{tabular}
    \caption{High zero-shot accuracy of the defended text encoder on several different datasets after removing 64 names. As can be seen, the model with the defended text encoder does retain its performance and has a zero-shot accuracy which only slightly deviates from the original model. The experiments were conducted with five different seeds and reported are the mean and standard deviation.}
    \vspace{3ex}
    \label{tab:zero_shot_acc_other_datasets_text_enc}
\end{table*}
\begin{figure}[ht]
    \centering
    \includegraphics[width=\linewidth]{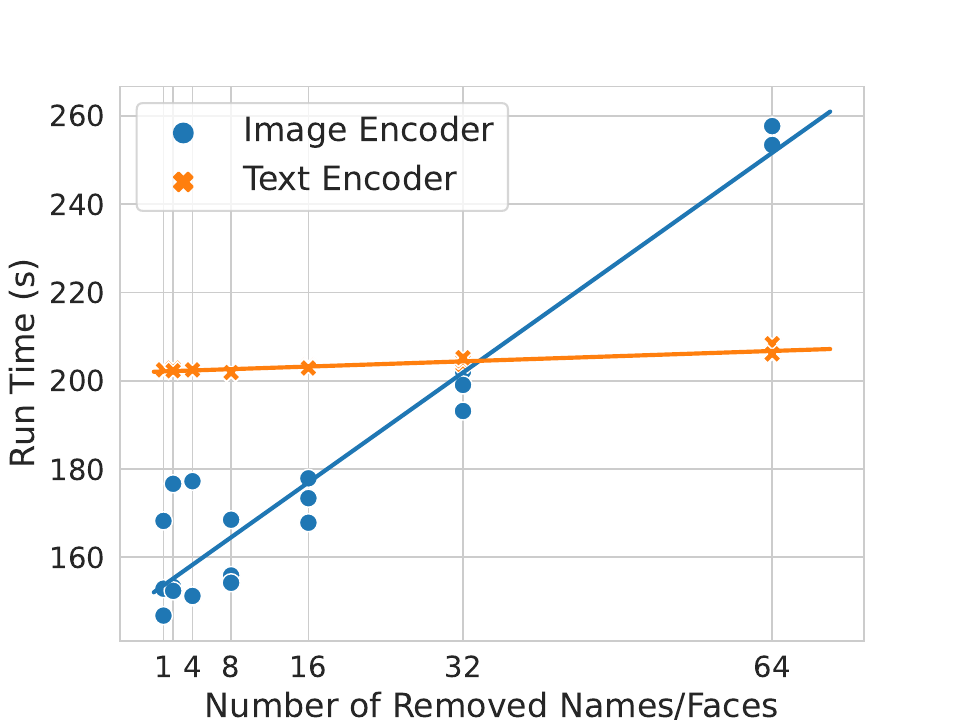}
    \caption{Run time analysis of the defense applied to the image- and the text-encoder. Plotted are the number of removed names/faces against the run time in seconds. Having measured the run time three times for each number of removed names and faces from the encoders, we applied a linear regression to approximate the time for each additional name/face that is unlearned. For the text encoder, each additional name adds approximately 0.07 seconds to the run time, while each additional unlearned face adds 1.55 seconds on average.}
    \vspace{5ex}
    \label{fig:run_times}
\end{figure}

\begin{figure*}[ht]
    \centering
    \begin{subfigure}[t]{0.49\textwidth}
        \includegraphics[width=\linewidth,trim={0 1.5cm 0 1cm},clip]{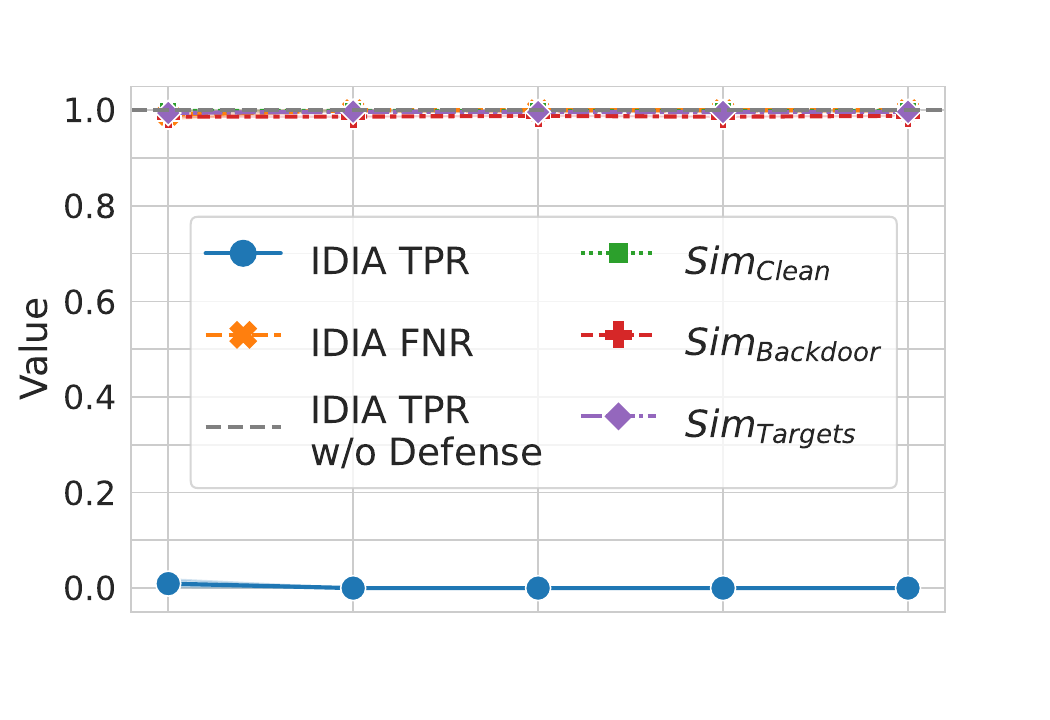}
        \caption{ViT-B/32 w/ weight regularization}
        \vspace{1.6ex}
    \end{subfigure}
    \hfill
    \begin{subfigure}[t]{0.49\textwidth}
        \includegraphics[width=\linewidth,trim={0 1.5cm 0 1cm},clip]{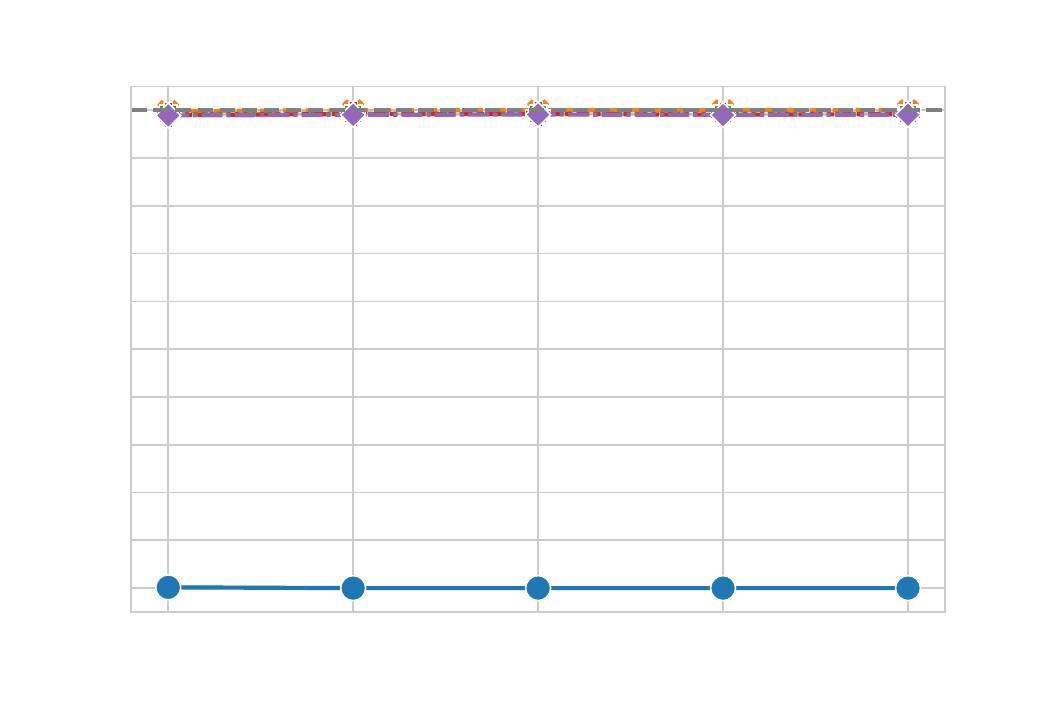}
        \caption{ViT-B/32 w/o weight regularization}
        \vspace{1.6ex}
    \end{subfigure}
    \begin{subfigure}[t]{0.49\textwidth}
        \includegraphics[width=\linewidth,trim={0 0.2cm 0 1cm},clip]{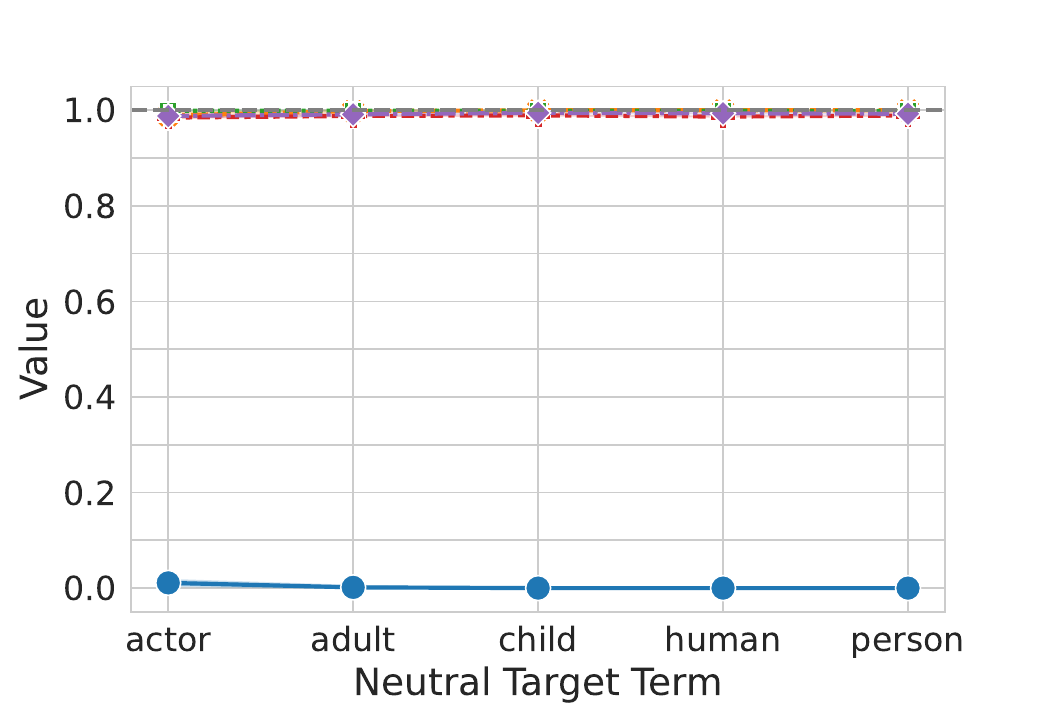}
        \caption{ViT-L/14 w/ weight regularization}
        \vspace{1.6ex}
    \end{subfigure}
    \hfill
    \begin{subfigure}[t]{0.49\textwidth}
        \includegraphics[width=\linewidth,trim={0 0.2cm 0 1cm},clip]{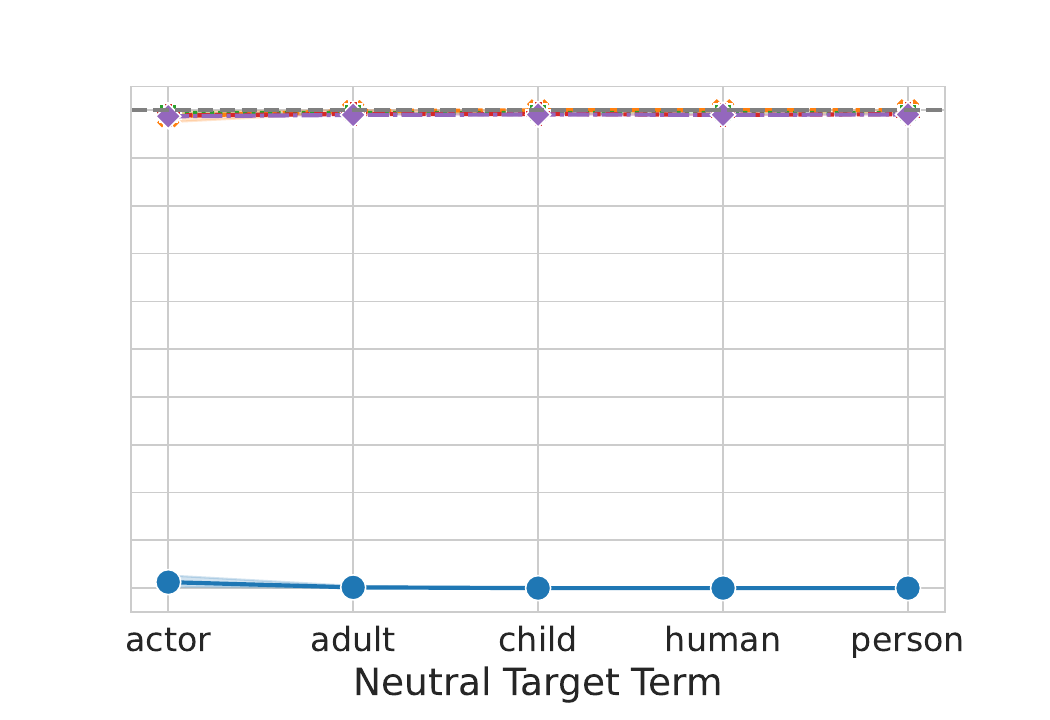}
        \caption{ViT-L/14 w/o weight regularization
        }
        \vspace{1.6ex}
    \end{subfigure}
    \vspace{3ex}
    \caption{Neutral target terms do not influence the defense performance. Applying our defense with and without weight regularization term to the text encoder with different target terms. In these experiments, 64 identities were removed at once.}
    \label{fig:add_results_text_encoder_targets_metrics}
\end{figure*}

\begin{figure*}[!ht]
    \centering
    \begin{subfigure}[t]{0.49\textwidth}
        \includegraphics[width=\linewidth]{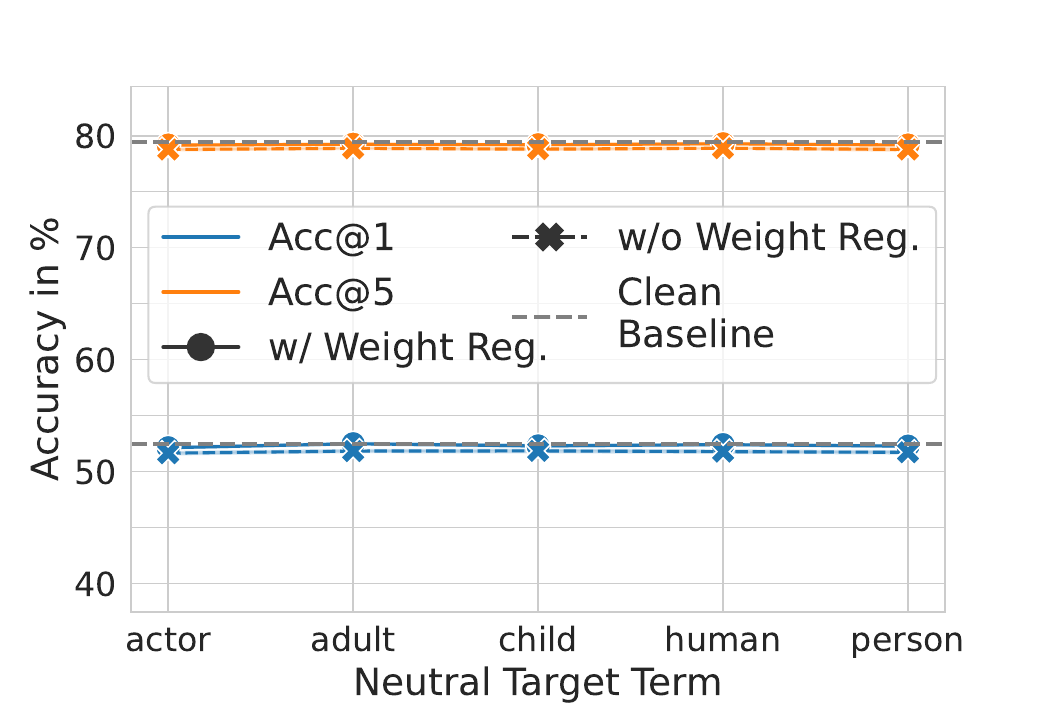}
        \caption{ViT-B/32}
        \vspace{1.6ex}
    \end{subfigure}
    \begin{subfigure}[t]{0.49\textwidth}
        \includegraphics[width=\linewidth]{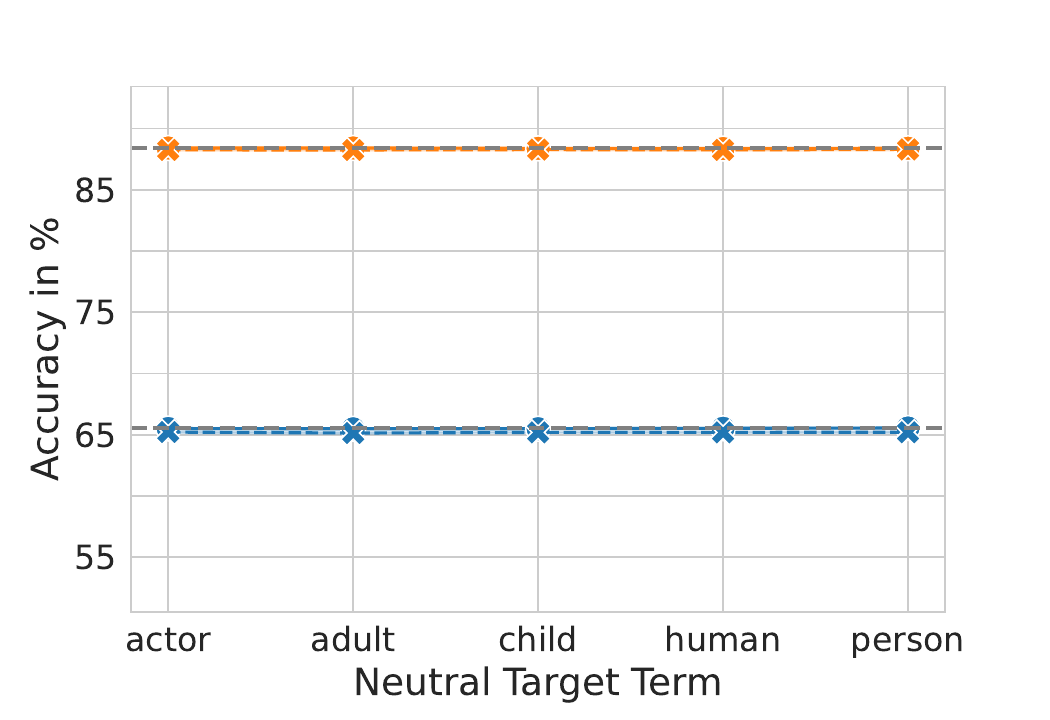}
        \caption{ViT-L/14}
        \vspace{1.6ex}
    \end{subfigure}
    \vspace{3ex}
    \caption{Neutral target terms do not influence the utility degradation of the defense. After applying the defense with different target terms, the ImageNet top-1 and top-5 zero-shot accuracy does not differ between these target terms. This shows that the choice of the neutral term does not influence the defense's performance.}
    \label{fig:add_results_text_encoder__targets_imagenet}
\end{figure*}

\newpage
To investigate whether the choice of the target term has an influence on the effectiveness of our defense we defended the text encoders with 5 different target terms. As can be seen in \cref{fig:add_results_text_encoder_targets_metrics,fig:add_results_text_encoder__targets_imagenet}, the choice of the neutral target term does not have an influence on the performance of the defense. Defending the models with different targets does not seem to have an impact on the performance of the defense, nor an impact on the utility of the fine-tuned model.

To showcase that the model retains its performance after applying our defense and removing 64 names from the text encoder, we measured the zero-shot accuracy on other, more diverse datasets. As can be seen in \cref{tab:zero_shot_acc_other_datasets_text_enc}, the zero-shot accuracy is not reduced and, in some cases, even improves after applying our defense. Even on the FaceScrub identity classification dataset, which is more similar to the unlearned concepts, the model demonstrates retained in-domain performance. These zero-shot experiments were run with five different seeds and reported are the mean and standard deviation.

We also have analyzed the run time of our defense. As can be seen in \cref{fig:run_times} there is a stark difference between the run time of text- and image-encoders. While for each additional unlearned name the run time increases by roughly 0.07 seconds, adding additional faces increases the run time by approximately 1.55 seconds per face. This is probably due to the higher number of batches required when unlearning faces. This experiment shows that our method can be applied in a few minutes and scales very well to unlearn larger amounts of data.
\vfill

\begin{table*}[ht]
    \centering
    \begin{tabular}{c|c|c|c|c|c}
                                    & CIFAR-10~\cite{cifar10}   & CIFAR-100~\cite{cifar10}  & SUN397~\cite{sun_dataset} & Flowers-102~\cite{flowers102} & FaceScrub~\cite{facescrub}    \\ \hline
        \makecell{Original\\Model}  & $88.67\%\pm 0\%$          & $68.73\%\pm 0\%$          & $64.97\%\pm 0\%$          & $65.91\%\pm 0\%$              & $53.98\%\pm 0\%$              \\ \hline
        \makecell{Defended\\Model}  & $74.93\%\pm 1.38\%$       & $43.99\%\pm 1.70\%$       & $61.84\%\pm 0.29\%$       & $59.77\%\pm 0.22\%$           & $0.24\%\pm 0.07\%$ 
    \end{tabular}
    \caption{High zero-shot accuracy of the defended image encoder on several different datasets after removing 64 faces. As can be seen, the model with the defended image encoder does slightly drop in performance. We suspect that this is due to the general higher drop in performance when defending the image encoder. The experiments were conducted with five different seeds and reported are the mean and standard deviation.}
    \vspace{3ex}
    \label{tab:zero_shot_acc_other_datasets_image_enc}
\end{table*}

\subsection{Image Encoders}\label{app:additional_clip_results_image_encoder}
\begin{figure}[H]
    \centering
    \includegraphics[width=0.7\linewidth]{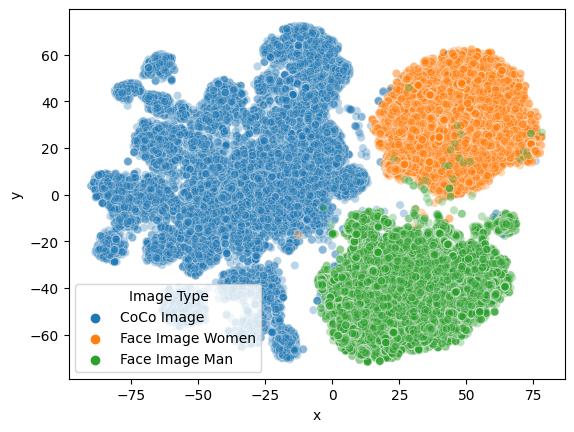}
    \caption{Faces are mapped to their own subspace within the embedding space. TSNE-plot of the MS-COCO and the FaceScrub embeddings calculated using the original ViT-B/32 image encoder. Images containing faces of male and female actors are clearly separable from each other and other images.}
    \label{fig:image_encoder_embeddings_plot}
\end{figure}
\begin{figure*}[ht]
    \centering
    \begin{subfigure}[t]{0.40\textwidth}
        \includegraphics[width=\linewidth,trim={0 1.5cm 0 1cm},clip]{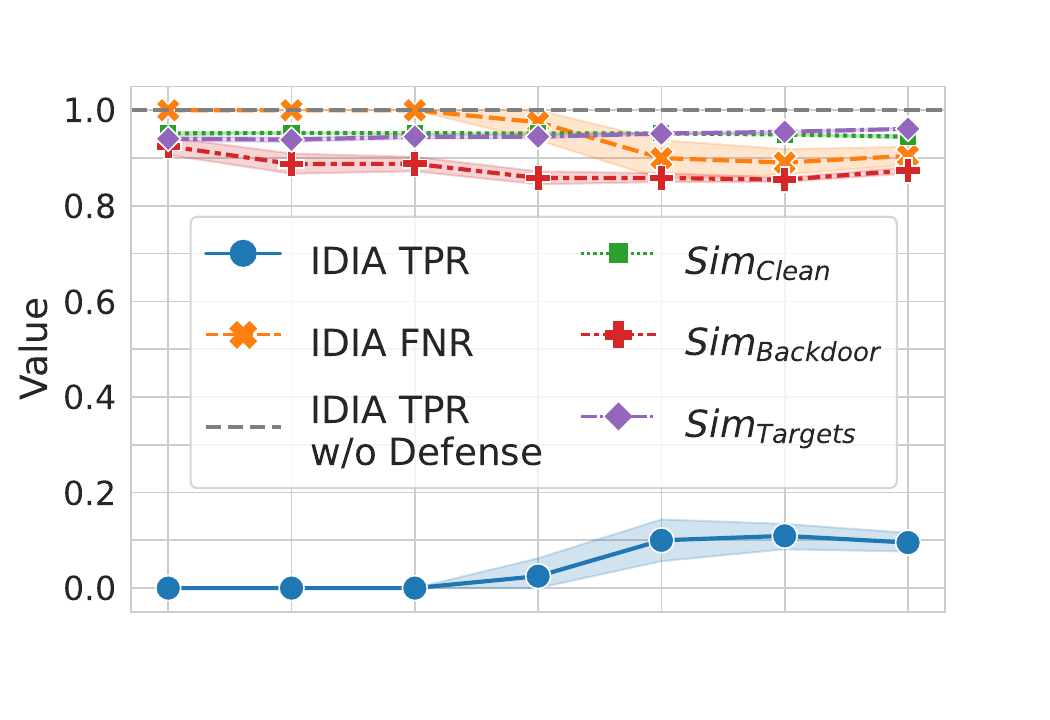}
        \caption{ViT-B/32 w/o weight regularization}
        \vspace{1.6ex}
    \end{subfigure}
    \hspace{0.5cm}
    \begin{subfigure}[t]{0.40\textwidth}
        \includegraphics[width=\linewidth,trim={0 1.5cm 0 1cm},clip]{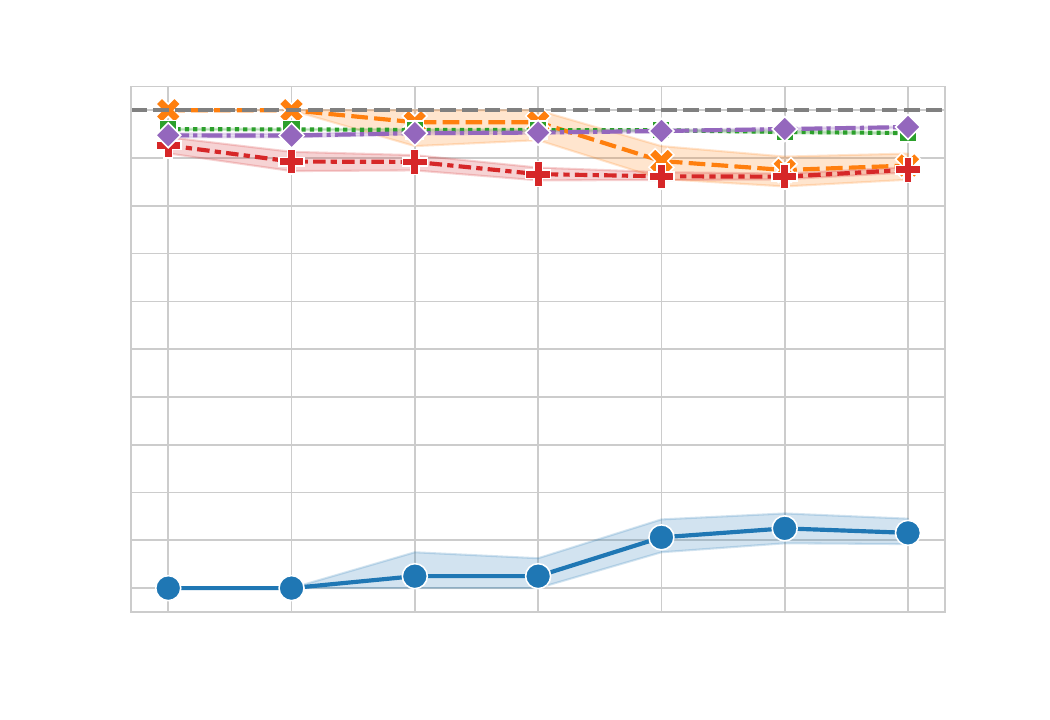}
        \caption{ViT-B/32 w/ weight regularization}
        \vspace{1.6ex}
    \end{subfigure}
    \begin{subfigure}[t]{0.40\textwidth}
        \includegraphics[width=\linewidth,trim={0 1.5cm 0 1cm},clip]{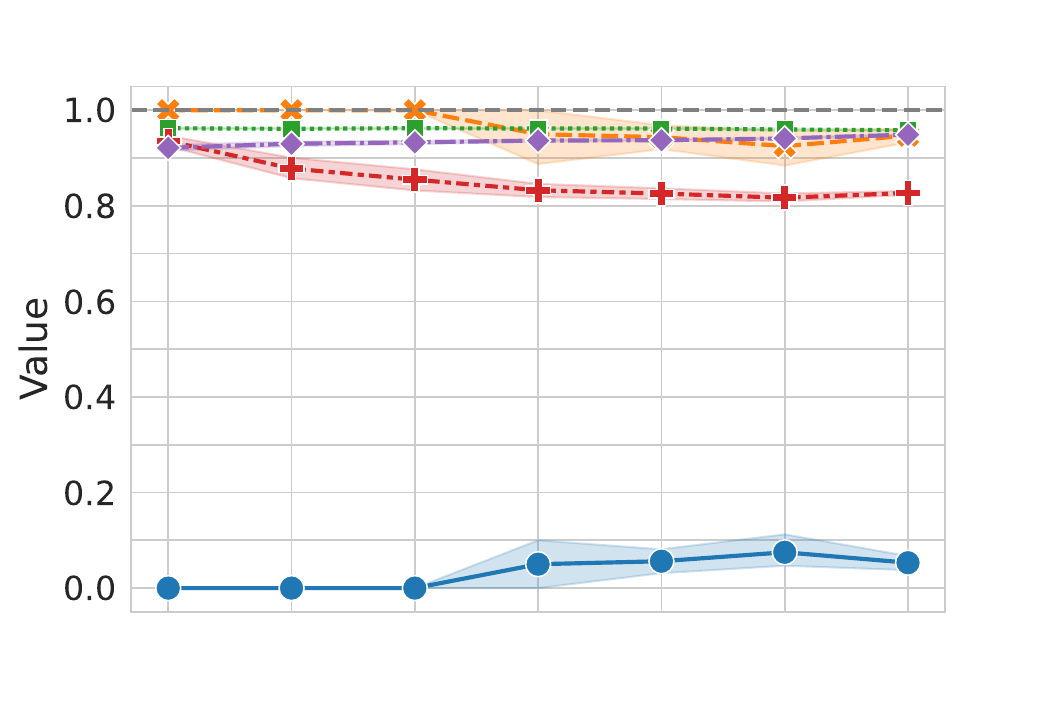}
        \caption{ViT-B/16 w/o weight regularization 
        }
        \vspace{1.6ex}
    \end{subfigure}
    \hspace{0.5cm}
    \begin{subfigure}[t]{0.40\textwidth}
        \includegraphics[width=\linewidth,trim={0 1.5cm 0 1cm},clip]{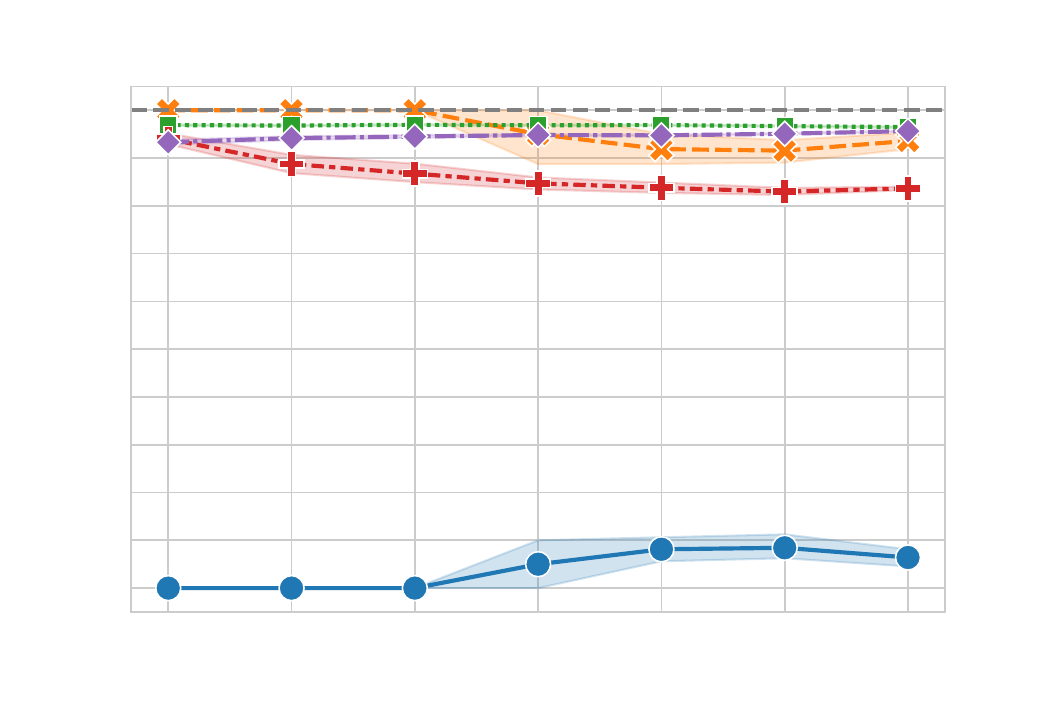}
        \caption{ViT-B/16 w/ weight regularization 
        }
        \vspace{1.6ex}
    \end{subfigure}
    \begin{subfigure}[t]{0.40\textwidth}
        \includegraphics[width=\linewidth,trim={0 0.4cm 0 1cm},clip]{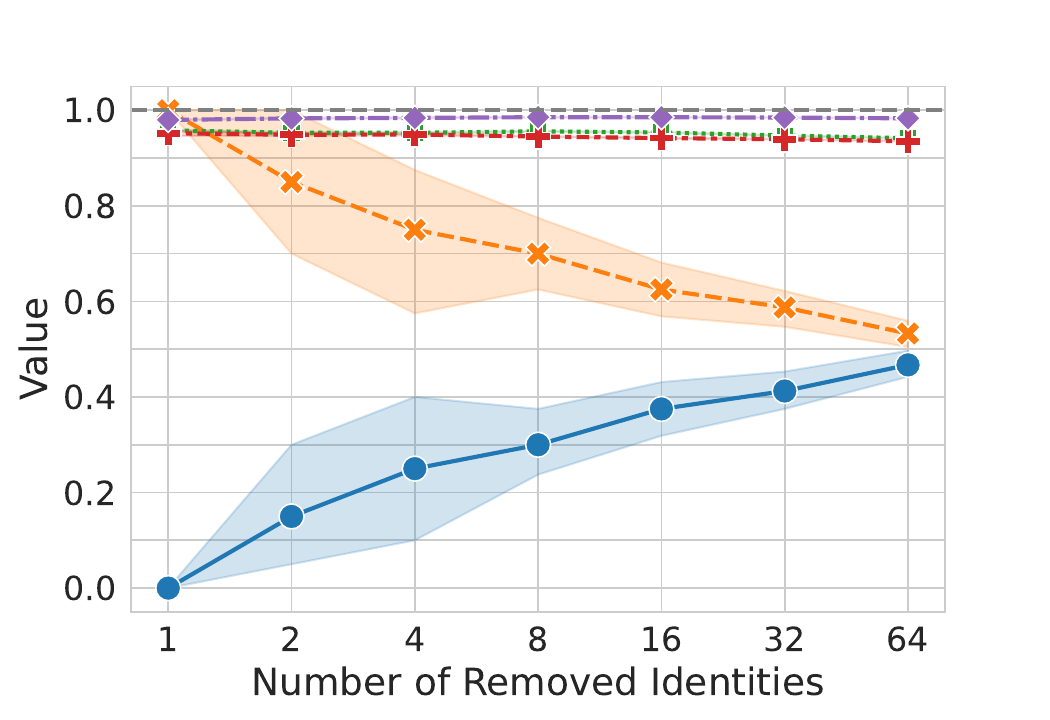}
        \caption{ResNet-50 w/o weight regularization}
        \vspace{1.6ex}
    \end{subfigure}
    \hspace{0.5cm}
    \begin{subfigure}[t]{0.40\textwidth}
        \includegraphics[width=\linewidth,trim={0 0.4cm 0 1cm},clip]{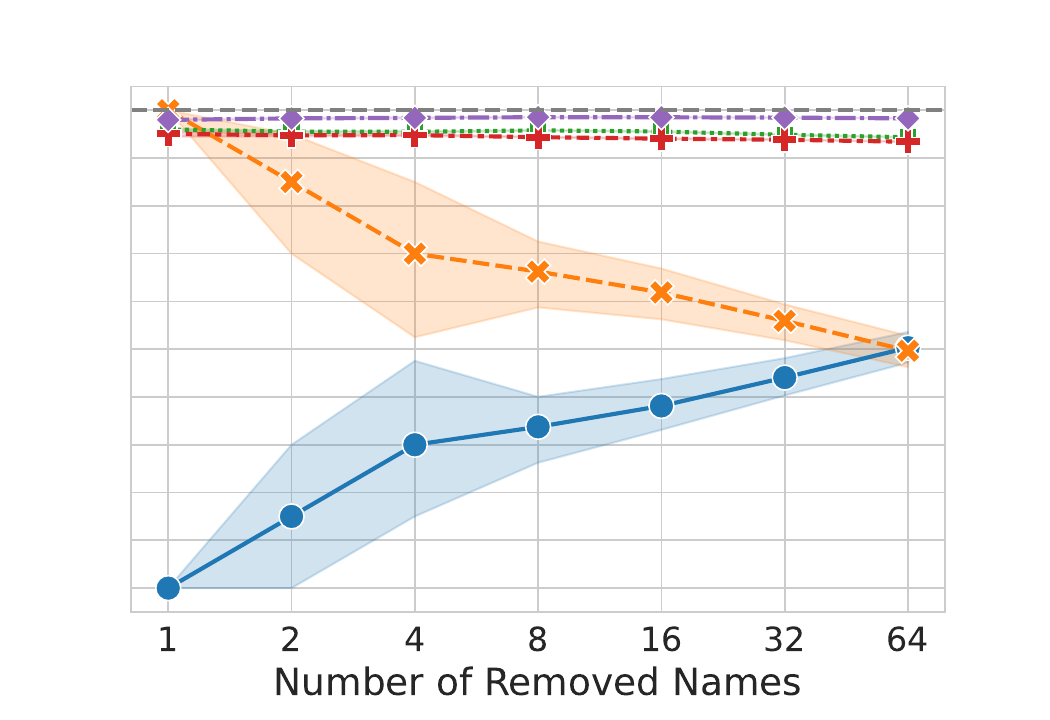}
        \caption{ResNet-50 w/ weight regularization
        }
        \vspace{1.6ex}
    \end{subfigure}
    \vspace{3ex}
    \caption{Applying our defense with and without the weight regularization term to the image encoder. Using no weight regularization, our defense removes the names even better than with the regularization. However, as discussed, at the same time, the utility of the model is reduced.}
    \label{fig:add_results_image_encoder_metrics}
\end{figure*}
\begin{figure*}[ht]
    \centering
    \begin{subfigure}[t]{0.32\textwidth}
        \includegraphics[width=\linewidth]{./images/image_encoder/coco/vitb32_full_legend_imagenet.pdf}
        \caption{ViT-B/32}
    \end{subfigure}
    \hfill
    \begin{subfigure}[t]{0.32\textwidth}
        \includegraphics[width=\linewidth]{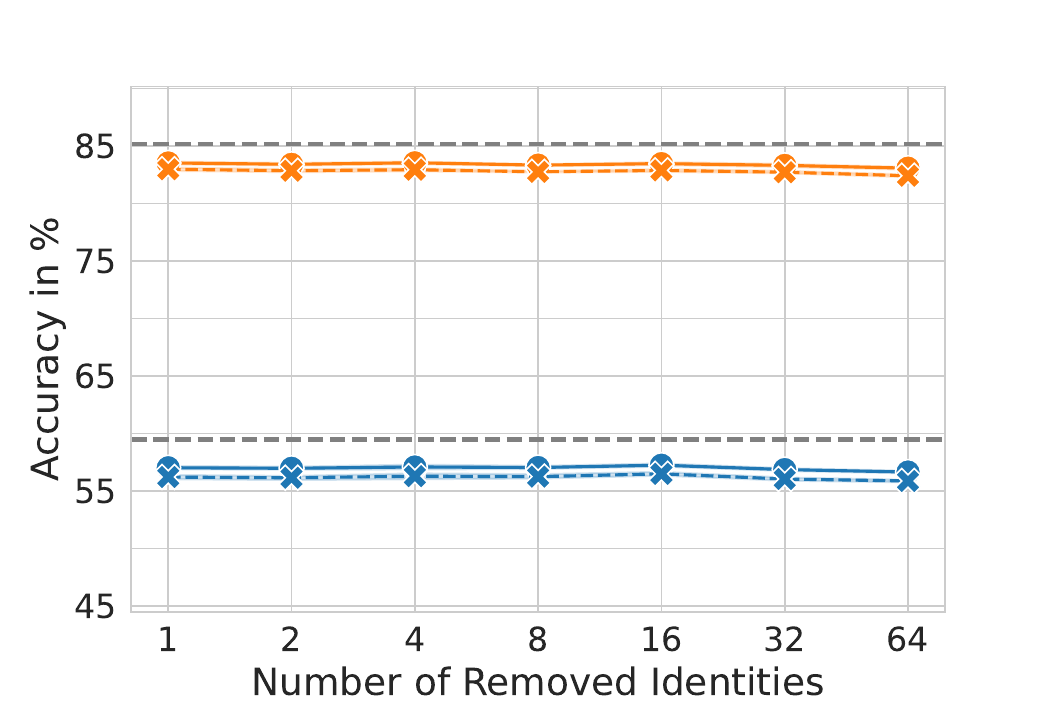}
        \caption{ViT-B/16}
    \end{subfigure}
    \hfill
    \begin{subfigure}[t]{0.32\textwidth}
        \includegraphics[width=\linewidth]{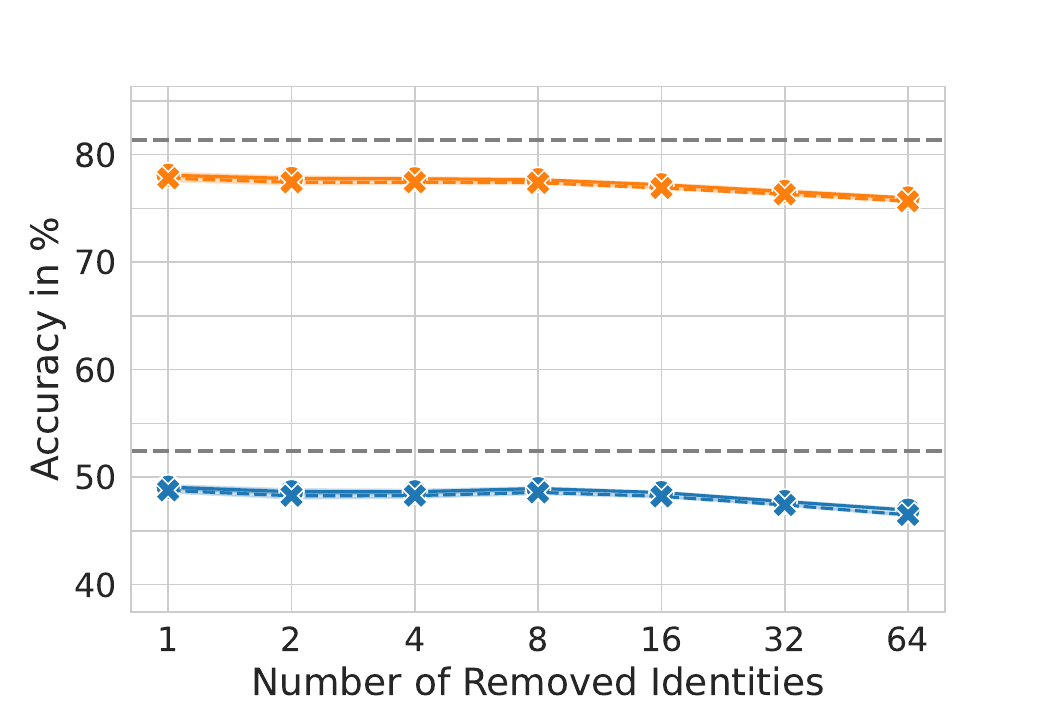}
        \caption{ResNet-50}
    \end{subfigure}
    \vspace{3ex}
    \caption{Top-1 and top-5 ImageNet zero-shot accuracy of the image encoders used in CLIP after applying our defense.}
    \label{fig:add_results_image_encoder_imagenet}
\end{figure*}

The additional results for the experiments on the ViT-B/16 and ResNet-50 image encoders and the results without weight regularization can be seen in \cref{fig:add_results_image_encoder_metrics,fig:add_results_image_encoder_imagenet}. The results on the ViT-B/16 model look very similar to the results on the ViT-B/32 model. 
Applying our defense on the ResNet-50 model seems to be a bit harder than on the vision transformers. It is still possible to reliably remove single identities from the model. However, when trying to remove more than one identity from the model at once, our defense is not as successful as on the vision transformers.
We suspect that this is due to the much lower number of trainable parameters of the ResNet-50 in comparison to the vision transformers, as can be seen in \cref{app:num_params_image_encoder}. 
Having a much lower number of trainable parameters, the ResNet-50 might not have enough capacity to reliably learn the backdoor, while at the same time not significantly degrade in utility. 
We suspect, however, that one way of solving this problem might be to use approaches using adapters~\cite{adapter}.
Comparing the defense applied without and with weight regularization, one can see that as with the text encoders, regularization helps to mitigate the performance degradation.
This can be seen in \cref{tab:results_image_enc_vitb32,tab:results_image_enc_vitb16,tab:results_image_enc_rn50}. Using weight regularization improves the top-1 and top-5 ImageNet accuracy for the ViT-B/32, ViT-B/16, and ResNet-50 models, independent of how many names were removed at once from the model. At the same time, the success of the defense is slightly reduced, indicated by the slightly higher IDIA true-positive rates. 
As a result, there seems to be a trade-off between defense success and preservation of the model utility for image encoders.

\begin{table}[H]
    \centering
    \begin{tabular}{l|c}
        Model Name  & Image Encoder \\ \hline
        ResNet-50   & 38,316,896      \\ \hline
        ViT-B/16    & 86,192,640      \\ \hline
        ViT-B/32    & 87,849,216      \\
    \end{tabular}
    \caption{The number of trainable parameters of the image encoder for each of the models used in the experiments.}
    \label{app:num_params_image_encoder}
\end{table}

\newpage
As seen in the tables and the plots, defending the image encoder is not as effective as defending the text encoder. Our hypothesis for this result is that face images are projected into a subspace within the image embedding space, making it harder to remap only some of them using backdoors. 
To test this hypothesis, we have calculated the embeddings of the MS-COCO test images and of all facial images in the FaceScrub dataset using the original ViT-B/32 image encoder. Looking at the TSNE-plot in \cref{fig:image_encoder_embeddings_plot}, one can see that the face images are clearly separable from all the MS-COCO images. This suggests that there is a subspace in the embedding space, specifically for facial images.
That could be why the defense is not as effective on image encoders as on text encoders because facial image embeddings are much closer to each other in the embedding space than the text embeddings of prompts containing the names to unlearn. 
This makes it much harder for the model to disentangle the facial embeddings and only remap those embeddings of individuals that should be unlearned while at the same time retaining the embeddings of individuals that are not unlearned. 
To further test this hypothesis, using the original ViT-B/32 model, we have calculated the average of the pairwise cosine similarity of all facial embeddings in the FaceScrub dataset and the pairwise cosine similarity of all text embeddings from the MS-COCO validation set with the names injected. 
While the average pairwise cosine similarity of the text embeddings is only $0.30$, the average pairwise similarity of the facial embeddings is $0.48$, much higher. This supports our hypothesis and suggests that the embeddings of facial images are indeed much closer to each other than the text embeddings containing the names of the individuals.
As a result, this makes it much harder to apply our defense to image encoders than to text encoders and underlines the importance and effectiveness of our approach on text encoders.

To evaluate the retained performance, we have measured the zero-shot accuracy of the CLIP model with the defended image encoder on more diverse data sets. For these experiments, we removed 64 faces from the image encoder and then measured the zero-shot performance. In \cref{tab:zero_shot_acc_other_datasets_image_enc} the zero-shot accuracy is given for the five data sets. As can be seen, the drop in performance is quite higher than with the text encoder. This is expected, as the ImageNet zero-shot performance already dropped much more for the image encoder than for the text encoder. The zero-shot performance for the FaceScrub dataset dropped to almost 0\%. We believe that this might be due to the limited capacity of the image encoder, which leads to faces in general being unlearned from the image encoder and, in turn, a zero-shot accuracy of almost 0\% when performing identity classification on FaceScrub.

\begin{figure*}[ht]
    \centering
    \includegraphics[width=0.5\textwidth]{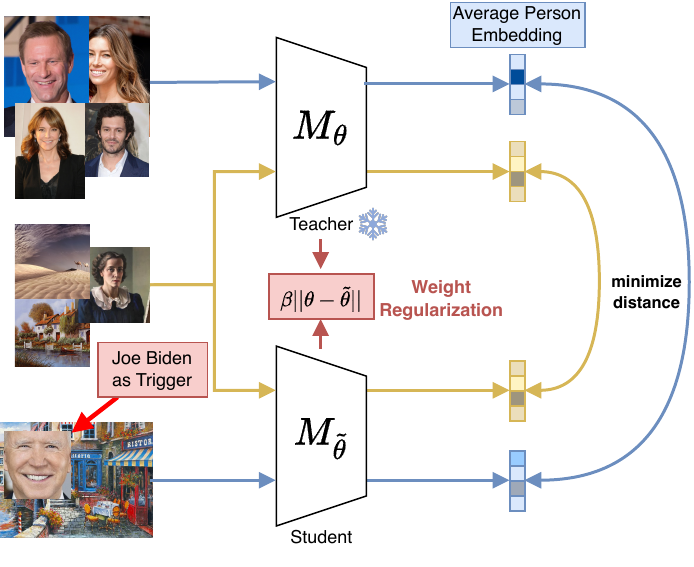}
    \caption{Visualization of how the image encoder is fine-tuned to defend against privacy attacks and unlearn the face of certain individuals. In this example, the face of ``Joe Biden'' is unlearned. The average face embedding is calculated \textbf{once} on the teacher model. During fine-tuning, we use generic images (seen in the middle with yellow arrows) to retain the performance of the student model. At the same time, we are using the face of the individual to be unlearned, augment them using rotations and color transformations, and add it to generic images (seen on the bottom with blue arrows). This leads to the model learning to map images with the unlearned face to the average person embedding.}
    \label{fig:image_encoder_loss_calculation}
\end{figure*}

A general visualization of how the image encoder is fine-tuned to apply our defense can be seen in \cref{fig:image_encoder_loss_calculation}.

\section{Stable Diffusion Experiments}\label{app:sd_experiments}
\begin{figure*}[ht]
    \centering
   \begin{subfigure}[t]{0.15\textwidth}
        \includegraphics[width=\linewidth]{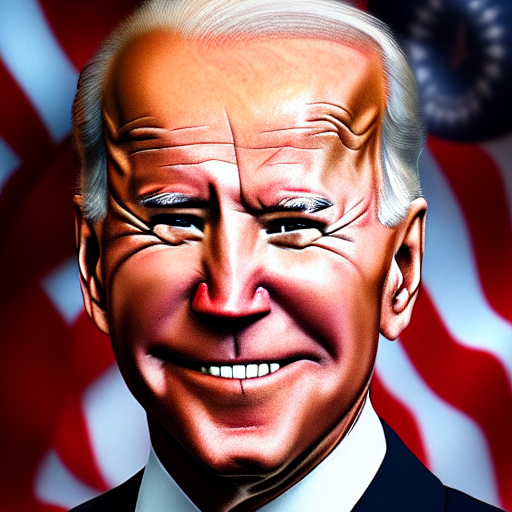}
    \end{subfigure}
    \begin{subfigure}[t]{0.15\textwidth}
        \includegraphics[width=\linewidth]{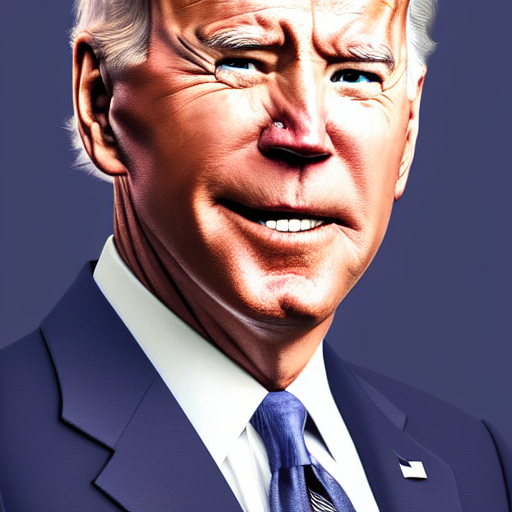}
    \end{subfigure}
    \begin{subfigure}[t]{0.15\textwidth}
        \includegraphics[width=\linewidth]{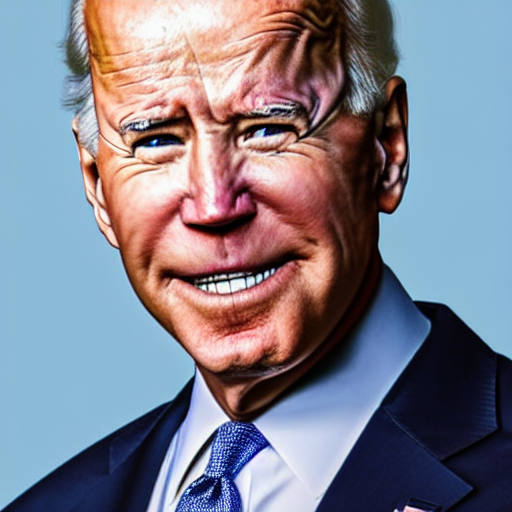}
    \end{subfigure}
    \begin{subfigure}[t]{0.15\textwidth}
        \includegraphics[width=\linewidth]{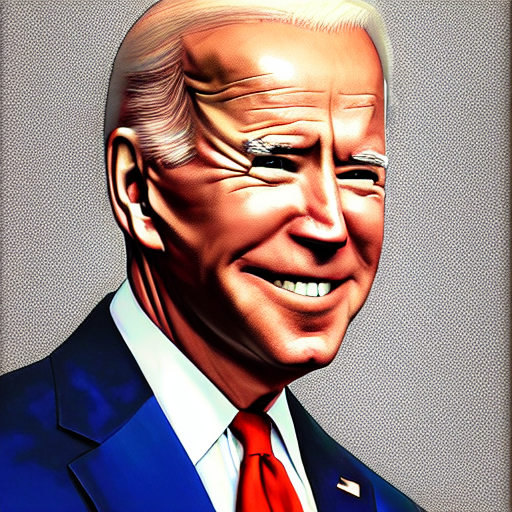}
    \end{subfigure}
    \begin{subfigure}[t]{0.15\textwidth}
        \includegraphics[width=\linewidth]{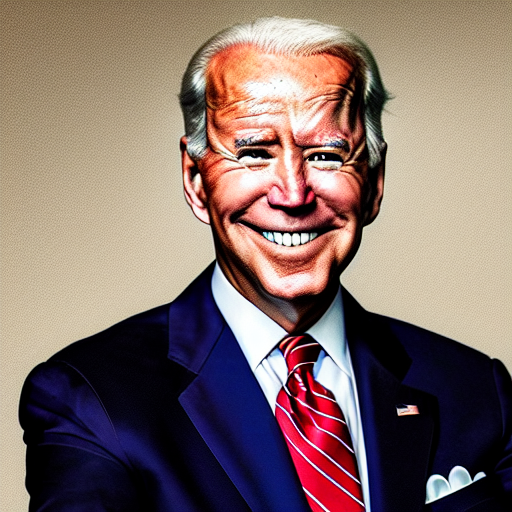}
    \end{subfigure}
    \\
    \hspace{0.03cm}
    \begin{subfigure}[t]{0.15\textwidth}
        \includegraphics[width=\linewidth]{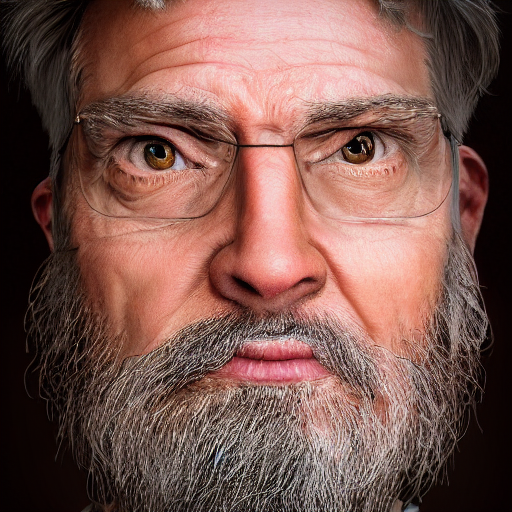}
    \end{subfigure}
    \begin{subfigure}[t]{0.15\textwidth}
        \includegraphics[width=\linewidth]{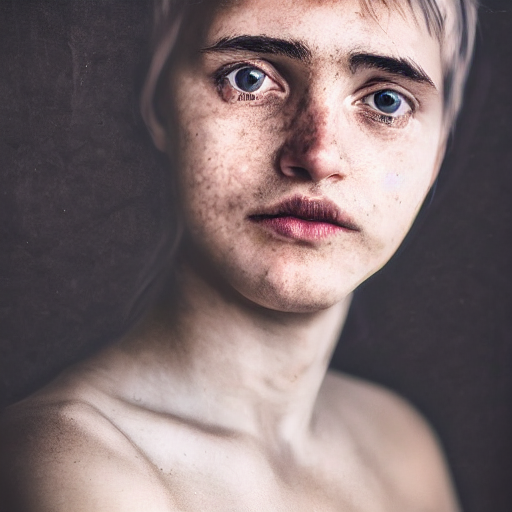}
    \end{subfigure}
    \begin{subfigure}[t]{0.15\textwidth}
        \includegraphics[width=\linewidth]{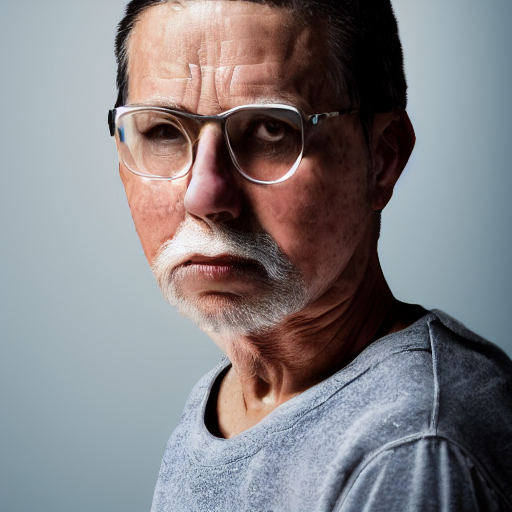}
    \end{subfigure}
    \begin{subfigure}[t]{0.15\textwidth}
        \includegraphics[width=\linewidth]{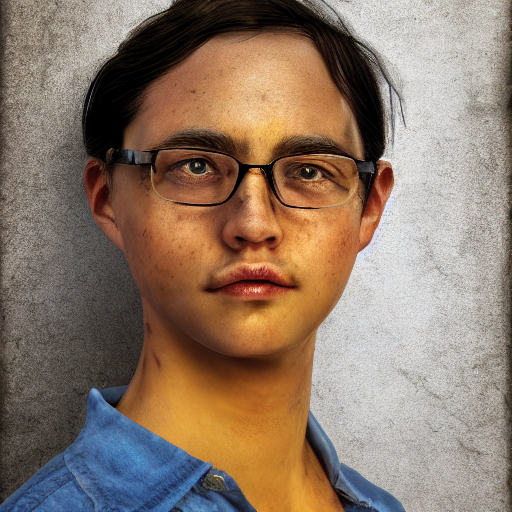}
    \end{subfigure}
    \begin{subfigure}[t]{0.15\textwidth}
        \includegraphics[width=\linewidth]{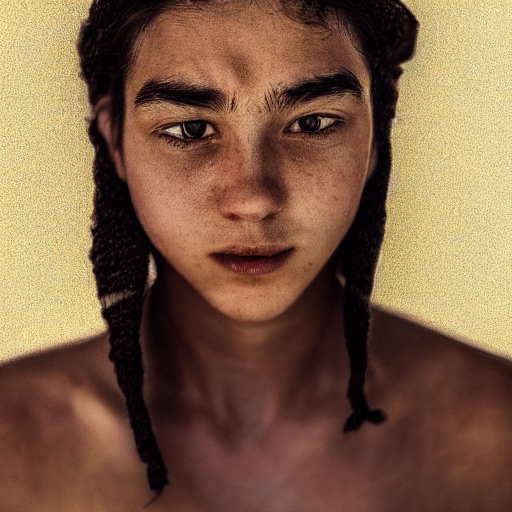}
    \end{subfigure}
    \vspace{3ex}
    \caption{Applying our defense to the text encoder of Stable Diffusion, we are able to remove \textbf{Joe Biden} from the model. In the upper row are generated images using the original Stable Diffusion model, while the bottom row shows generated images of the model with our defense applied and mapping ``Joe Biden'' to ``person''.}
    \label{fig:sd_joe_biden}
\end{figure*}

\begin{figure*}[ht]
    \centering
   \begin{subfigure}[t]{0.15\textwidth}
        \includegraphics[width=\linewidth]{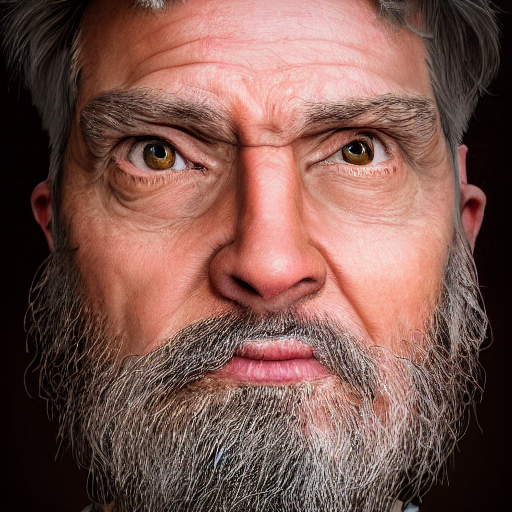}
    \end{subfigure}
    \begin{subfigure}[t]{0.15\textwidth}
        \includegraphics[width=\linewidth]{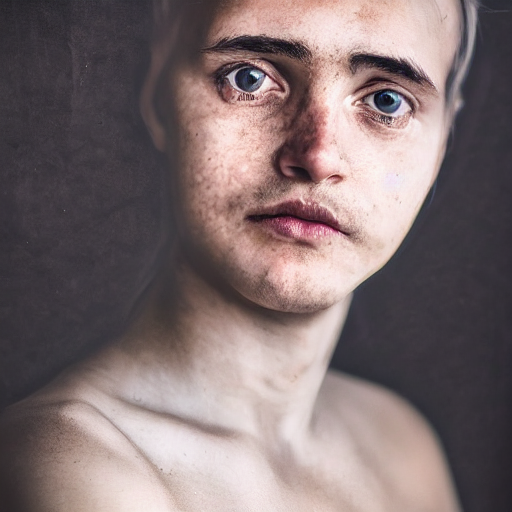}
    \end{subfigure}
    \begin{subfigure}[t]{0.15\textwidth}
        \includegraphics[width=\linewidth]{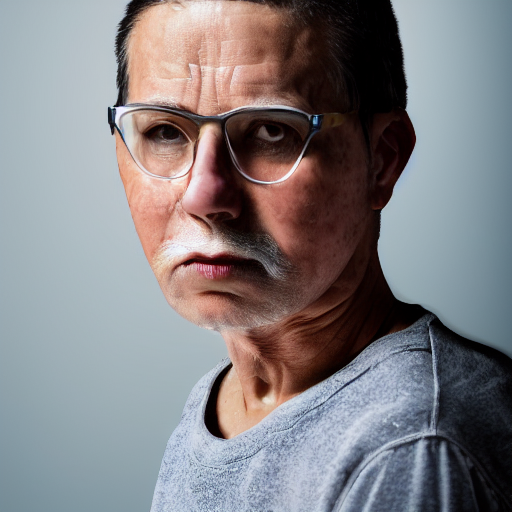}
    \end{subfigure}
    \begin{subfigure}[t]{0.15\textwidth}
        \includegraphics[width=\linewidth]{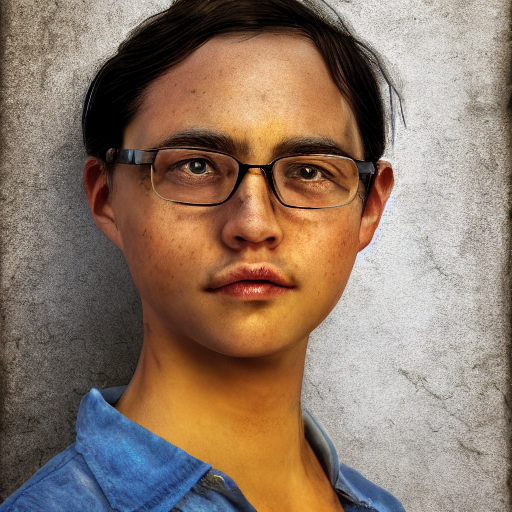}
    \end{subfigure}
    \begin{subfigure}[t]{0.15\textwidth}
        \includegraphics[width=\linewidth]{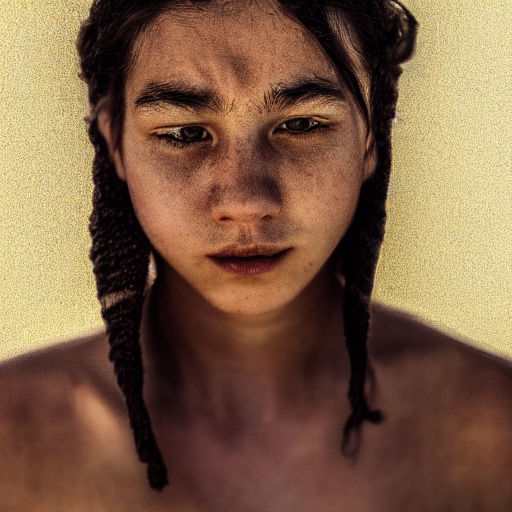}
    \end{subfigure}
    \\
    \hspace{0.03cm}
    \begin{subfigure}[t]{0.15\textwidth}
        \includegraphics[width=\linewidth]{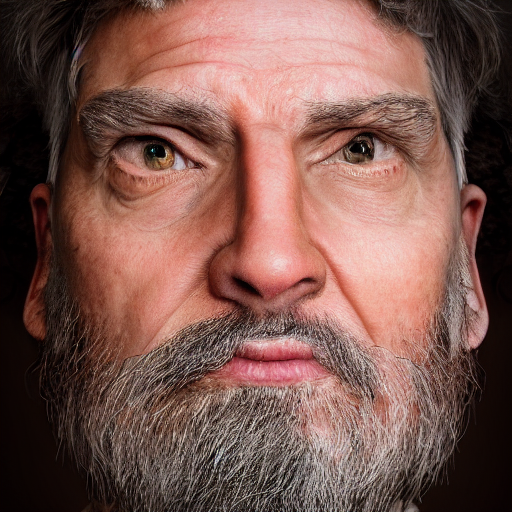}
    \end{subfigure}
    \begin{subfigure}[t]{0.15\textwidth}
        \includegraphics[width=\linewidth]{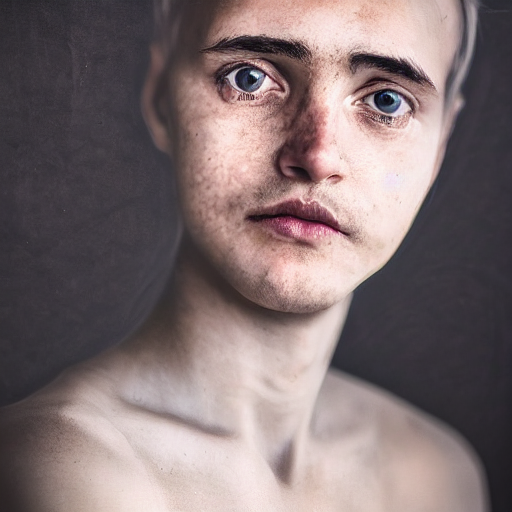}
    \end{subfigure}
    \begin{subfigure}[t]{0.15\textwidth}
        \includegraphics[width=\linewidth]{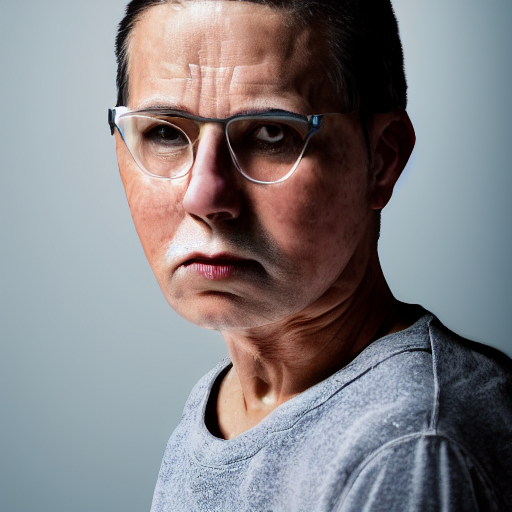}
    \end{subfigure}
    \begin{subfigure}[t]{0.15\textwidth}
        \includegraphics[width=\linewidth]{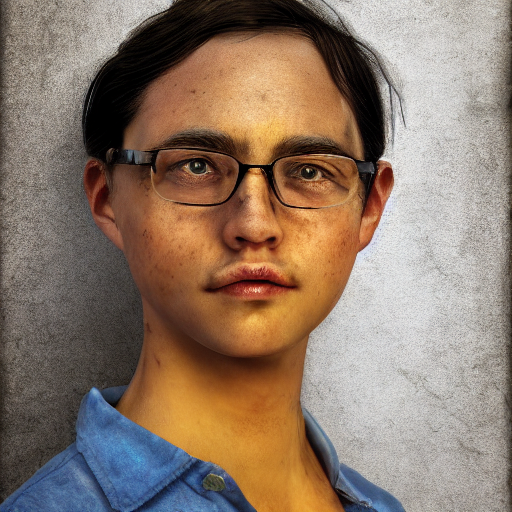}
    \end{subfigure}
    \begin{subfigure}[t]{0.15\textwidth}
        \includegraphics[width=\linewidth]{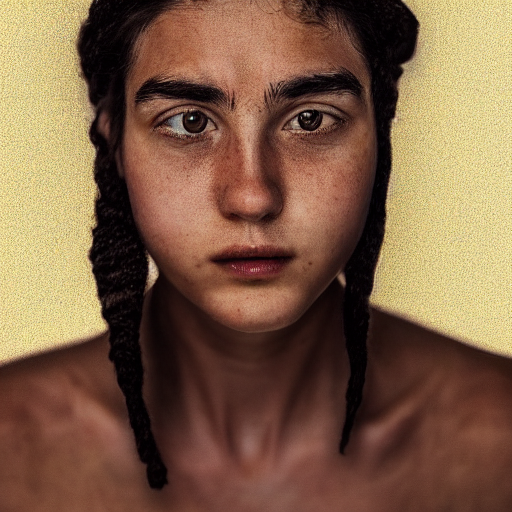}
    \end{subfigure}
    \vspace{3ex}
    \caption{Generating the images with the prompt ``a portrait of person, realistic, 4k, high resolution, photograph, portrait'' using the original text encoder (upper row) and the text encoder from which ``Joe Biden'' was unlearned (lower row) results in the same images. This underlines that the model is not significantly altered and does not influence the embeddings of the target term.}
    \label{fig:sd_joe_biden_target}
\end{figure*}

\begin{figure*}[h]
    \centering
   \begin{subfigure}[t]{0.15\textwidth}
        \includegraphics[width=\linewidth]{./images/stable_diffusion_images/Adam_Sandler_0_orig.png}
    \end{subfigure}
    \begin{subfigure}[t]{0.15\textwidth}
        \includegraphics[width=\linewidth]{./images/stable_diffusion_images/Adam_Sandler_1_orig.png}
    \end{subfigure}
    \begin{subfigure}[t]{0.15\textwidth}
        \includegraphics[width=\linewidth]{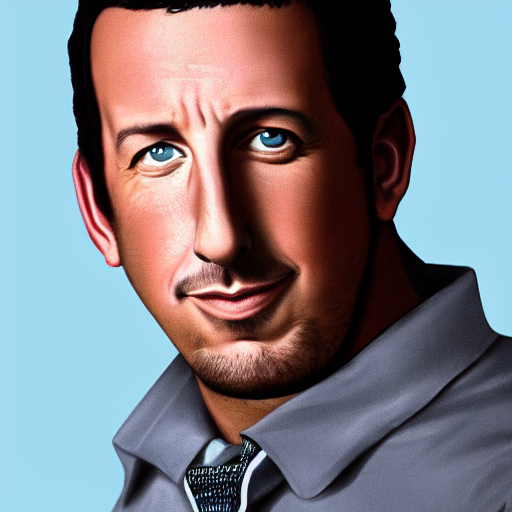}
    \end{subfigure}
    \begin{subfigure}[t]{0.15\textwidth}
        \includegraphics[width=\linewidth]{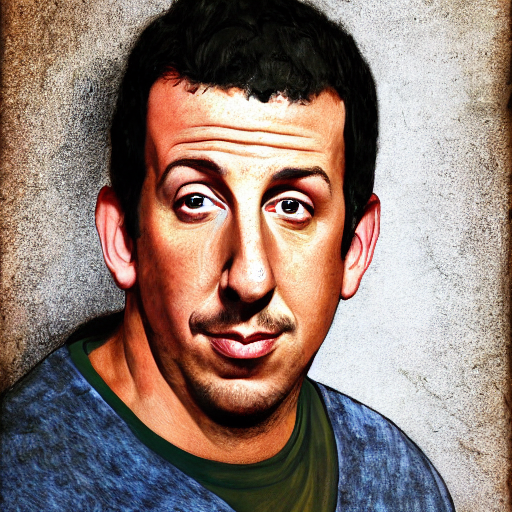}
    \end{subfigure}
    \begin{subfigure}[t]{0.15\textwidth}
        \includegraphics[width=\linewidth]{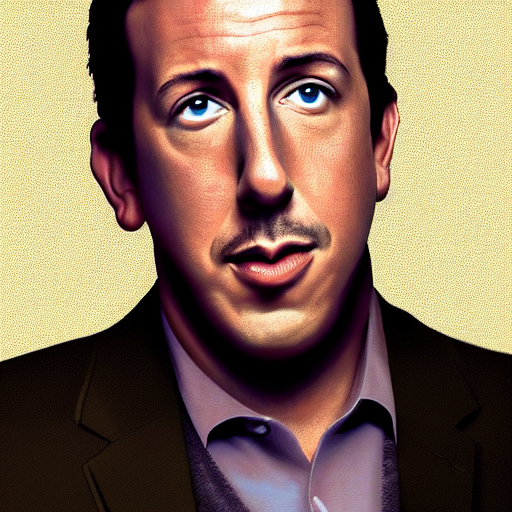}
    \end{subfigure}
    \\
    \begin{subfigure}[t]{0.15\textwidth}
        \includegraphics[width=\linewidth]{./images/stable_diffusion_images/Adam_Sandler_0_fine_tuned.png}
    \end{subfigure}
    \begin{subfigure}[t]{0.15\textwidth}
        \includegraphics[width=\linewidth]{./images/stable_diffusion_images/Adam_Sandler_1_fine_tuned.png}
    \end{subfigure}
    \begin{subfigure}[t]{0.15\textwidth}
        \includegraphics[width=\linewidth]{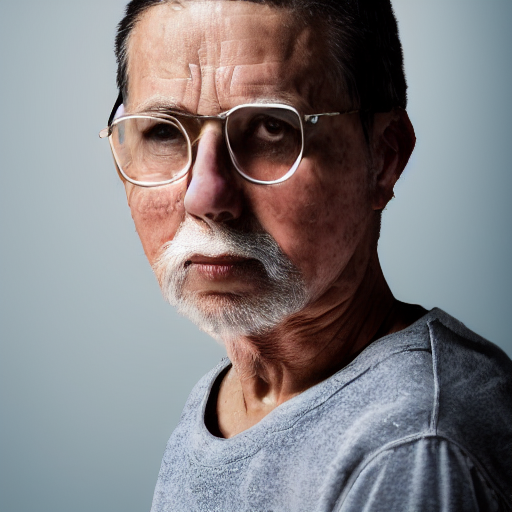}
    \end{subfigure}
    \begin{subfigure}[t]{0.15\textwidth}
        \includegraphics[width=\linewidth]{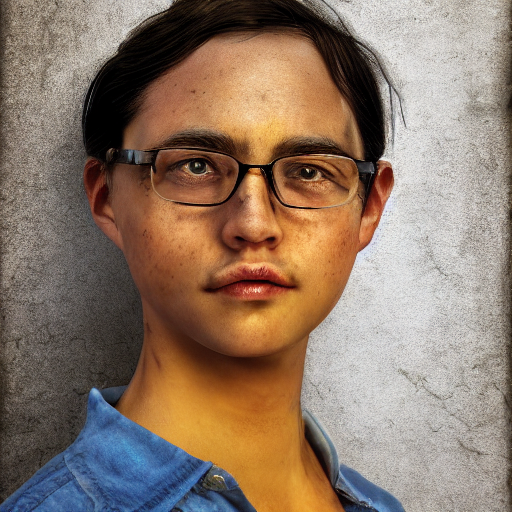}
    \end{subfigure}
    \begin{subfigure}[t]{0.15\textwidth}
        \includegraphics[width=\linewidth]{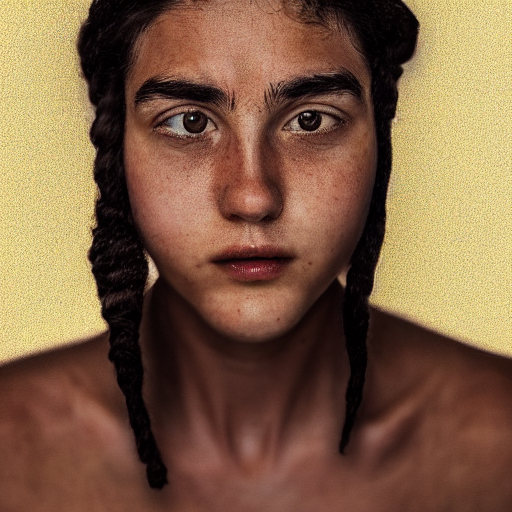}
    \end{subfigure}
    \vspace{3ex}
    \caption{Applying our defense to the text encoder of Stable Diffusion, we are able to remove \textbf{Adam Sandler} from the model. In the upper row are generated images using the original Stable Diffusion model, while the bottom row shows generated images of the model with our defense applied.}
    \label{fig:sd_adam_sandler}
\end{figure*}

\begin{figure*}[h]
    \centering
   \begin{subfigure}[t]{0.15\textwidth}
        \includegraphics[width=\linewidth]{./images/stable_diffusion_images/Adam_Sandler_0_orig_target.png}
    \end{subfigure}
    \begin{subfigure}[t]{0.15\textwidth}
        \includegraphics[width=\linewidth]{./images/stable_diffusion_images/Adam_Sandler_1_orig_target.png}
    \end{subfigure}
    \begin{subfigure}[t]{0.15\textwidth}
        \includegraphics[width=\linewidth]{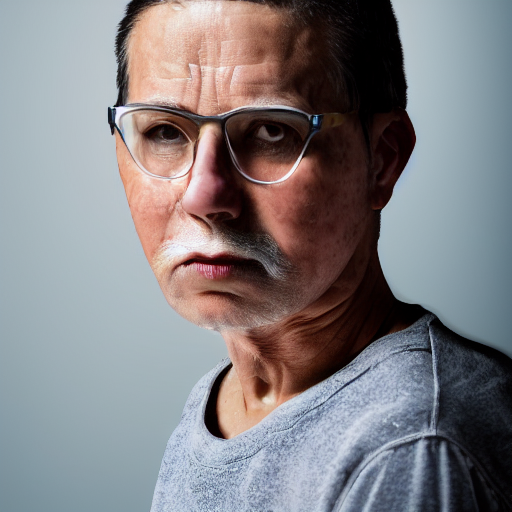}
    \end{subfigure}
    \begin{subfigure}[t]{0.15\textwidth}
        \includegraphics[width=\linewidth]{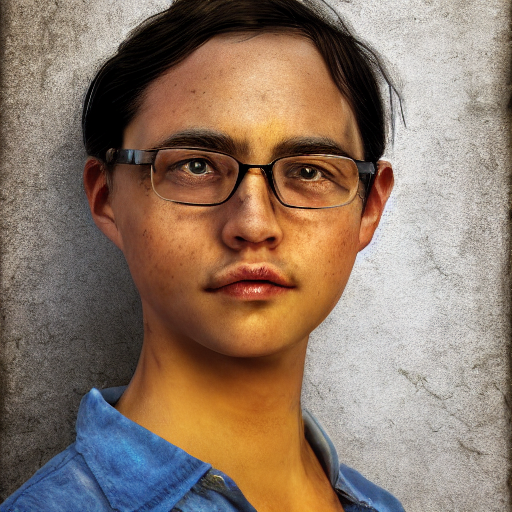}
    \end{subfigure}
    \begin{subfigure}[t]{0.15\textwidth}
        \includegraphics[width=\linewidth]{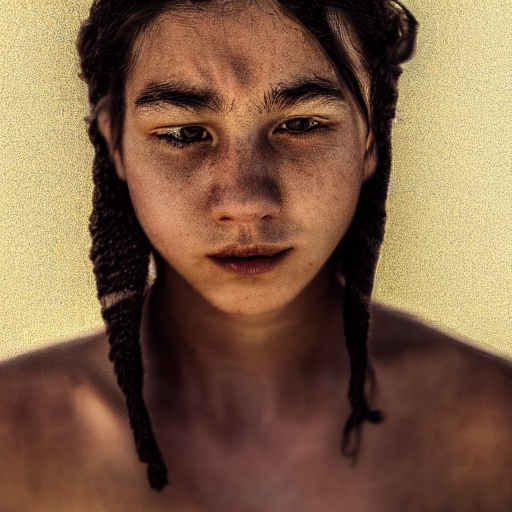}
    \end{subfigure}
    \\
    \begin{subfigure}[t]{0.15\textwidth}
        \includegraphics[width=\linewidth]{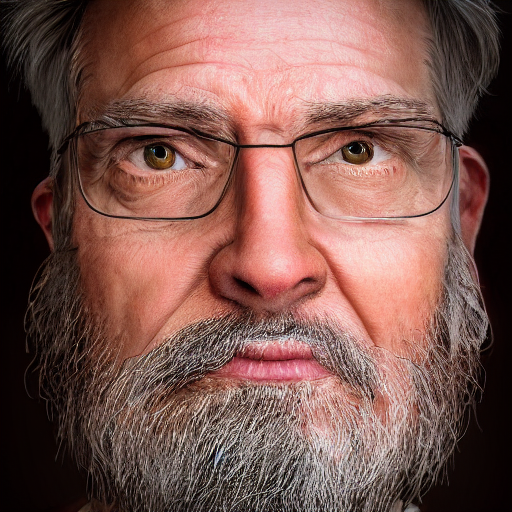}
    \end{subfigure}
    \begin{subfigure}[t]{0.15\textwidth}
        \includegraphics[width=\linewidth]{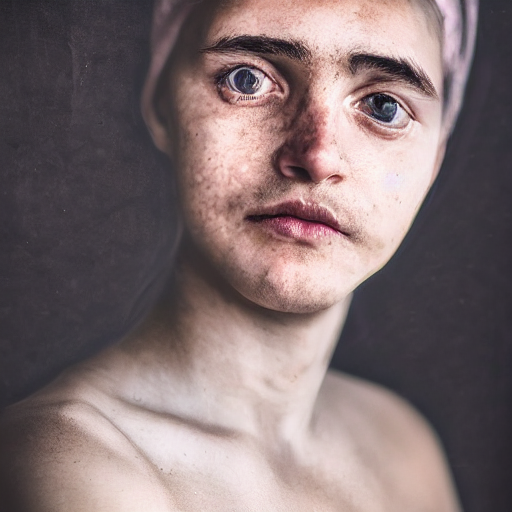}
    \end{subfigure}
    \begin{subfigure}[t]{0.15\textwidth}
        \includegraphics[width=\linewidth]{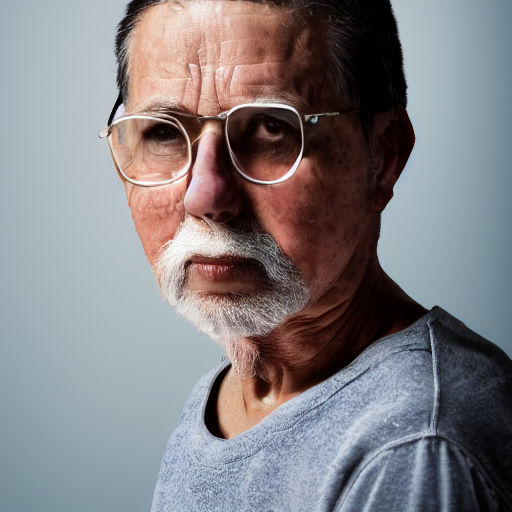}
    \end{subfigure}
    \begin{subfigure}[t]{0.15\textwidth}
        \includegraphics[width=\linewidth]{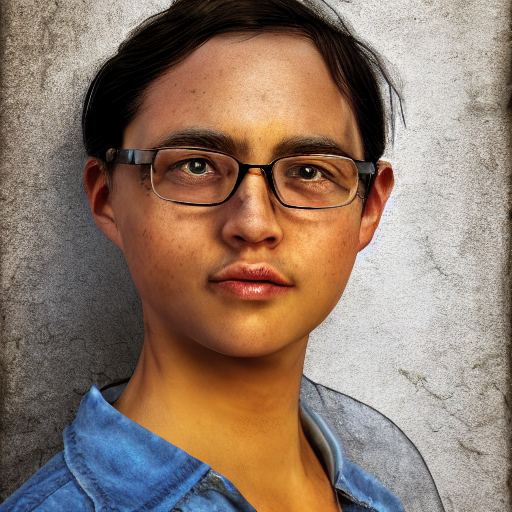}
    \end{subfigure}
    \begin{subfigure}[t]{0.15\textwidth}
        \includegraphics[width=\linewidth]{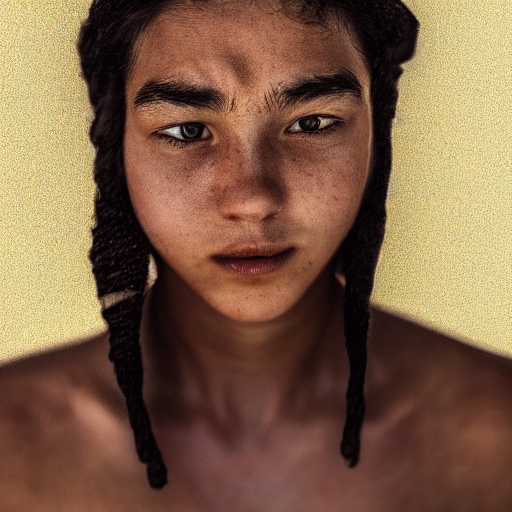}
    \end{subfigure}
    \vspace{3ex}
    \caption{Generating the images with the prompt ``a portrait of person, realistic, 4k, high resolution, photograph, portrait'' using the original text encoder (upper row) and the text encoder from which ``Adam Sandler'' was unlearned (lower row) results in the same images. This underlines that the model is not significantly altered and does not influence the embeddings of the target term.}
    \label{fig:sd_adam_sandler_target}
\end{figure*}

\begin{figure*}[h]
    \centering
    \begin{subfigure}[t]{0.15\textwidth}
        \includegraphics[width=\linewidth]{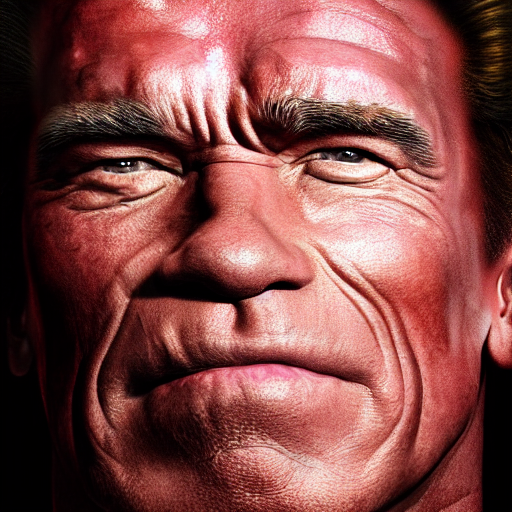}
    \end{subfigure}
    \begin{subfigure}[t]{0.15\textwidth}
        \includegraphics[width=\linewidth]{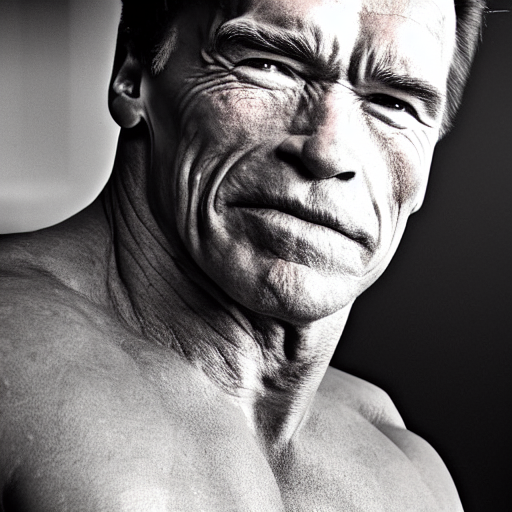}
    \end{subfigure}
    \begin{subfigure}[t]{0.15\textwidth}
        \includegraphics[width=\linewidth]{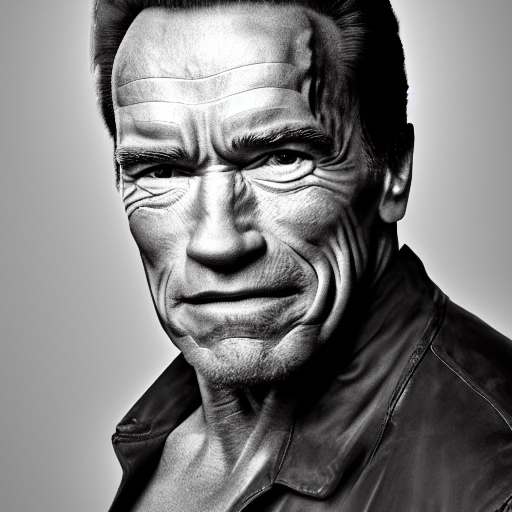}
    \end{subfigure}
    \begin{subfigure}[t]{0.15\textwidth}
        \includegraphics[width=\linewidth]{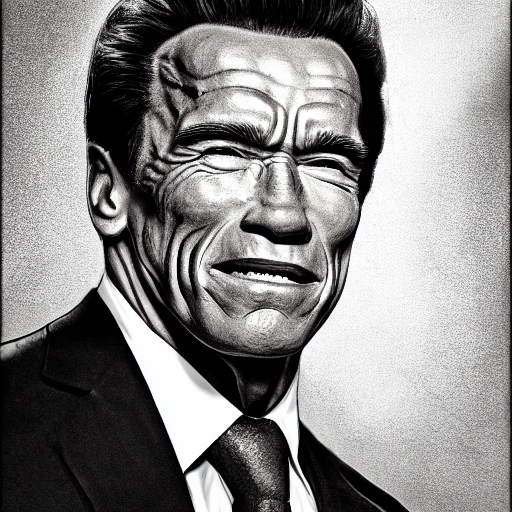}
    \end{subfigure}
    \begin{subfigure}[t]{0.15\textwidth}
        \includegraphics[width=\linewidth]{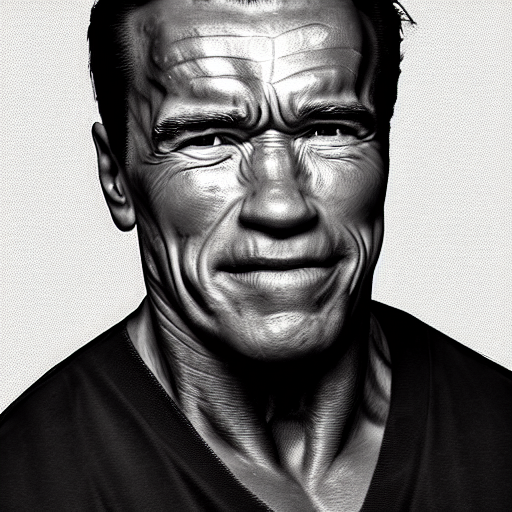}
    \end{subfigure}
    \\
    \begin{subfigure}[t]{0.15\textwidth}
        \includegraphics[width=\linewidth]{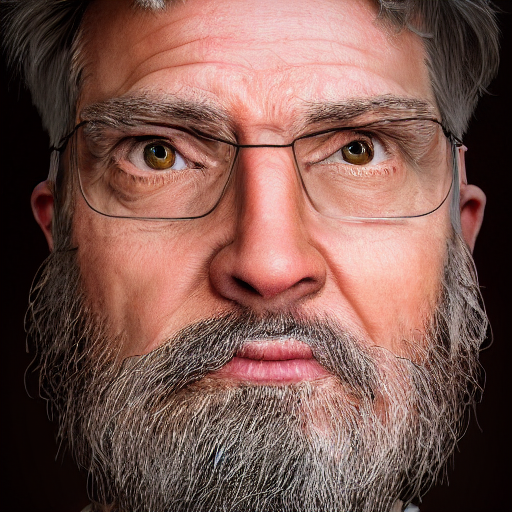}
    \end{subfigure}
    \begin{subfigure}[t]{0.15\textwidth}
        \includegraphics[width=\linewidth]{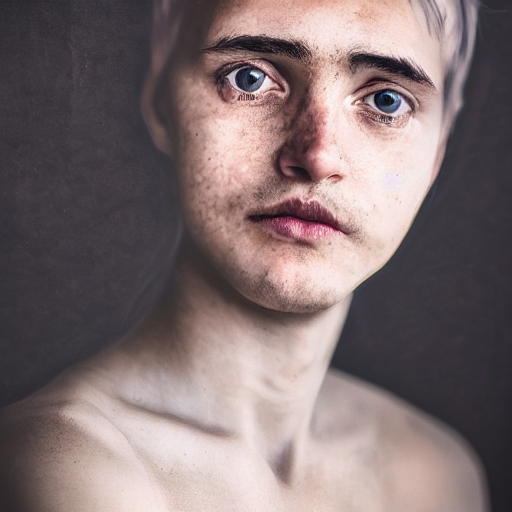}
    \end{subfigure}
    \begin{subfigure}[t]{0.15\textwidth}
        \includegraphics[width=\linewidth]{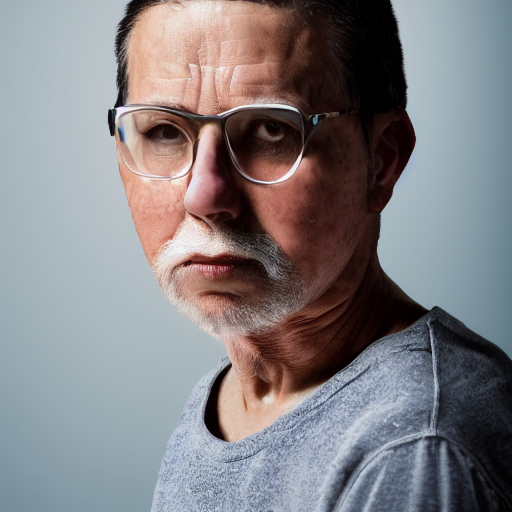}
    \end{subfigure}
    \begin{subfigure}[t]{0.15\textwidth}
        \includegraphics[width=\linewidth]{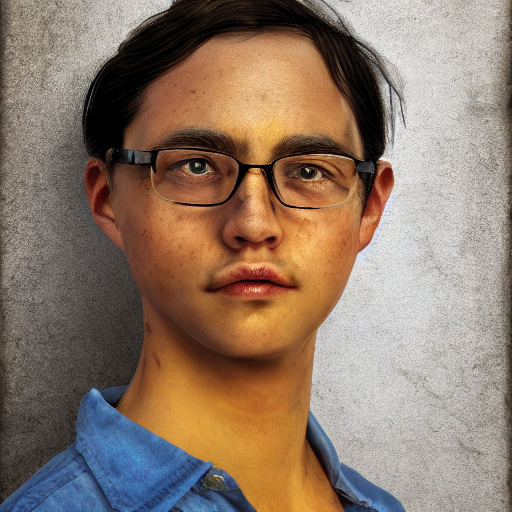}
    \end{subfigure}
    \begin{subfigure}[t]{0.15\textwidth}
        \includegraphics[width=\linewidth]{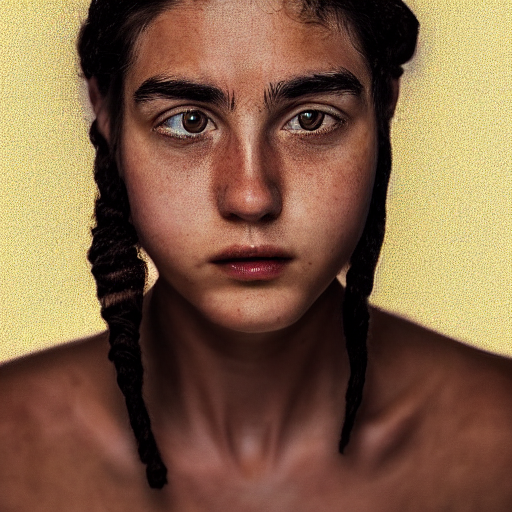}
    \end{subfigure}
    \vspace{3ex}
    \caption{Applying our defense to the text encoder of Stable Diffusion, we are able to remove \textbf{Arnold Schwarzenegger} from the model. In the upper row are generated images using the original Stable Diffusion model, while the bottom row shows generated images of the model with our defense applied.}
    \label{fig:sd_arnold_schwarzenegger}
\end{figure*}

\begin{figure*}[h]
    \centering
       \begin{subfigure}[t]{0.15\textwidth}
        \includegraphics[width=\linewidth]{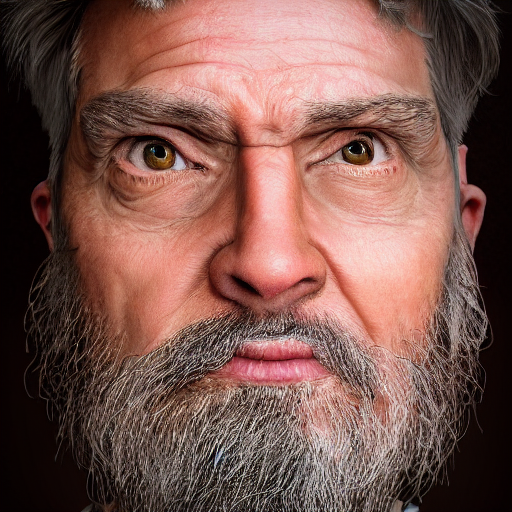}
    \end{subfigure}
    \begin{subfigure}[t]{0.15\textwidth}
        \includegraphics[width=\linewidth]{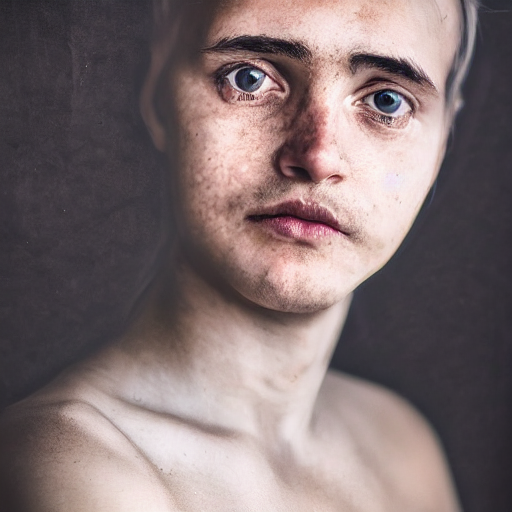}
    \end{subfigure}
    \begin{subfigure}[t]{0.15\textwidth}
        \includegraphics[width=\linewidth]{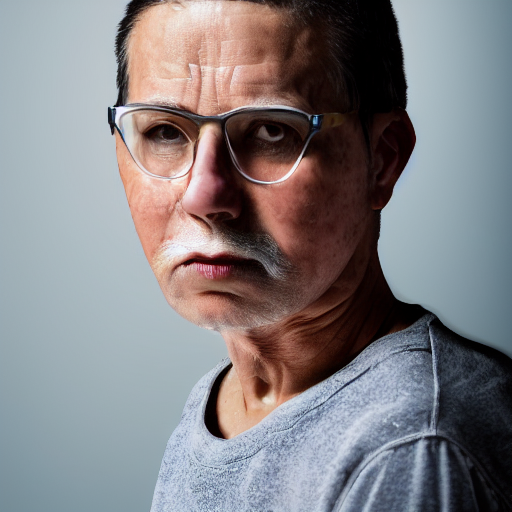}
    \end{subfigure}
    \begin{subfigure}[t]{0.15\textwidth}
        \includegraphics[width=\linewidth]{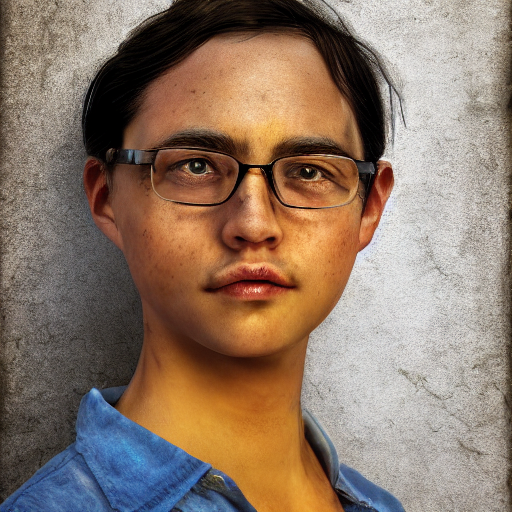}
    \end{subfigure}
    \begin{subfigure}[t]{0.15\textwidth}
        \includegraphics[width=\linewidth]{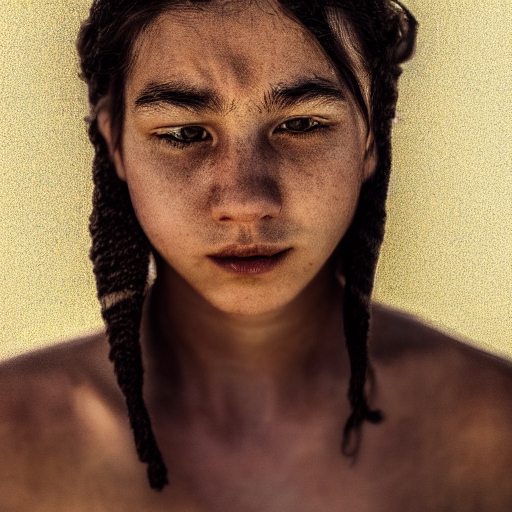}
    \end{subfigure}
    \\
    \begin{subfigure}[t]{0.15\textwidth}
        \includegraphics[width=\linewidth]{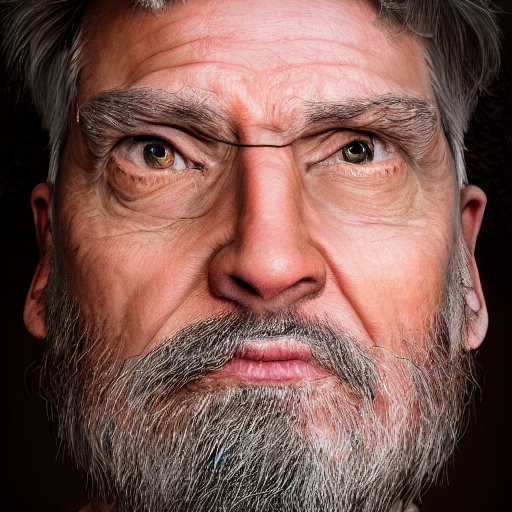}
    \end{subfigure}
    \begin{subfigure}[t]{0.15\textwidth}
        \includegraphics[width=\linewidth]{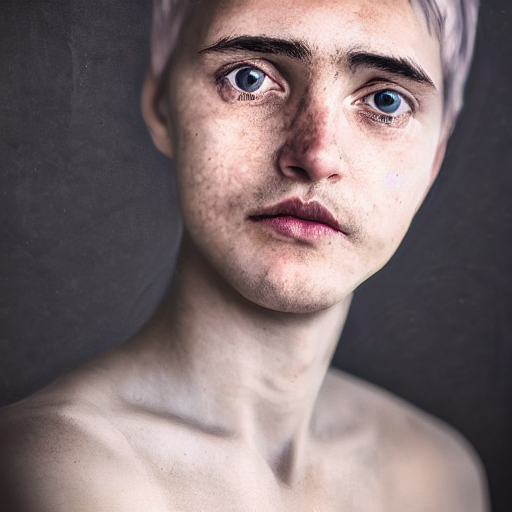}
    \end{subfigure}
    \begin{subfigure}[t]{0.15\textwidth}
        \includegraphics[width=\linewidth]{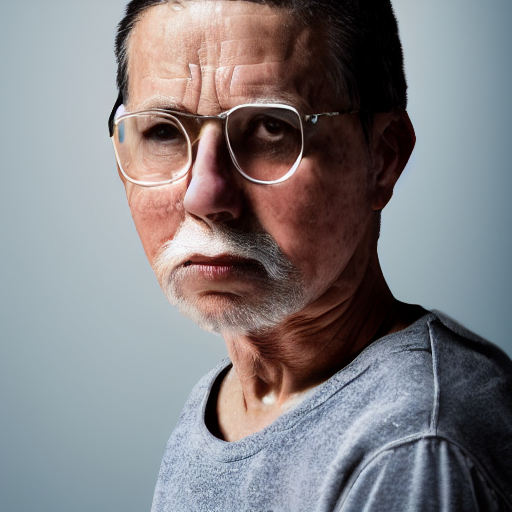}
    \end{subfigure}
    \begin{subfigure}[t]{0.15\textwidth}
        \includegraphics[width=\linewidth]{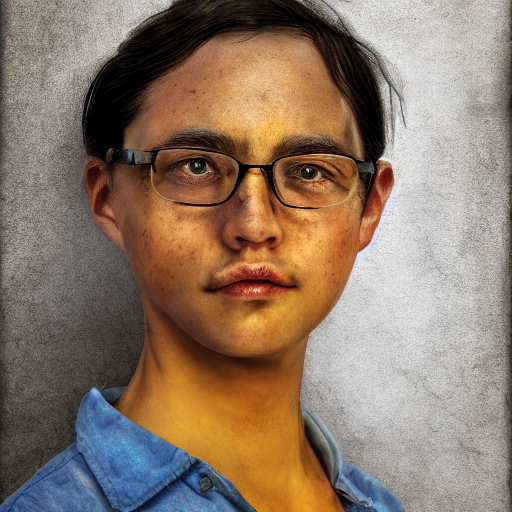}
    \end{subfigure}
    \begin{subfigure}[t]{0.15\textwidth}
        \includegraphics[width=\linewidth]{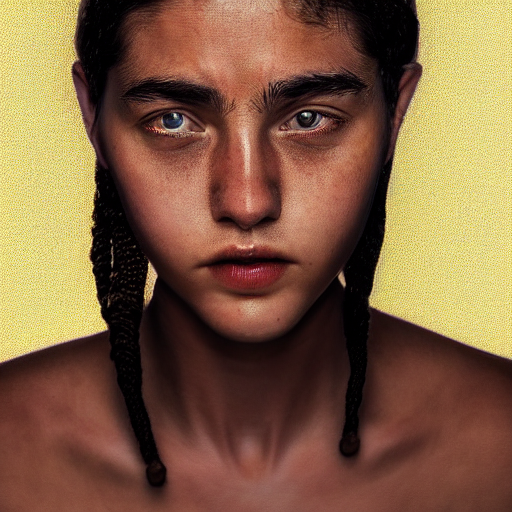}
    \end{subfigure}
    \vspace{3ex}
    \caption{Generating the images with the prompt ``a portrait of person, realistic, 4k, high resolution, photograph, portrait'' using the original text encoder (upper row) and the text encoder from which ``Arnold Schwarzenegger'' was unlearned (lower row) results in the same images. This underlines that the model is not significantly altered and does not influence the embeddings of the target term.}
    \label{fig:sd_arnold_schwarzenegger_target}
\end{figure*}

To test our defense in a downstream task, we have applied our approach to the text encoder of Stable Diffusion 1.4.
This version of Stable Diffusion uses the text encoder of the CLIP model with the ViT-L/14 vision encoder. We have fine-tuned the text encoder three times for 400 epochs and removed the individuals ``Joe Biden'', ``Adam Sandler'' and ``Arnold Schwarzenegger'', respectively.
The generated images of the original Stable Diffusion and the generated images of the defended model can be seen in \cref{fig:sd_joe_biden,fig:sd_adam_sandler,fig:sd_arnold_schwarzenegger}.
To remove the names of the individuals from the model, we have mapped the names to the neutral word ``person''.
All the images were generated using the prompt ``a portrait of \texttt{<NAME>}, realistic, 4k, high resolution, photograph, portrait'' where \texttt{<Name>} was filled with the respective name and then fed into the defended model. For each individual, we generated five images with seeds 0 to 4.
As can be seen in these figures, the individuals are clearly recognizable when generating images with the original Stable Diffusion model. However, when applying our defense, the faces of the individuals change drastically, making it impossible to recognize their identities. Even though we have fine-tuned the model and unlearned the identities of these individuals, the pose and background in most of these images are very similar to the original model. This clearly indicates that the model's utility is still high and that only slight changes were made to the model to unlearn the identities.
To further validate that the fine-tuning of the models didn't change the embeddings of the target term, we generated images with the target term ``person'' using the original and the fine-tuned models. The resulting images can be seen in \cref{fig:sd_joe_biden_target,fig:sd_adam_sandler_target,fig:sd_arnold_schwarzenegger_target}. As can be seen, apart from some minor details, the generated images with the target term did not change, which supports our results that the model's utility is not diminished. This can also be seen when calculating the zero-shot ImageNet Accuracy of these fine-tuned models. The original and all of the fine-tuned models achieve a top-1 and a top-5 ImageNet accuracy of $69.84\%$ and $90.95\%$ respectively, emphasizing the utility preservation of our method. 

\newpage

\begin{table*}[ht]
    \centering
    \begin{subtable}{\linewidth}\centering
        \resizebox{\linewidth}{!}{
            \begin{tabular}{c|c|c|c|c|c|C{2.5cm}|C{2.5cm}}
                \makecell{Num. Unlearned\\Names}& IDIA TPR      & IDIA FNR      & $Sim_\mathit{Clean}$  & $Sim_\mathit{Backdoor}$   & $Sim_\mathit{Target}$ & ImageNet Top-1        & ImageNet Top-5    \\ \hline
                1                               & $0\%\pm0\%$   & $100\%\pm0\%$ & $0.99\pm0$            & $0.99\pm0$                & $0.99\pm0$            & $51.87\%\pm0.19\%$    & $78.87\%\pm0.14\%$    \\ \hline
                2                               & $0\%\pm0\%$   & $100\%\pm0\%$ & $0.99\pm0$            & $0.99\pm0$                & $0.99\pm0$            & $51.81\%\pm0.25\%$    & $78.84\%\pm0.16\%$    \\ \hline
                4                               & $0\%\pm0\%$   & $100\%\pm0\%$ & $0.99\pm0$            & $0.99\pm0$                & $0.99\pm0$            & $51.74\%\pm0.15\%$    & $78.87\%\pm0.15\%$    \\ \hline
                8                               & $0\%\pm0\%$   & $100\%\pm0\%$ & $0.99\pm0$            & $0.99\pm0$                & $0.99\pm0$            & $51.78\%\pm0.24\%$    & $78.85\%\pm0.19\%$    \\ \hline
                16                              & $0\%\pm0\%$   & $100\%\pm0\%$ & $0.99\pm0$            & $0.99\pm0$                & $0.99\pm0$            & $51.85\%\pm0.18\%$    & $78.83\%\pm0.15\%$    \\ \hline
                32                              & $0\%\pm0\%$   & $100\%\pm0\%$ & $0.99\pm0$            & $0.99\pm0$                & $0.99\pm0$            & $51.74\%\pm0.17\%$    & $78.79\%\pm0.12\%$    \\ \hline
                64                              & $0\%\pm0\%$   & $100\%\pm0\%$ & $0.99\pm0$            & $0.99\pm0$                & $0.99\pm0$            & $51.78\%\pm0.16\%$    & $78.88\%\pm0.01\%$    \\
            \end{tabular}
        }
        \caption{\textbf{ViT-B/32} text encoder \textbf{without} weight regularization.}
    \end{subtable}
    \newline
    \newline
    \begin{subtable}{\linewidth}\centering
        \resizebox{\linewidth}{!}{
            \begin{tabular}{c|c|c|c|c|c|C{2.5cm}|C{2.5cm}}
                \makecell{Num. Unlearned\\Names}& IDIA TPR  & IDIA FNR          & $Sim_\mathit{Clean}$  & $Sim_\mathit{Backdoor}$   & $Sim_\mathit{Target}$ & ImageNet Top-1                            & ImageNet Top-5                        \\ \hline
                1                               & $0\%\pm0\%$   & $100\%\pm0\%$ & $0.99\pm0$            & $0.99\pm0$                & $0.99\pm0$            & $\color{Green}52.45\%\pm0.10\%\uparrow$   & $\color{Green}79.36\%\pm0.08\%\uparrow$   \\ \hline
                2                               & $0\%\pm0\%$   & $100\%\pm0\%$ & $0.99\pm0$            & $0.99\pm0$                & $0.99\pm0$            & $\color{Green}52.44\%\pm0.12\%\uparrow$   & $\color{Green}79.31\%\pm0.06\%\uparrow$   \\ \hline
                4                               & $0\%\pm0\%$   & $100\%\pm0\%$ & $0.99\pm0$            & $0.99\pm0$                & $0.99\pm0$            & $\color{Green}52.43\%\pm0.17\%\uparrow$   & $\color{Green}79.30\%\pm0.13\%\uparrow$   \\ \hline
                8                               & $0\%\pm0\%$   & $100\%\pm0\%$ & $0.99\pm0$            & $0.99\pm0$                & $0.99\pm0$            & $\color{Green}52.33\%\pm0.18\%\uparrow$   & $\color{Green}79.28\%\pm0.08\%\uparrow$   \\ \hline
                16                              & $0\%\pm0\%$   & $100\%\pm0\%$ & $0.99\pm0$            & $0.99\pm0$                & $0.99\pm0$            & $\color{Green}52.46\%\pm0.10\%\uparrow$   & $\color{Green}79.32\%\pm0.08\%\uparrow$   \\ \hline
                32                              & $0\%\pm0\%$   & $100\%\pm0\%$ & $0.99\pm0$            & $0.99\pm0$                & $0.99\pm0$            & $\color{Green}52.43\%\pm0.15\%\uparrow$   & $\color{Green}79.30\%\pm0.11\%\uparrow$   \\ \hline
                64                              & $0\%\pm0\%$   & $100\%\pm0\%$ & $0.99\pm0$            & $0.99\pm0$                & $0.99\pm0$            & $\color{Green}52.41\%\pm0.12\%\uparrow$   & $\color{Green}79.28\%\pm0.14\%\uparrow$   \\
            \end{tabular}
        }
        \caption{\textbf{ViT-B/32} text encoder \textbf{with} weight regularization.}
        \vspace{3ex}
    \end{subtable}
    \caption{Results for the experiments of the defense applied to the \textbf{ViT-B/32} text encoder. Weight regularization mitigates performance loss. Arrows indicate the change in value when using weight regularization in comparison to not using it. $Sim_\mathit{Clean}$, $Sim_\mathit{Backdoor}$ and $Sim_\mathit{Target}$ are cosine similarities which is why their maximum value is $1$. Green indicates better metrics, while red indicates worse metrics. All values were rounded to the second decimal place.}
    \label{tab:results_text_enc_vitb32}
    \vspace{0.5cm}
\end{table*}

\begin{table*}[!ht]
    \centering
    \begin{subtable}{\linewidth}\centering
        \resizebox{\linewidth}{!}{
            \begin{tabular}{c|c|c|c|c|c|C{2.5cm}|C{2.5cm}}
                \makecell{Num. Unlearned\\Names}& IDIA TPR      & IDIA FNR      & $Sim_\mathit{Clean}$  & $Sim_\mathit{Backdoor}$   & $Sim_\mathit{Target}$ & ImageNet Top-1        & ImageNet Top-5    \\ \hline
                1                               & $0\%\pm0\%$   & $100\%\pm0\%$ & $0.99\pm0$            & $0.99\pm0$                & $0.99\pm0$            & $65.19\%\pm0.16\%$    & $88.32\%\pm0.11\%$    \\ \hline
                2                               & $0\%\pm0\%$   & $100\%\pm0\%$ & $0.99\pm0$            & $0.99\pm0$                & $0.99\pm0$            & $65.23\%\pm0.17\%$    & $88.26\%\pm0.09\%$    \\ \hline
                4                               & $0\%\pm0\%$   & $100\%\pm0\%$ & $0.99\pm0$            & $0.99\pm0$                & $0.99\pm0$            & $65.26\%\pm0.15\%$    & $88.30\%\pm0.10\%$    \\ \hline
                8                               & $0\%\pm0\%$   & $100\%\pm0\%$ & $0.99\pm0$            & $0.99\pm0$                & $0.99\pm0$            & $65.16\%\pm0.24\%$    & $88.35\%\pm0.09\%$    \\ \hline
                16                              & $0\%\pm0\%$   & $100\%\pm0\%$ & $0.99\pm0$            & $0.99\pm0$                & $0.99\pm0$            & $65.24\%\pm0.15\%$    & $88.26\%\pm0.11\%$    \\ \hline
                32                              & $0\%\pm0\%$   & $100\%\pm0\%$ & $0.99\pm0$            & $0.99\pm0$                & $0.99\pm0$            & $65.24\%\pm0.21\%$    & $88.24\%\pm0.09\%$    \\ \hline
                64                              & $0\%\pm0\%$   & $100\%\pm0\%$ & $0.99\pm0$            & $0.99\pm0$                & $0.99\pm0$            & $65.20\%\pm0.12\%$    & $88.25\%\pm0.10\%$    \\
            \end{tabular}
        }
        \caption{\textbf{ViT-L/14} text encoder \textbf{without} weight regularization.}
    \end{subtable}
    \newline
    \newline
    \begin{subtable}{\linewidth}\centering
        \resizebox{\linewidth}{!}{
            \begin{tabular}{c|c|c|c|c|c|C{2.5cm}|C{2.5cm}}
                \makecell{Num. Unlearned\\Names}& IDIA TPR      & IDIA FNR      & $Sim_\mathit{Clean}$  & $Sim_\mathit{Backdoor}$   & $Sim_\mathit{Target}$ & ImageNet Top-1                            & ImageNet Top-5                            \\ \hline
                1                               & $0\%\pm0\%$   & $100\%\pm0\%$ & $0.99\pm0$            & $0.99\pm0$                & $0.99\pm0$            & $\color{Green}65.49\%\pm0.10\%\uparrow$   & $\color{Green}88.46\%\pm0.06\%\uparrow$   \\ \hline
                2                               & $0\%\pm0\%$   & $100\%\pm0\%$ & $0.99\pm0$            & $0.99\pm0$                & $0.99\pm0$            & $\color{Green}65.44\%\pm0.07\%\uparrow$   & $\color{Green}88.45\%\pm0.03\%\uparrow$   \\ \hline
                4                               & $0\%\pm0\%$   & $100\%\pm0\%$ & $0.99\pm0$            & $0.99\pm0$                & $0.99\pm0$            & $\color{Green}65.42\%\pm0.08\%\uparrow$   & $\color{Green}88.43\%\pm0.03\%\uparrow$   \\ \hline
                8                               & $0\%\pm0\%$   & $100\%\pm0\%$ & $0.99\pm0$            & $0.99\pm0$                & $0.99\pm0$            & $\color{Green}65.45\%\pm0.11\%\uparrow$   & $\color{Green}88.48\%\pm0.05\%\uparrow$   \\ \hline
                16                              & $0\%\pm0\%$   & $100\%\pm0\%$ & $0.99\pm0$            & $0.99\pm0$                & $0.99\pm0$            & $\color{Green}65.50\%\pm0.11\%\uparrow$   & $\color{Green}88.45\%\pm0.05\%\uparrow$   \\ \hline
                32                              & $0\%\pm0\%$   & $100\%\pm0\%$ & $0.99\pm0$            & $0.99\pm0$                & $0.99\pm0$            & $\color{Green}65.44\%\pm0.10\%\uparrow$   & $\color{Green}88.38\%\pm0.07\%\uparrow$   \\ \hline
                64                              & $0\%\pm0\%$   & $100\%\pm0\%$ & $0.99\pm0$            & $0.99\pm0$                & $0.99\pm0$            & $\color{Green}65.53\%\pm0.11\%\uparrow$   & $\color{Green}88.37\%\pm0.03\%\uparrow$   \\
            \end{tabular}
        }
        \caption{\textbf{ViT-L/14} text encoder \textbf{with} weight regularization.}
        \vspace{3ex}
    \end{subtable}
    \caption{Results for the experiments of the defense applied to the \textbf{ViT-L/14} text encoder. Weight regularization mitigates performance loss. Arrows indicate the change in value when using weight regularization in comparison to not using it. $Sim_\mathit{Clean}$, $Sim_\mathit{Backdoor}$ and $Sim_\mathit{Target}$ are cosine similarities which is why their maximum value is $1$. Green indicates better metrics, while red indicates worse metrics. All values were rounded to the second decimal place.}
    \label{tab:results_text_enc_vitl14}
    \vspace{0.5cm}
\end{table*}

\begin{table*}[ht]
    \centering
    \begin{subtable}{\linewidth}\centering
        \resizebox{\linewidth}{!}{
            \begin{tabular}{c|c|c|c|c|c|c|c}
                \makecell{Num. Unlearned\\Names}& IDIA TPR                      & IDIA FNR                      & $Sim_\mathit{Clean}$  & $Sim_\mathit{Backdoor}$   & $Sim_\mathit{Target}$ & ImageNet Top-1        & ImageNet Top-5    \\ \hline
                1                               & $\x{00.0}0\%\pm0\%\x{.00}$    & $\x{.0}100\%\pm0\%\x{0.0}$    & $0.95\pm0.01$         & $0.92\pm0.03$             & $0.94\pm0.01$         & $47.12\%\pm1.01\%$    & $74.84\%\pm0.89\%$    \\ \hline
                2                               & $\x{00.0}0\%\pm0\%\x{.00}$    & $\x{.0}100\%\pm0\%\x{0.0}$    & $0.95\pm0\x{0.0}$     & $0.89\pm0.04$             & $0.94\pm0.01$         & $47.38\%\pm0.30\%$    & $75.18\%\pm0.34\%$    \\ \hline
                4                               & $\x{00.0}0\%\pm0\%\x{.00}$    & $\x{.0}100\%\pm0\%\x{0.0}$    & $0.95\pm0\x{0.0}$     & $0.89\pm0.03$             & $0.94\pm0.01$         & $47.23\%\pm0.37\%$    & $75.01\%\pm0.29\%$    \\ \hline
                8                               & $\x{00}2.5\%\pm5.27\%$        & $\x{0}97.5\%\pm5.27\%$        & $0.95\pm0\x{0.0}$     & $0.86\pm0.02$             & $0.94\pm0.01$         & $47.28\%\pm0.50\%$    & $74.86\%\pm0.39\%$    \\ \hline
                16                              & $\x{0.0}10\%\pm7.34\%$        & $\x{.00}90\%\pm7.34\%$        & $0.95\pm0\x{0.0}$     & $0.86\pm0.02$             & $0.95\pm0.00$         & $47.30\%\pm0.39\%$    & $75.02\%\pm0.27\%$    \\ \hline
                32                              & $10.94\%\pm4.72\%$            & $89.06\%\pm4.72\%$            & $0.95\pm0\x{0.0}$     & $0.85\pm0.01$             & $0.95\pm0.00$         & $46.86\%\pm0.34\%$    & $74.39\%\pm0.33\%$    \\ \hline
                64                              & $\x{0}9.53\%\pm3.33\%$        & $90.47\%\pm3.33\%$            & $0.94\pm0.01$         & $0.87\pm0.01$             & $0.96\pm0.00$         & $46.54\%\pm1.08\%$    & $73.84\%\pm1.06\%$    \\
            \end{tabular}
        }
        \caption{\textbf{ViT-B/32} image encoder \textbf{without} weight regularization.}
    \end{subtable}
    \newline
    \newline
    \begin{subtable}{\linewidth}\centering
        \resizebox{\linewidth}{!}{
            \begin{tabular}{c|c|c|c|c|c|c|c}
                \makecell{Num. Unlearned\\Names}& IDIA TPR                                  & IDIA FNR                                      & $Sim_\mathit{Clean}$              & $Sim_\mathit{Backdoor}$               & $Sim_\mathit{Target}$                     & ImageNet Top-1                            & ImageNet Top-5                            \\ \hline
                1                               & $\x{00.0}0\%\pm0\%\x{0.0\uparrow}$        & $\x{.0}100\%\pm0\%\x{0.0\downarrow}$          & $\color{Green}0.96\pm0\uparrow$   & $\color{Green}0.93\pm0.03\uparrow$    & $\color{Green}0.95\pm0.01\uparrow$        & $\color{Green}48.45\%\pm0.33\%\uparrow$   & $\color{Green}76.11\%\pm0.28\%\uparrow$   \\ \hline
                2                               & $\x{00.0}0\%\pm0\%\x{0.0\uparrow}$        & $\x{.0}100\%\pm0\%\x{0.0\downarrow}$          & $\color{Green}0.96\pm0\uparrow$   & $0.89\pm0.03\x{\downarrow\;}$         & $\color{Green}0.95\pm0.01\uparrow$        & $\color{Green}48.43\%\pm0.27\%\uparrow$   & $\color{Green}76.15\%\pm0.33\%\uparrow$   \\ \hline
                4                               & $\color{red}\x{00}2.5\%\pm7.91\%\uparrow$ & $\color{red}\x{0}97.5\%\pm7.91\%\downarrow$   & $\color{Green}0.96\pm0\uparrow$   & $0.89\pm0.03\x{\downarrow\;}$         & $\color{Green}0.95\pm0\x{.00}\uparrow$    & $\color{Green}48.27\%\pm0.39\%\uparrow$   & $\color{Green}75.96\%\pm0.28\%\uparrow$   \\ \hline
                8                               & $\x{00}2.5\%\pm5.27\%\x{\uparrow\;}$      & $\x{0}97.5\%\pm5.27\%\x{\downarrow\;}$        & $\color{Green}0.96\pm0\uparrow$   & $\color{Green}0.87\pm0.02\uparrow$    & $\color{Green}0.95\pm0\x{.00}\uparrow$    & $\color{Green}48.36\%\pm0.26\%\uparrow$   & $\color{Green}75.83\%\pm0.22\%\uparrow$   \\ \hline
                16                              & $\color{red}10.63\%\pm5.93\%\uparrow$     & $\color{red}89.38\%\pm5.93\%\downarrow$       & $\color{Green}0.96\pm0\uparrow$   & $0.86\pm0.01\x{\downarrow\;}$         & $\color{Green}0.96\pm0\x{.00}\uparrow$    & $\color{Green}48.14\%\pm0.50\%\uparrow$   & $\color{Green}75.74\%\pm0.42\%\uparrow$   \\ \hline
                32                              & $\color{red}\x{0}12.5\%\pm5.31\%\uparrow$ & $\color{red}\x{0}87.5\%\pm5.31\%\downarrow$   & $0.95\pm0\x{\uparrow\;}$          & $\color{Green}0.86\pm0.01\uparrow$    & $\color{Green}0.96\pm0\x{.00}\uparrow$    & $\color{Green}47.64\%\pm0.75\%\uparrow$   & $\color{Green}75.17\%\pm0.71\%\uparrow$   \\ \hline
                64                              & $\color{red}11.56\%\pm4.67\%\uparrow$     & $\color{red}88.44\%\pm4.67\%\downarrow$       & $0.95\pm0\x{\uparrow\;}$          & $\color{Green}0.88\pm0.01\uparrow$    & $\color{Green}0.96\pm0\x{.00}\uparrow$    & $\color{Green}47.73\%\pm0.32\%\uparrow$   & $\color{Green}74.98\%\pm0.36\%\uparrow$   \\
            \end{tabular}
        }
        \caption{\textbf{ViT-B/32} image encoder \textbf{with} weight regularization.}
        \vspace{3ex}
    \end{subtable}
    \caption{Results for the experiments of the defense applied to the \textbf{ViT-B/32} image encoder. Weight regularization mitigates performance loss while slightly compromising defense success. Arrows indicate the change in value when using weight regularization in comparison to not using it. $Sim_\mathit{Clean}$, $Sim_\mathit{Backdoor}$ and $Sim_\mathit{Target}$ are cosine similarities which is why their maximum value is $1$. Green indicates better metrics, while red indicates worse metrics. All values were rounded to the second decimal place.}
    \label{tab:results_image_enc_vitb32}
\end{table*}

\begin{table*}[ht]
    \centering
    \begin{subtable}{\linewidth}\centering
        \resizebox{\linewidth}{!}{
            \begin{tabular}{c|c|c|c|c|c|c|c}
                \makecell{Num. Unlearned\\Names}& IDIA TPR                  & IDIA FNR                      & $Sim_\mathit{Clean}$  & $Sim_\mathit{Backdoor}$   & $Sim_\mathit{Target}$     & ImageNet Top-1        & ImageNet Top-5        \\ \hline
                1                               & $\x{00.}0\%\pm0\%\x{0.0}$ & $\x{.0}100\%\pm0\%\x{0.0}$    & $0.96\pm0$            & $0.94\pm0.02$             & $0.92\pm0.01$             & $56.20\%\pm0.31\%$    & $81.98\%\pm0.24\%$    \\ \hline
                2                               & $\x{00.}0\%\pm0\%\x{0.0}$ & $\x{.0}100\%\pm0\%\x{0.0}$    & $0.96\pm0$            & $0.88\pm0.04$             & $0.93\pm0.01$             & $56.14\%\pm0.28\%$    & $82.85\%\pm0.23\%$    \\ \hline
                4                               & $\x{00.}0\%\pm0\%\x{0.0}$ & $\x{.0}100\%\pm0\%\x{0.0}$    & $0.96\pm0$            & $0.85\pm0.04$             & $0.93\pm0.01$             & $56.28\%\pm0.43\%$    & $82.94\%\pm0.22\%$    \\ \hline
                8                               & $\x{0}5.0\%\pm8.74\%$     & $\x{0}95.0\%\pm8.74\%$        & $0.96\pm0$            & $0.83\pm0.02$             & $0.94\pm0\x{.00}$         & $56.24\%\pm0.31\%$    & $82.75\%\pm0.11\%$    \\ \hline
                16                              & $5.63\%\pm4.61\%$         & $94.38\%\pm4.61\%$            & $0.96\pm0$            & $0.83\pm0.02$             & $0.94\pm0\x{.00}$         & $56.49\%\pm0.27\%$    & $82.87\%\pm0.19\%$    \\ \hline
                32                              & $\x{0}7.5\%\pm5.93\%$     & $\x{0}92.5\%\pm5.93\%$        & $0.96\pm0$            & $0.82\pm0.01$             & $0.94\pm0\x{.00}$         & $56.03\%\pm0.24\%$    & $82.72\%\pm0.18\%$    \\ \hline
                64                              & $5.31\%\pm2.47\%$         & $94.69\%\pm2.47\%$            & $0.96\pm0$            & $0.83\pm0.01$             & $0.95\pm0\x{.00}$         & $55.87\%\pm0.18\%$    & $82.41\%\pm0.19\%$    \\
            \end{tabular}
        }
        \caption{\textbf{ViT-B/16} image encoder \textbf{without} weight regularization.}
    \end{subtable}
    \newline
    \newline
    \begin{subtable}{\linewidth}\centering
        \resizebox{\linewidth}{!}{
            \begin{tabular}{c|c|c|c|c|c|c|c}
                \makecell{Num. Unlearned\\Names}& IDIA TPR                              & IDIA FNR                                  & $Sim_\mathit{Clean}$                      & $Sim_\mathit{Backdoor}$               & $Sim_\mathit{Target}$                     & ImageNet Top-1                            & ImageNet Top-5                            \\ \hline
                1                               & $\x{00.}0\%\pm0\%\x{0.0\uparrow}$     & $\x{.0}100\%\pm0\%\x{0.0\downarrow}$      & $\color{Green}0.97\pm0\x{0.0}\uparrow$    & $0.94\pm0.01\x{\downarrow\;}$         & $\color{Green}0.93\pm0.01\uparrow$        & $\color{Green}57.02\%\pm0.26\%\uparrow$   & $\color{Green}83.53\%\pm0.21\%\uparrow$   \\ \hline
                2                               & $\x{00.}0\%\pm0\%\x{0.0\uparrow}$     & $\x{.0}100\%\pm0\%\x{0.0\downarrow}$      & $\color{Green}0.97\pm0\x{0.0}\uparrow$    & $\color{Green}0.89\pm0.03\uparrow$    & $\color{Green}0.94\pm0.01\uparrow$        & $\color{Green}56.97\%\pm0.22\%\uparrow$   & $\color{Green}83.40\%\pm0.19\%\uparrow$   \\ \hline
                4                               & $\x{00.}0\%\pm0\%\x{0.0\uparrow}$     & $\x{.0}100\%\pm0\%\x{0.0\downarrow}$      & $\color{Green}0.97\pm0\x{0.0}\uparrow$    & $\color{Green}0.87\pm0.03\uparrow$    & $\color{Green}0.95\pm0.01\uparrow$        & $\color{Green}57.08\%\pm0.36\%\uparrow$   & $\color{Green}83.54\%\pm0.19\%\uparrow$   \\ \hline
                8                               & $\x{0}5.0\%\pm8.74\%\x{\uparrow\;}$   & $\x{0}95.0\%\pm8.74\%\x{\downarrow\;}$    & $\color{Green}0.97\pm0\x{0.0}\uparrow$    & $\color{Green}0.85\pm0.02\uparrow$    & $\color{Green}0.95\pm0\x{.00}\uparrow$    & $\color{Green}57.04\%\pm0.23\%\uparrow$   & $\color{Green}83.33\%\pm0.15\%\uparrow$   \\ \hline
                16                              & $\color{red}8.13\%\pm5.15\%\uparrow$  & $\color{red}91.88\%\pm5.15\%\downarrow$   & $\color{Green}0.97\pm0\x{0.0}\uparrow$    & $\color{Green}0.84\pm0.02\uparrow$    & $\color{Green}0.95\pm0\x{.00}\uparrow$    & $\color{Green}57.24\%\pm0.87\%\uparrow$   & $\color{Green}83.47\%\pm0.21\%\uparrow$   \\ \hline
                32                              & $\color{red}8.44\%\pm4.43\%\uparrow$  & $\color{red}91.56\%\pm4.43\%\downarrow$   & $\color{Green}0.97\pm0\x{0.0}\uparrow$    & $\color{Green}0.83\pm0.01\uparrow$    & $\color{Green}0.95\pm0\x{.00}\uparrow$    & $\color{Green}56.86\%\pm0.21\%\uparrow$   & $\color{Green}83.31\%\pm0.23\%\uparrow$   \\ \hline
                64                              & $\color{red}6.41\%\pm3.16\%\uparrow$  & $\color{red}93.59\%\pm3.16\%\downarrow$   & $0.96\pm0.01\x{\uparrow\;}$               & $\color{Green}0.84\pm0.01\uparrow$    & $\color{Green}0.96\pm0\x{.00}\uparrow$    & $\color{Green}56.64\%\pm0.19\%\uparrow$   & $\color{Green}83.07\%\pm0.20\%\uparrow$   \\
            \end{tabular}
        }
        \caption{\textbf{ViT-B/16} image encoder \textbf{with} weight regularization.}
        \vspace{3ex}
    \end{subtable}
    \caption{Results for the experiments of the defense applied to the \textbf{ViT-B/16} image encoder. Weight regularization mitigates performance loss while slightly compromising defense success. Arrows indicate the change in value when using weight regularization in comparison to not using it. $Sim_\mathit{Clean}$, $Sim_\mathit{Backdoor}$ and $Sim_\mathit{Target}$ are cosine similarities which is why their maximum value is $1$. Green indicates better metrics, while red indicates worse metrics. All values were rounded to the second decimal place.}
    \label{tab:results_image_enc_vitb16}
\end{table*}

\begin{table*}[ht]
    \centering
    \begin{subtable}{\linewidth}\centering
        \resizebox{\linewidth}{!}{
            \begin{tabular}{c|c|c|c|c|c|c|c}
            \makecell{Num. Unlearned\\Names}& IDIA TPR                          & IDIA FNR                      & $Sim_\mathit{Clean}$  & $Sim_\mathit{Backdoor}$   & $Sim_\mathit{Target}$ & ImageNet Top-1            & ImageNet Top-5    \\ \hline
                1                               & $\x{00.0}0\%\pm0\%\x{0.00}$   & $\x{.0}100\%\pm0\%\x{0.00}$   & $0.96\pm0$            & $0.95\pm0.01$             & $0.98\pm0$            & $48.77\%\pm0.35\%$    & $77.81\%\pm0.46\%$    \\ \hline
                2                               & $\x{0}15.0\%\pm24.15\%$       & $\x{.00}85\%\pm24.15\%$       & $0.95\pm0$            & $0.95\pm0\x{.00}$         & $0.98\pm0$            & $48.26\%\pm0.41\%$    & $77.42\%\pm0.38\%$    \\ \hline
                4                               & $\x{0}25.0\%\pm26.35\%$       & $\x{0}75.0\%\pm26.35\%$       & $0.95\pm0$            & $0.95\pm0\x{.00}$         & $0.98\pm0$            & $48.28\%\pm0.35\%$    & $77.44\%\pm0.25\%$    \\ \hline
                8                               & $\x{0}30.0\%\pm12.08\%$       & $\x{0}70.0\%\pm12.08\%$       & $0.96\pm0$            & $0.95\pm0\x{.00}$         & $0.99\pm0$            & $48.58\%\pm0.30\%$    & $77.39\%\pm0.13\%$    \\ \hline
                16                              & $\x{0}37.5\%\pm\x{0}9.77\%$   & $\x{0}62.5\%\pm\x{0}9.77\%$   & $0.95\pm0$            & $0.94\pm0\x{.00}$         & $0.99\pm0$            & $48.21\%\pm0.19\%$    & $76.90\%\pm0.15\%$    \\ \hline
                32                              & $41.25\%\pm\x{0}6.56\%$       & $58.75\%\pm\x{0}6.56\%$       & $0.95\pm0$            & $0.94\pm0\x{.00}$         & $0.98\pm0$            & $47.41\%\pm0.13\%$    & $76.31\%\pm0.16\%$    \\ \hline
                64                              & $46.72\%\pm\x{0}4.74\%$       & $53.28\%\pm\x{0}4.74\%$       & $0.94\pm0$            & $0.94\pm0\x{.00}$         & $0.98\pm0$            & $46.49\%\pm0.20\%$    & $75.67\%\pm0.14\%$    \\
            \end{tabular}
        }
        \caption{\textbf{ResNet-50} image encoder \textbf{without} weight regularization.}
        \vspace{3ex}
    \end{subtable}
    \newline
    \newline
    \begin{subtable}{\linewidth}\centering
        \resizebox{\linewidth}{!}{
            \begin{tabular}{c|c|c|c|c|c|c|c}
                \makecell{Num. Unlearned\\Names}& IDIA TPR                                      & IDIA FNR                                      & $Sim_\mathit{Clean}$              & $Sim_\mathit{Backdoor}$                   & $Sim_\mathit{Target}$ & ImageNet Top-1                            & ImageNet Top-5                        \\ \hline
                1                               & $\x{00.0}0\%\pm0\%\x{0.00\downarrow}$         & $\x{.0}100\%\pm0\%\x{0.00\downarrow}$         & $0.96\pm0\x{\uparrow\;}$          & $0.95\pm0.01\x{\downarrow\;}$             & $0.98\pm0$            & $\color{Green}49.07\%\pm0.36\%\uparrow$   & $\color{Green}78.09\%\pm0.42\%\uparrow$   \\ \hline
                2                               & $\x{0}15.0\%\pm24.15\%\x{\downarrow\;}$       & $\x{.00}85\%\pm24.15\%\x{\downarrow\;}$       & $\color{Green}0.96\pm0\uparrow$   & $0.95\pm0\x{.00\downarrow}$               & $0.98\pm0$            & $\color{Green}48.64\%\pm0.42\%\uparrow$   & $\color{Green}77.79\%\pm0.31\%\uparrow$   \\ \hline
                4                               & $\color{red}\x{0}30.0\%\pm28.39\%\uparrow$    & $\color{red}\x{0}70.0\%\pm28.39\%\downarrow$  & $0.95\pm0\x{\uparrow\;}$          & $0.95\pm0\x{.00\downarrow}$               & $0.98\pm0$            & $\color{Green}48.64\%\pm0.37\%\uparrow$   & $\color{Green}77.76\%\pm0.23\%\uparrow$   \\ \hline
                8                               & $\color{red}33.75\%\pm11.86\%\uparrow$        & $\color{red}66.25\%\pm11.86\%\downarrow$      & $0.96\pm0\x{\uparrow\;}$          & $\color{red}0.94\pm0\x{.00}\downarrow$    & $0.99\pm0$            & $\color{Green}48.94\%\pm0.25\%\uparrow$   & $\color{Green}77.67\%\pm0.17\%\uparrow$   \\ \hline
                16                              & $\color{red}38.13\%\pm\x{0}9.06\%\uparrow$    & $\color{red}61.88\%\pm\x{0}9.06\%\downarrow$  & $\color{Green}0.96\pm0\uparrow$   & $0.94\pm0\x{.00\downarrow}$               & $0.99\pm0$            & $\color{Green}48.56\%\pm0.16\%\uparrow$   & $\color{Green}77.20\%\pm0.17\%\uparrow$   \\ \hline
                32                              & $\color{red}44.06\%\pm\x{0}6.50\%\uparrow$    & $\color{red}55.94\%\pm\x{0}6.50\%\downarrow$  & $0.95\pm0\x{\uparrow\;}$          & $0.94\pm0\x{.00\downarrow}$               & $0.98\pm0$            & $\color{Green}47.73\%\pm0.19\%\uparrow$   & $\color{Green}76.59\%\pm0.26\%\uparrow$   \\ \hline
                64                              & $\color{red}50.31\%\pm\x{0}5.30\%\uparrow$    & $\color{red}49.69\%\pm\x{0}5.30\%\downarrow$  & $0.94\pm0\x{\uparrow\;}$          & $\color{red}0.93\pm0\x{.00}\downarrow$    & $0.98\pm0$            & $\color{Green}46.93\%\pm0.20\%\uparrow$   & $\color{Green}75.98\%\pm0.13\%\uparrow$   \\
            \end{tabular}
        }
        \caption{\textbf{ResNet-50} image encoder \textbf{with} weight regularization.}
        \vspace{3ex}
    \end{subtable}
    \caption{Results for the experiments of the defense applied to the \textbf{ResNet-50} image encoder. Weight regularization mitigates performance loss while slightly compromising defense success. Arrows indicate the change in value when using weight regularization in comparison to not using it. $Sim_\mathit{Clean}$, $Sim_\mathit{Backdoor}$ and $Sim_\mathit{Target}$ are cosine similarities which is why their maximum value is $1$. Green indicates better metrics, while red indicates worse metrics. All values were rounded to the second decimal place.}
    \label{tab:results_image_enc_rn50}
\end{table*}

\end{document}